\documentclass[12pt, twoside]{report}

\usepackage[utf8]{inputenc}
\usepackage{graphicx}
\graphicspath{ {Images/} }
\usepackage{lipsum}
\usepackage[a4paper,width=150mm,top=25mm,bottom=25mm,bindingoffset=10mm]{geometry}
\usepackage{fancyhdr}
\usepackage{caption}
\usepackage{subcaption}

\usepackage{cite}

\usepackage{setspace}
\usepackage{algorithm}
\usepackage{algpseudocode}
\usepackage{tablefootnote}
\usepackage{textcomp}
\usepackage{amsmath,amssymb,amsfonts}
\usepackage{color}
 
\DeclareMathOperator*{\argmaxB}{argmax} 
\newcommand{\R}{\mathbb{R}}
\usepackage{soul}

\soulregister{\ref}7 
\soulregister{\cite}7 

\begin{document}

\begin{titlepage}
    \begin{center}
        \vspace*{1cm}
        
        \LARGE
        \textbf{Topics in Deep Learning and Optimization Algorithms for IoT Applications in Smart Transportation}
        
%
        
        \vspace{1.5cm}
        
        \Large
        Hongde Wu, B.Eng\\
        
        \vspace{0.5cm}
        
        Supervised by Dr. Mingming Liu\\
        
        \vfill
        
        \includegraphics[width=0.25\textwidth]{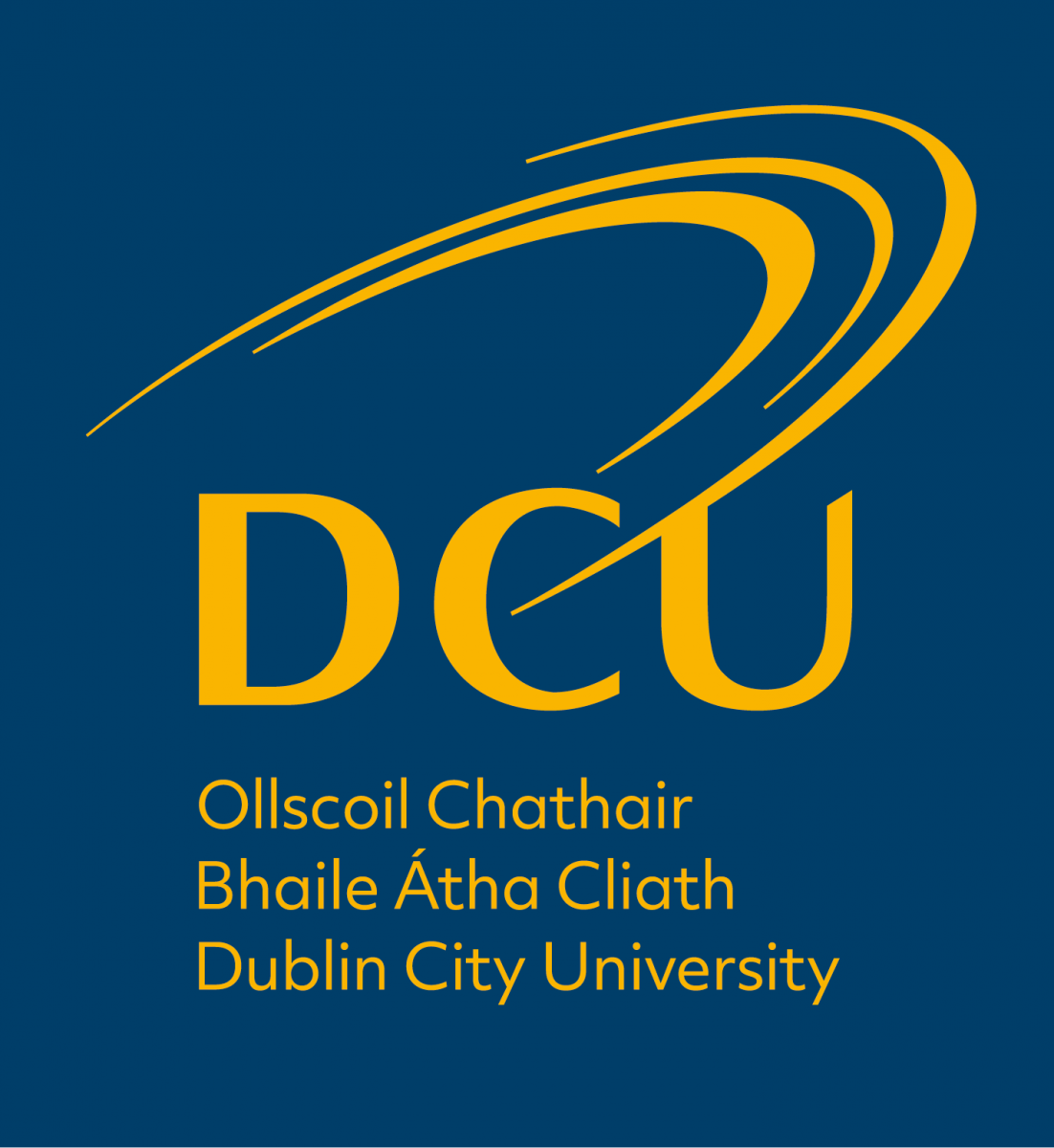}
        
        \vspace{0.8cm}
        
        \Large
        A thesis submitted for the award of Master of Engineering (M.Eng)
        
        \vspace{0.5cm}
        
        \Large
	    \textsc{School of Electronic Engineering\\
	    Dublin City University}
	    
	    \begin{flushright}
	
	    \Large
	    August 2022
	
	    \end{flushright}
        
    \end{center}
\end{titlepage}


\chapter*{Declaration}
I hereby certify that this thesis, which I now submit for assessment on the programme of study leading to the award of Master of Engineering is entirely my own work, and that I have exercised reasonable care to ensure that the work is original, and does not to the best of my knowledge breach any law of copyright, and has not been taken from the work of others save and to the extent that such work has been cited and acknowledged within the text of my work. \\ \hspace*{\fill} \\

\noindent Signed: 

\noindent \rule[-7pt]{7.0cm}{0.5pt}  \\ \hspace*{\fill} \\
\noindent ID No.: 20216606

\noindent Date: 28/08/2022

\chapter*{Acknowledgements}
I would first like to thank my supervisor Dr. Mingming Liu for giving me the great support during my Master of research life. Without his constant encouragement, guidance and pursuit of scientific rigor, my achievements would not exist. There are too many people at DCU that I would like to thank and feel very happy to work with them, including Prof. Noel E. O’Connor, Prof. Jennifer Bruton, Dr. Amy Hall and my lab colleagues Sen Yan, Dr. Hoa Xuan Nguyen and Shaoshu Zhu.

Finally, I would like to thank the research master scholarship sponsored by the Faculty of Engineering and Computing at DCU as well was the generous support from the SFI Insight Centre for Data Analytics under grant number SFI/12/RC/2289\_P2 for some related research works reported in this thesis.

Most importantly, I would like to thank the mental supports from my family, my dear friends and Wenting Luo in the duration of the master study.

\chapter*{List of Publications}

The following \textbf{journals/conference} papers have been \textbf{submitted/published} during the course of my MEng:
\\

\begin{itemize}
	\item[1.]  \textbf{Hongde Wu}, Noel E. O’Connor, Jennifer Bruton, Amy Hall and Mingming Liu, ``Real-Time Anomaly Detection for an ADMM-based Optimal Transmission Frequency Management System for IoT Edge Devices", Sensors, 2022. \textbf{[published]}
	\\
	
	\item[2.] \textbf{Hongde Wu} and Mingming Liu, ``Lane-GNN: Integrating GNN for Predicting Drivers’ Lane Change Intention", IEEE International Intelligent Transportation Systems Conference (ITSC), 2022. \textbf{[accepted]}
	\\
	
	\item[3.] \textbf{Hongde Wu}, Noel E. O’Connor, Jennifer Bruton and Mingming Liu, ``An ADMM-based Optimal Transmission Frequency Management System for IoT Edge Intelligence", IEEE World Forum on Internet of Things (WF-IoT), 2021. \textbf{[published]}
	\\
	
	\item[4.] Zhengyong Chen, \textbf{Hongde Wu}, Noel E. O’Connor and Mingming Liu, ``A Comparative Study of Using Spatial-Temporal Graph Convolutional Networks for Predicting Availability in Bike Sharing Schemes", IEEE International Intelligent Transportation Systems Conference (ITSC), 2021. \textbf{[published]}
	\\

	\item[5.] \textbf{Hongde Wu}, Zhengyong Chen, Noel E. O’Connor and Mingming Liu, ``Optimal Distributed Bandwidth Allocation in NB-IoT Networks", International Conference on Internet-of-Things Design and Implementation (IoTDI), 2021. \textbf{[published]}

\end{itemize}

\begin{singlespacing}
\tableofcontents
\end{singlespacing}

\listoffigures

\listoftables

\chapter*{List of Abbreviations}
ADMM \makebox[1.1cm]{\dotfill} Alternating Direction Method of Multipliers

\noindent MWF \makebox[1.4cm]{\dotfill} Maximum Writing Frequency

\noindent DFWF \makebox[1.2cm]{\dotfill} Data Flow Writing Frequency

\noindent SS \makebox[2.05cm]{\dotfill} Simulation System

\noindent RS \makebox[2.0cm]{\dotfill} Real-world System

\noindent ASTGCN \makebox[0.8cm]{\dotfill} Attention-based Spatial-Temporal Graph Convolutional Network

\noindent AAM \makebox[1.6cm]{\dotfill} Adaptive Adjacency Matrix 

\noindent SAS \makebox[1.85cm]{\dotfill} Speed Advisory System

\noindent TCNN \makebox[1.40cm]{\dotfill} Temporal Convolutional Neural Network

\noindent ATGCN \makebox[1.15cm]{\dotfill} Attention-based Temporal Graph Convolutional Networks

\noindent SID \makebox[2.0cm]{\dotfill} Spectral Information Divergence

\chapter*{}
\vspace{-1.5in}
\begin{center}

    \large
    \textbf{Topics in Deep Learning and Optimization Algorithms for IoT Applications in Smart Transportation}
    
    \large
    \vspace{0.4cm}
    \textbf{Hongde Wu}
    
    \vspace{0.4cm}
    \large
    \textbf{Abstract}
    
\end{center}

\vspace{0.9cm}
\begin{singlespacing}

Nowadays, the Internet of Things (IoT) has become one of the most important technologies which enables a variety of connected and intelligent applications in smart cities. The smart decision making process of IoT devices not only relies on the large volume of data collected from their sensors, but also depends on advanced optimization theories and novel machine learning technologies which can process and analyse the collected data in specific network structure. Therefore, it becomes practically important to investigate how different optimization algorithms and machine learning techniques can be leveraged to improve system performance for real world IoT applications in a graph-based environment. 

As one of the most important vertical domains for IoT applications, smart transportation system has played a key role for providing real-world information and services to citizens by making their access to transport facilities easier and thus it is one of the key application areas to be explored in this thesis.  

In a nutshell, this thesis covers three key topics related to applying mathematical optimization and deep learning methods to IoT networks. In the first topic, we propose an optimal transmission frequency management scheme using decentralized ADMM-based method in a IoT network and introduce a mechanism to identify anomalies in data transmission frequency using an LSTM-based architecture. In the second topic, we leverage graph neural network (GNN) for demand prediction for shared bikes. In particular, we introduce a novel architecture, i.e., attention-based spatial temporal graph convolutional network (AST-GCN), to improve the prediction accuracy in real world datasets. In the last topic, we consider a highway traffic network scenario where frequent lane changing behaviors may occur with probability. A specific GNN based anomaly detector is devised to reveal such a probability driven by data collected in a dedicated mobility simulator.

\end{singlespacing}


\chapter{Introduction}

\textbf{Abstract:} In this chapter, we present an overview knowledge of the Internet of things and smart transportation as our research background. With these background, we highlight the research objective and key contributions. We organise this chapter as follows: in section \ref{overview} we introduce the Internet of things and smart transportation to readers as our work is based on this context; in section \ref{objectives} we discuss the research problems and objectives and highlight the research contributions in section \ref{contributions}; in section \ref{structure} we describe the thesis structure which matches with our research contributions in specific chapters.

\section{Overview}\label{overview}

Internet of Things (IoT) has played a key role in our daily life as it enables various intelligent applications in our cities. As one of the applications of IoT, which is most related to our daily travelling, smart transportation has served our citizens by offering real-world information and making transport facilities more convenient. Here we give a short background of IoT and smart transportation to provide a better scope that this thesis will cover.

\subsection{Internet of things} \label{IoTintro}
The Internet of Things (IoT) is a paradigm which is increasingly getting attention in modern wireless telecommunications. The basic concept of IoT is that ubiquitous objects around us, such as sensors and mobile phones, are able to communicate and cooperate with each other to solve a common problem \cite{giusto2010internet}. Specifically, the IoT network includes a variety of smart devices with the functions of connecting, exchanging and sharing data with each other over the Internet \cite{madakam2015internet}. In order to enable these functions in IoT networks, one of the key technologies is the Radio-Frequency IDentification (RFID) technology, which allows smart devices to exchange the information of device identification to the target receivers (e.g., Cloud facilities) by using RFID identifier \cite{jia2012rfid}. Another foundational technique is the wireless network used for connecting intelligent devices to monitor the environment. With these two techniques, an IoT system can capture real-time environmental data through sensors embedded in IoT devices. The data emitted from the system can be transmitted to the Cloud via gateways for further storage, process and analysis \cite{2011A}. Typically, in a cloud-dominant centralised architecture, Artificial Intelligence (AI) enabled computing nodes are often integrated and implemented at the cloud side, with an intention to collect the useful information from the transmitted data centrally and provide better insight for users to make decisions. Some recent IoT applications relying on this architecture are described in the following works, such as in the field of healthcare monitoring, traffic monitoring and environmental resource monitoring \cite{2018Brokering, 2018Economic, St2016Cloud, li2012compressed, he2012integration}. 

In a word, with the advances in wireless communication and sensor networks, IoT has been gaining attention in the area related to our daily life and more and more 'things' or smart objects are being involved in IoT networks. As a result, these IoT-related technologies have also made a large impact on new information and communications technology (ICT). However, the advanced IoT networks also come with inevitable shortcomings, especially those usually require the decision-making process to be conducted at the device side or edge side for better security \cite{9301390} and privacy protection \cite{9163078}.  Specifically, in a typical IoT scenario where data streams from various IoT devices can be transmitted to the Cloud and stored on a cloud database.  Our initial observation is that most IoT devices start to transmit data at a fixed transmission frequency, and such a transmission frequency is typically set by default or pre-defined by the device manufacturer with limited options made available to users. However, some advanced IoT devices with edge intelligence, e.g. Raspberry Pis and the Jetson series toolkit from Nvidia, can now be programmed to promptly respond to changes in the external environment \cite{10.1117/12.2571307, 8230004}, and can also be deployed with deep learning algorithms to satisfy stringent low-latency transmission requirements for time-sensitive IoT applications \cite{9287960, 9289509}. This approach does not sufficiently cater for a practical situation where groups of IoT devices may work collaboratively with limited system resources restricted by the operational environment. In fact, implementing IoT devices in a resource-constrained environment may impose two interesting problems in the design of IoT networks: 1) how to determine an adaptive transmission frequency for each IoT device so that an overall utility of the group of devices can be maximised in response to the dynamic changes of the environment; 2) how to ensure that different kinds of network resources can be better managed in a way that heterogeneous IoT devices can be engaged with the network in a secure, privacy-aware and plug-and-play manner. In order to address the mentioned problems, the first topic of this thesis is to propose a transmission frequency system for edge devices in an IoT network with a robust anomaly detection mechanism.


\subsection{Smart transportation} \label{smartcityinro}


On the one hand, IoT has played a key role in enabling the smart city, which combines data collection, analysis and decision making \cite{batty2012smart}. On the other hand, the smart city has become a terminology along with IoT, a novel city management approach to establish a collaborative society, where the data from daily life is leveraged to provide decisions for city management \cite{albino2015smart}.

Obviously, as the population is growing, the need for transportation increase dramatically and therefore smart transportation becomes the most challenging part of a smart city. To enable smart applications in modern transportation, advanced technologies, e.g. intelligent transportation system (ITS), have been proposed to provide creative insight for traffic management and improve user experience by providing proper information about the traffic network \cite{lin2017intelligent}. For instance, a smart parking system can save time for drivers, by informing drivers of the availability of parking spaces \cite{khanna2016iot}; carbon emissions and pollutants may also be minimised by recommending a shortest path to drivers for their parking search process \cite{agarana2017minimizing}. To sum up, smart transportation has shed a light on modern traffic management and satisfied the need of citizens in daily commuting. However, there are still open problems in smart transportation, such as traffic demand prediction, accident prevention, traffic flow prediction; cloud-based multi-agents planning; energy consumption \cite{rodrigue2020geography}, which are more challenging to deal with using conventional means of traffic management.

\subsubsection{Bike availability prediction}

As one of the common modes of transportation, bikes provide a healthy and convenient way for short-distance travel and sharing bikes have become prevalent in our cities. Also, an efficient bike-sharing system can not only reduce cost and commute time for urban commuters but also effectively mitigate the level of air pollution emissions generated in cities \cite{otero2018health}. 
However, bike availability prediction is one of the challenging problems in traffic demand prediction because the available number of bikes tends to be unbalanced, particularly at peak demand dates and hours \cite{hulot2018towards}. Therefore, an important consideration to make the bike-sharing system efficient is to balance supply and demand in the bike-sharing network \cite{raviv2013optimal}. To do this, traditional management methods such as manual monitoring systems, have been deployed to enable the relocation of bikes across different stations using other means of transportation, e.g. trucks \cite{7313194, raviv2013static}. However, this approach can easily lead to supply-demand imbalance due to estimation errors of system operators and unexpected traffic delays during the bike transition. Thus, due to the uncertainty of departure and arrival of bikes at any bike station, it is important to take a more proactive approach by accurately predicting the number of bikes that will be available for users to access at any given time and location. However, on the topic of traffic demand prediction, most of the work focus on taxi demand/availability prediction \cite{yao2018deep, yao2019revisiting} and limited work discusses the topic of availability prediction for sharing bike. Meanwhile, the current approaches are not able to forecast the availability precisely because of the weakness in traffic feature extraction and modelling. Therefore, in this thesis, the problem of sharing-bike availability prediction using graph neural network (GNN) is our second topic to discuss.

\subsubsection{Lane change detection}
As another challenge of smart transportation, accident prevention ensures driving safety and deserves more attention. Even if the traffic suggestions and regulations have been authorized to ensure a safe driving environment and minimise the chances of a traffic accident as much as possible, malicious driving intentions (e.g., acute acceleration; frequent lane changing) still play a threat to traffic safety and disturb the normal traffic flow. For instance, a speed advisory system (SAS) offers speed guidance for ensuring driving safety, but the vehicles tend to be driven with unexpected acceleration and lane changing behaviours \cite{jeon2014effects}, once they leave the road segment with SAS. Therefore, detection of driving intentions has been involved in traffic management, alarming for intervention when the driving safety may be under threat, such as traffic incidents\cite{asakura2017incident} \cite{hawas2007fuzzy}, traffic congestion \cite{wang2018locality} and malicious driving \cite{8886013}.
It is worth paying attention to frequent lane changing, which may easily result in severe traffic accidents on highway networks. Existing approaches, such as hidden Markov model (HMM) \cite{li2016lane} and LSTM-based methods \cite{tang2020driver, 8813987}, have been found less capable in dealing with the lane changing detection problems as they can not model the traffic data with natural geographical information (e.g, the connection between lanes) sufficiently. Therefore, the last topic in this thesis concerns the detection for lane changing intention using GNN to leverage the geographical information on the highway network, to improve the detection performance.


\section{Research objectives}\label{objectives}

\textbf{Research objectives 1:} Optimise transmission frequencies for edge devices in IoT network with robust anomaly detection mechanism.

We consider the two problems of the design of IoT networks, as discussed in section \ref{IoTintro}. Our key assumption is that different IoT devices may have different priority levels when transmitting data in a resource-constrained environment and that those priority levels may only be locally defined and accessible by edge devices for privacy concerns. With these in mind, the research objective is to optimise the transmission frequencies for a group of IoT edge devices under practical constraints. We aim at establishing a transmission frequency management system which can allocate optimal transmission frequencies to IoT devices and maximise the overall utility of the edge devices in the IoT network in a decentralised manner. In order to ensure the security of the system, we shall also devise an anomaly detector, on top of the designed optimal transmission management system, which can effectively identify abnormal transmission frequencies in different settings. The anomaly detector is expected to only leverage limited information from the IoT system. We will investigate both mathematical rule-based and deep learning based approaches, and examine their efficacy in tackling such challenges.

\noindent \textbf{Research objectives 2:} Availability prediction for the sharing-bike scheme using spatial-temporal graph convolutional network. 

As to a research topic related to smart transportation, we first consider the problem of availability prediction for sharing bikes. The research objective is to present a availability prediction system which can forecast the available number of sharing bikes among different bike stations accurately and promptly using models trained on realistic data. In particular, spatial-temporal graph convolutional network (ST-GCN), as a powerful variant of graph convolutional networks (GCN) which aims to capture the relationship of data contained in the graphical nodes across both spatial and temporal dimensions, is applied for improving the prediction accuracy. Recently, graph based solutions have caught much attention in the literature as they have shown efficacy in improving traffic management. We shall apply spatial-temporal graph convolutional network (ST-GCN) to capture the relationship of data between graph nodes and compare its performance with other schemes to illustrate its efficacy in chapter \ref{bike}. Moreover, the impacts of different modelling methods of adjacency matrices shall be investigated.

\noindent \textbf{Research objectives 3:} Detecting lane changing intention on highway network scenario using graph neural network.

The last research objective is related to driving safety. As mentioned previously in section \ref{smartcityinro}, frequent lane changing intention threatens driving safety on the highway network. The objective of this part is to develop an algorithm which is able to detect the frequent lane changing behaviour on highway network using graph-based deep learning methods. As we shall see, the proposed algorithm will be able to forecast the lane changing probability of vehicles on a segment of the highway network in real-time.


\section{Thesis contributions}\label{contributions}
The thesis discusses three topics related to IoT and smart transportation. The contributions of the thesis can be summarised as followed:

\begin{itemize}
	\item In chapter \ref{IoT}, we propose a transmission frequency management system which is able to find the optimal transmission frequency for each IoT device, in order to maximise the overall utility in a resource-constrained, privacy-aware environment. Design an anomaly detector to ensure the transmission frequencies of the proposed IoT transmission frequency management system are in good order. 
	
	\item In chapter \ref{bike}, we design a deep learning architecture by combining attention mechanism with the spatial-temporal graph neural network, to better predict the sharing-bike availability based on realistic datasets. Furthermore, we also discuss the impacts of different modelling methods of adjacency matrices on the proposed architecture.

	\item In chapter \ref{lane}, we apply a refined version of a graph neural network, to predict the lane changing intention and analysis the pattern of driving data for the purpose of model interpretability.
	
\end{itemize}

\section{Thesis structure}\label{structure}

The thesis is organised as follows:

\begin{itemize}
	\item Chapter 1 introduces the background, our research objectives, thesis contribution and structure.
	\item Chapter 2 approaches the first research objective by applying optimisation and deep learning method to IoT systems.
	\item Chapter 3 achieves the second research objective by leveraging graph neural networks to forecast the sharing-bike availability, based on the data collected by IoT devices embedded in bike stations.
	\item Chapter 4 tackles the third research problem by using graph neural network to analyse the driving patterns and predict the lane changing intention, based on the data generated from a novel mobility simulator.
	\item Chapter 5 summarises the thesis and highlights the potential directions for future work.
\end{itemize}

%

\chapter{Transmission frequency management system in IoT network} \label{IoT}
\graphicspath{ {Chapter02/} }
%
%
%
%

\textbf{Abstract:} In this chapter, we propose a transmission frequency management system with anomaly detector in the context of the Internet of Things. The anomaly detector is able to enhance the system security by detecting different types of manipulations, which lead the IoT devices to transmit data violating the desired transmission frequencies. The work presented in this chapter has been published in \cite{wu2021admm, wu2022real}.

\section{Introduction}

In a cloud-based IoT solution, data from various IoT devices need to be pushed to cloud-based database instances in real-time. However, the capacity of storage space is limited. For instance, an IBM Cloudant database instance allows 1 GB of data storage with 10 writes/sec for its Lite Plan users, and 20 GB of data storage with 50 writes/sec for its Standard Plan users \cite{bienko2015ibm}. Given this scenario with the limited storage resource, if the Maximum Writing Frequency (MWF) of the data is not managed properly, it can be envisioned that a writing congestion event, e.g. a REST-API writing failure, can be triggered for a group of IoT devices. Also, another concern is on privacy, which, in our context, refers to the fact that the mapping between the utility and the transmission dynamics of a given IoT device should not be revealed to any unrelated devices, third-party gateways and untrusted cloud units or instances. If this mapping information is revealed publicly it may be possible for an attacker to identify which IoT device is more vulnerable in a given system \cite{2020A}.

To solve this challenge, in this chapter we propose a transmission frequency management system for IoT edge devices in a decentralized architecture with anomaly detection mechanisms. Thus the MWF can be managed optimally by a group of IoT devices and any abnormal writing frequency occurrences can be detected by the gateway. To carry out  optimisation, we assume that each IoT device is associated with a utility function with some concavity \cite{7106504, 2019Utility}, in a way that only the user of the device can specify. Here, the utility refers to how a user can practically benefit from a given Data Flow Writing Frequency (DFWF). For instance, a utility function can easily describe the accuracy of a trained model with respect to DFWF of a given IoT device for an Edge AI type of IoT application \cite{LV202190}. Furthermore, as previously mentioned, such a utility function may also potentially reflect the significance or vulnerability of an IoT device in a specific scenario. For instance, a faster transmission frequency of a webcam in a bank system may be more desirable, i.e., have higher utility, especially in an emergency, than that  of a $CO_2$ detector.

With this idea in mind, our main objective in our system is to maximise the overall utility of the group of IoT devices given the predefined and limited MWF and storage capacity of the database. We will show that the presented challenge can be formulated as a concave optimisation problem with constraints. This problem will then be solved using the well-known Alternating Direction Method of Multipliers (ADMM) algorithm \cite{boyd2011distributed} in a decentralised optimisation framework where each utility function is locally defined on the edge device and will not be revealed to any unrelated devices and untrusted management platforms, such as other smart gateways and cloud units/instances. The proposed solution aims to provide flexibility in data transmission for IoT systems and applications, especially in resource-constrained environments. As we shall see, the designed system is fully autonomous and can be easily deployed to optimally manage various IoT transmission frequencies with anomaly detection capabilities. 

We note that significant work on anomaly detection has been undertaken in IoT context: for instance, Liu et al. \cite{8408731} proposed a detector for on and off attack by a malicious network node in an industrial IoT site; Anthi et al. \cite{8379722} represented an intrusion detection system for an IoT system to identify the Denial of Service (DoS) attacks; Ukil et al. \cite{7474197} discussed the detection of anomalies in healthcare analytics based on IoT by analysing the cardiac signal; and Hu et al. \cite{traffic} proposed a Context-augmented Graph Auto-encoder (Con-GAE) for anomaly detection in traffic monitoring. However, the anomalies defined in these works are largely based on \textit{tempering with contents in data packets} transmitted by IoT devices (e.g., changing a data value from ``A" to ``B" in the transmitted file \cite{miller2016detection}) and no approach has been found on anomaly detection for an IoT data transmission frequency system involved with an optimal iterative scheme. Therefore, in this thesis, we are interested in detecting the malicious manipulations leading to a change of transmission frequency as a result of the anomalies happening on the edge devices.

The contributions of this chapter can be summarised as follows:
\begin{itemize}    	
	\item[1.] We propose an optimisation framework for an IoT network so that the transmission frequency of the connected IoT devices can be dynamically adjusted to their optimal values in a low latency through an ADMM-based iterative optimisation method. 
	
	\item[2.] We design an anomaly detector on top of the frequency management system, which is able to infer anomalies that may occur in the underlying transmission management system in real-time.
	
	\item[3.] We propose both mathematical rule-based and deep-learning-based approaches for detecting anomalies in the IoT transmission frequency management system. In particular, the rule-based approach is designed to reveal anomalies in the system based on fundamental optimisation theory, and the deep-learning approach aims to establish a prediction model based on sequential data analysis in system implementations.
	
	\item[4.] We conduct a comprehensive comparative study using both anomaly detector strategies and demonstrate the strengths and weaknesses of the two approaches in both simulated and practical working environments. 
\end{itemize} 	



The remainder of this chapter is organised as follows. In section \ref{sec:System_model}, the architecture of the proposed system is presented. The optimisation problem is formulated in section \ref{probstatement} and its implementation is discussed in section \ref{implementation}. The experiments of transmission frequency management and results are discussed in section \ref{allocation}. The anomaly detection mechanisms are demonstrated in section \ref{anomalydefine}. The real-world experiment for anomaly detection is presented in section \ref{setup} and the corresponding results are discussed in section \ref{results}. Finally, a conclusion for this chapter is provided in section \ref{sec:Conclusion}.

\section{System Architecture}\label{sec:System_model}

\begin{figure}[ht]
	\vspace{-0.1in}
	\centering
	\includegraphics[width=1.0\textwidth, height=0.8\textwidth]{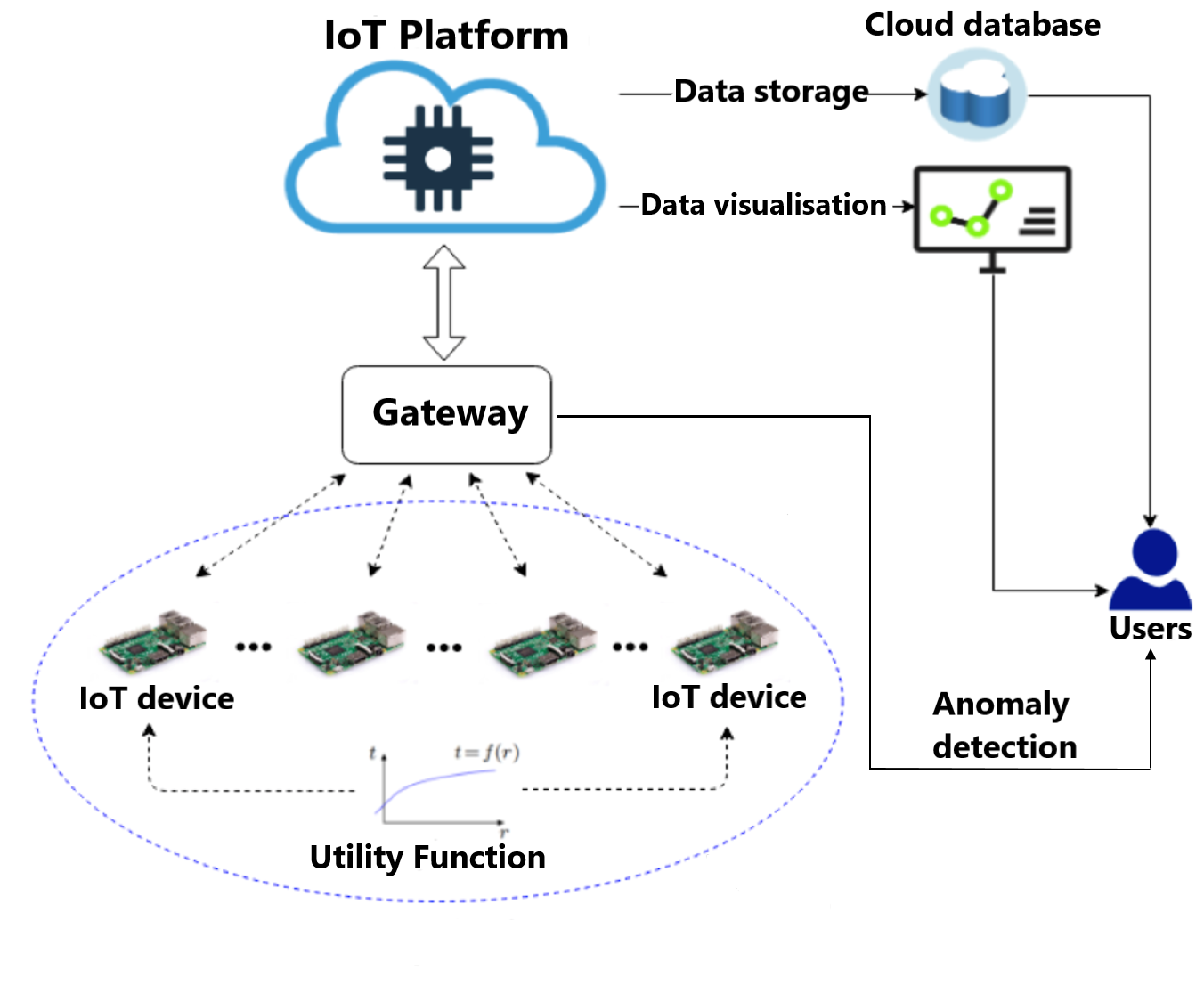}
	\caption{Schematic diagram of the system architecture.}
	\label{fig:architecture1}
	\vspace{-0.1in}
\end{figure}

Our proposed system architecture is illustrated in Fig. \ref{fig:architecture1}. The system consists of four main components, including IoT edge devices, gateways, a cloud platform and users. The main functionalities of each component are described as follows: 

\begin{itemize}
	\item[1.] \textbf{IoT devices}: sensors/devices connected to a gateway, having the capabilities of defining utility functions and the ability to solve a local optimisation problem in a decentralised manner.
	\item[2.] \textbf{Gateway}: collects data from IoT devices/sensors, passes data to the Cloud, and conducts basic data processing tasks including anomaly detection to protect and inform users. 
	\item[3.] \textbf{Cloud platform}: a central hub for data analysis, monitoring and storage. 
	\item[4.] \textbf{Users}: the owner of the IoT devices who wishes to use the IoT devices in some collaborative application scenarios.
\end{itemize}

In the proposed system, a gateway starts by waiting for a connection from IoT devices. When an IoT device initially connects to the gateway, the decentralised optimisation algorithm is activated to calculate the optimal transmission frequencies for all connected devices whilst taking account of the resource constraints of the system. After that, the gateway starts to collect data streams from all IoT devices after the transmission frequencies are established.  Finally, data collected by the gateway is transmitted to the cloud platform for data storage and further analysis of the IoT devices if specifically requested by the users. 

\section{Problem Statement} \label{probstatement}

We now present the specific problem statement to be solved in this chapter. A user wishes to determine the optimal DFWF of every IoT edge device so that the overall utility of the whole group can be maximised, given $N$, the number of devices connected to the gateway, the utility $f_i(x_i)$ of the $i^{th}$ device with current DFWF $x_i$, MWF $c$, total data storage (e.g., in unit MB) available per received data packet, $d$, and $a_i$ the data size (e.g., in unit MB) required for the $i$'th device per writing request.


Mathematically, this problem can be formulated as follows:

\begin{equation} \label{eq}
	\begin{gathered}
		\underset{x_1, x_2, \ldots, x_{N}}{\max} \quad
		\sum\limits_{i = 1}^{N} f_{i}\left(x_i \right),\\
		{\text{such that}} ~
		\sum\limits_{i = 1}^{N} x_i \leq c,  ~ \sum\limits_{i = 1}^{N} a_i x_i \leq d, ~ x_i \geq 0  \\
	\end{gathered}
\end{equation}

We shall only require that each utility function $f_i(x_i)$ can be modelled as a continuously differentiable, non-decreasing, strictly concave function, which is a common assumption for modelling the utility of internet data traffic \cite{srikant2004mathematics}. For example, utility functions may be modelled as a cluster of negative quadratic functions.

\section{System Implementation} \label{implementation}

The classic ADMM algorithm proposed in \cite{boyd2011distributed} is particularly suited to solving the formulated optimisation problem \eqref{eq} as the problem can be converted to a convex optimisation problem with convex constraints. Here we briefly recall the ADMM algorithm for solving \eqref{eq}, which is shown in Algorithm \ref{ADMM}, where $x$ and $z$ are
updated in an alternating fashion and $u$ is a dual update variable.

\begin{algorithm}[htbp]
	\caption{ADMM Algorithm}
	\begin{algorithmic}[1]
		\State  $\textbf{x}^{k+1}:=  \argmaxB_\textbf{x}  ( \sum\limits_{i = 1}^{N} f_i(x_i) + (\rho/2)  || \textbf{x} - \textbf{z}^{k} + \textbf{u}^{k} ||_{2}^{2})$
		\State  $\textbf{z}^{k+1}:= \Pi_{\mathcal{C}} (\textbf{x}^{k+1} + \textbf{u}^{k})$
		\State  $\textbf{u}^{k+1}:= \textbf{u}^{k} + \textbf{x}^{k+1} - \textbf{z}^{k+1}$
	\end{algorithmic}
	\label{ADMM}
\end{algorithm}

Note that the above ADMM algorithm can be implemented in a decentralised manner as our objective function is separable which implies that both $\textbf{x}$ and $\textbf{u}$ vector updates in the algorithm can be implemented in parallel. Finally, the $\textbf{z}$ update depends on inputs from both $\textbf{x}$ and $\textbf{u}$. Given these inputs, the projection operator $\Pi_{\mathcal{C}}$ projects the resulting vector to the constrained convex space $\mathcal{C}$. Thus, the $\textbf{z}$ update needs to be implemented on gateway. Note that $\rho$ is the augmented Lagrangian parameter and we take $\rho = 1.0$, being equivalent to a $\rho/2$ step size in $x$ update. The ADMM algorithm in its decentralised format is shown in Algorithm \ref{DecentralisedADMM}.

\begin{algorithm}[htbp]
	\caption{Decentralised ADMM Algorithm}
	\begin{algorithmic}[1]
		\State  ${x_i}^{k+1}:=  \argmaxB_{x_i}  (  f_i(x_i) + (\rho/2)  || {x_i}^{k} - {z_i}^{k} + {u_i}^{k} ||_{2}^{2})$
		\State  $\textbf{z}^{k+1}:= \Pi_{\mathcal{C}} (\textbf{x}^{k+1} + \textbf{u}^{k})$
		\State  ${u_i}^{k+1}:= {u_i}^{k} + {x_i}^{k+1} - {z_i}^{k+1}$
	\end{algorithmic}
	\label{DecentralisedADMM}
\end{algorithm}

\begin{figure}[ht]
	\vspace{-0.1in}
	\centering
	\includegraphics[width=0.8\textwidth]{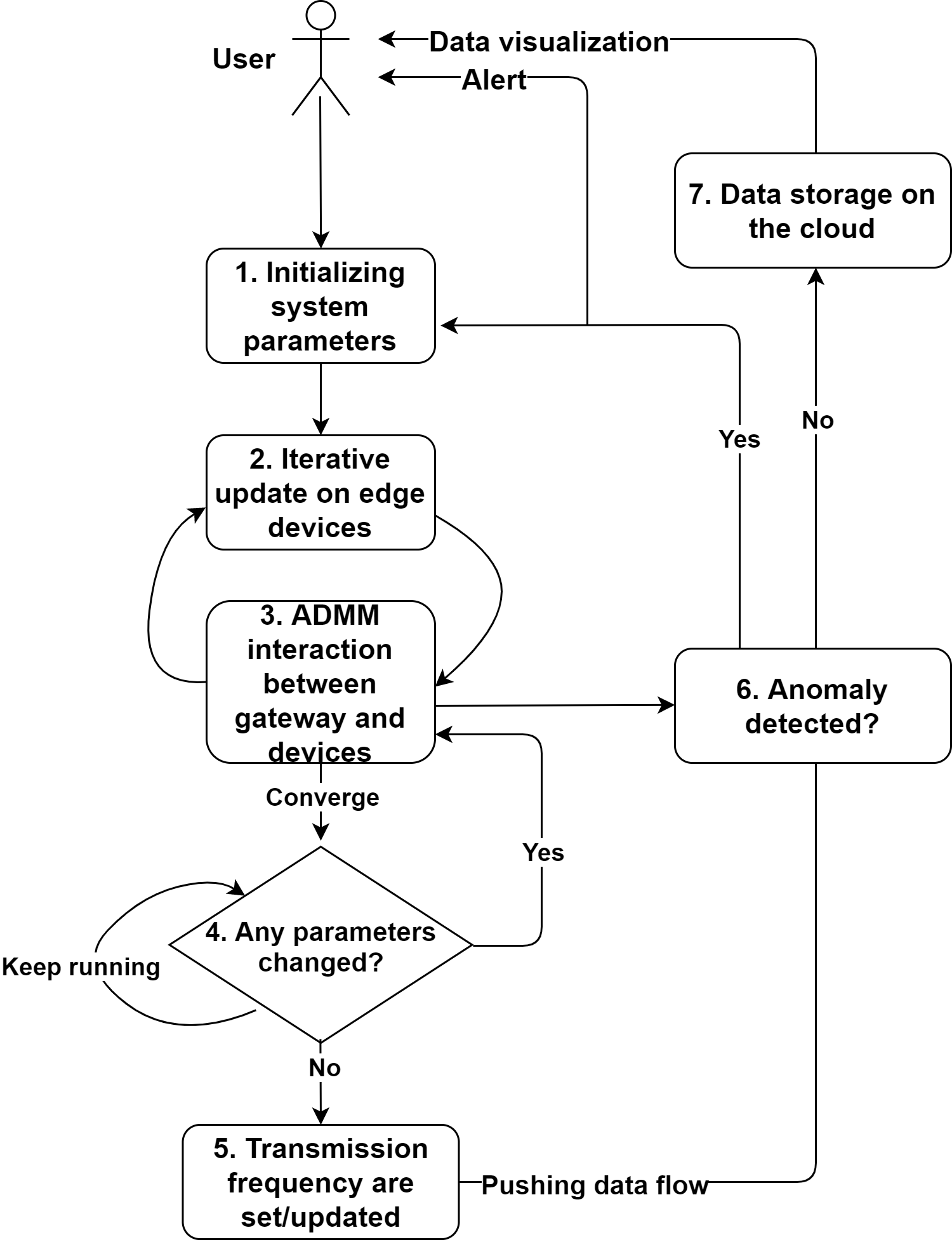}
	\caption{System implementation flowchart. }
	\label{fig:flow_chart}
\end{figure}

With this algorithm in mind, the proposed system can be implemented in the following steps, which are illustrated in Fig. \ref{fig:flow_chart}.

\begin{itemize}
	\item[S1:] During the initialisation stage, a user needs to specify some parameters before running the
	algorithm. This includes $N$, $c$, $d$, $a_i$ and the utility function $f_i(x_i)$ of each device.
	\item[S2:] When the initialisation step finishes, the ADMM algorithm will be implemented in an iterative manner on the edge IoT devices to determine the optimal DFWF by computing the optimal ${x_i}^{k+1}$ as per Algorithm \ref{DecentralisedADMM}.
	\item[S3:] During each iteration, the gateway gathers all the optimal ${x_i}^{k+1}$ from all devices, calculates and  broadcasts the updated $\textbf{z}$ value to local edge devices. Upon receiving the $\textbf{z}$ value, each edge device updates ${u_i}^{k+1}$ correspondingly.
	\item [S4:] If there are any resource changes during  runtime, the algorithm can dynamically capture the changes to recalculate the optimal solution given the new context. 
	\item[S5:] When the algorithm converges, the optimal DFWF will be set by each device, and these devices
	can then start pushing data to the cloud accordingly. 
	\item[S6:] The gateway keeps monitoring the data injection and detects if an anomaly happens on any of the transmission frequencies. If so, the user will be alerted and the optimal solution will be recalculated and reset after the anomaly has been remedied. We note that the legitimate reconfigurations of the system should not be identified as anomalies. Instead, the devices notify the gateway when legitimate changes happen, and the system executes step S4.
	\item[S7:] Finally, all transmitted data streams will be stored on the cloud and an authorised user can leverage the stored data for visualisation and analysis by making a request. 
	
\end{itemize}

\begin{figure}[ht]
	\vspace{0.1in}
	\centering
	\includegraphics[width=0.6\textwidth]{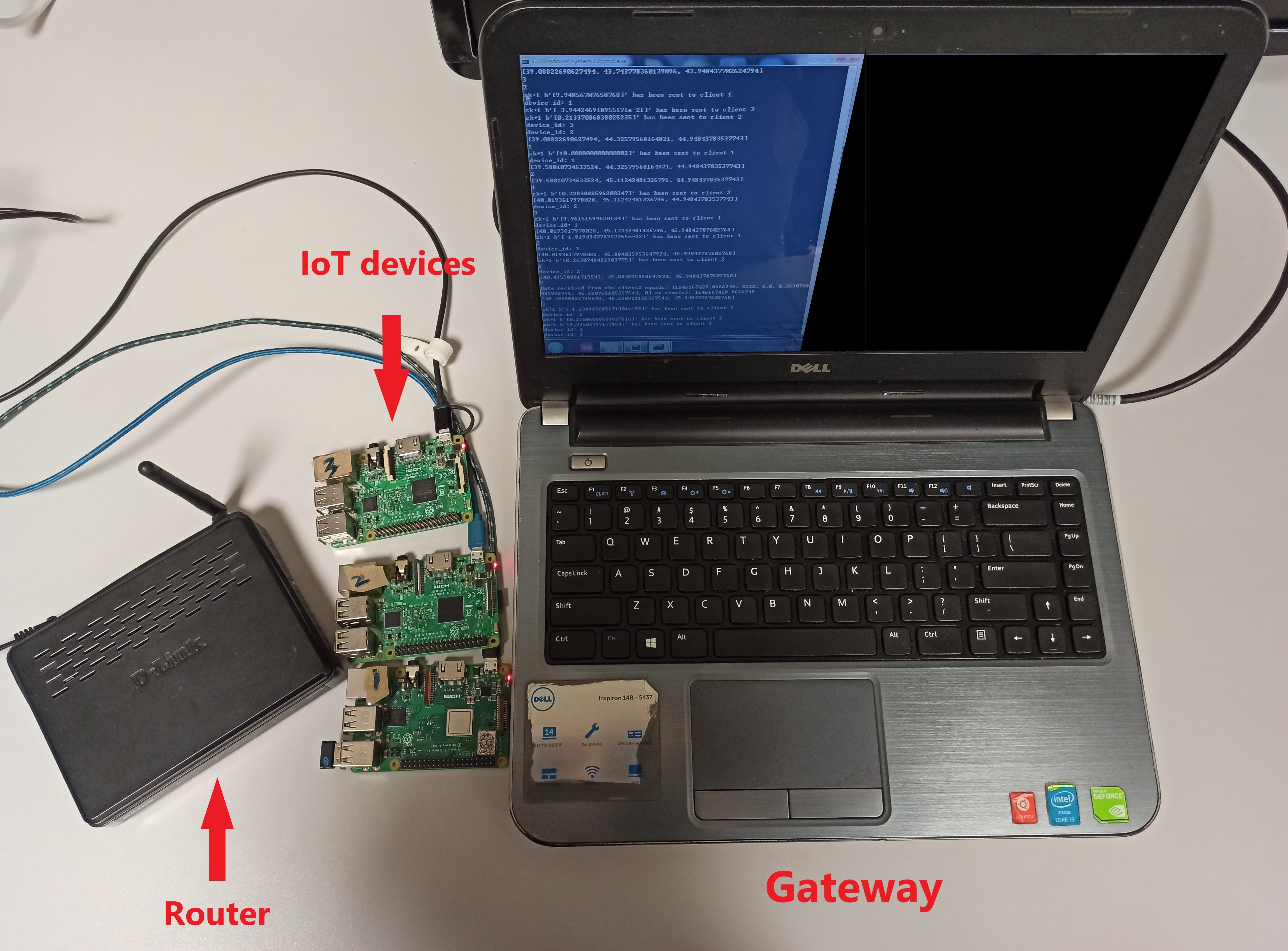}
	\caption{System simulation device setup.}
	\label{fig:devices}
\end{figure}

\section{Experiment results on optimal transmission frequency allocation} \label{allocation}

This section presents simulation results to evaluate the performance of the proposed system. As shown in Fig. \ref{fig:devices}, the system consists of a laptop as the central node (i.e., as a smart gateway in this work), three IoT devices (Raspberry Pi), and a router for the communication between the gateway and the IoT devices. Typically, IoT devices connect to the router in a wireless manner. However, in our setup, since the IoT devices do not have the capability of wireless transmission, they transmit data to the router via cables, and the laptop communicates with router wirelessly. Decentralised ADMM optimisation and data transmission are implemented on both the gateway and devices via socket programming. System parameters for the simulations are set as $N = 3$, $c = 10$, $d = 15$, $a_1 = 2$, $a_2 = 3$, and $a_3 = 5$. The utility functions in this simulation are presented in Table \ref{UtilityF} and have the characteristics previously specified to successfully apply the ADMM algorithm. We note that the utility functions are required to be concave based on optimisation problem \ref{eq} and the utility functions in Table \ref{UtilityF} are selected as our examples.  We simulate the system in two scenarios: a) resources are sufficient for the data transmission request, and b) resources are insufficient for the data transmission request from all devices. For each device $i$, its transmission frequency $x_i$ is defined as data is transmitted $x_i$ times per second. In particular, $x_i = 0$ implies that the $i^{th}$ device is not transmitting data. Thus, for each device, an extra constraint, $x_i >= \gamma_i$ applies to indicate the minimum transmission frequency. For simplicity, we set $\gamma_1 = \gamma_2 = \gamma_3 = 1$ in our simulation.

It is worth noting that the gateway is not able to access the utility function of each device in order to cater for privacy concerns, and also that the transmission frequency of each device is calculated locally and not explicitly exposed to the gateway. However, a DFWF may be estimated by the gateway by evaluating the time intervals of the consecutively received data packets and an averaged DFWF is calculated over 300 data packets after the optimal DFWF is assigned.


\renewcommand\arraystretch{1.5}
\begin{table}[ht!]
	\centering
	\caption{Utility Functions for transmission frequency allocation experiment}\label{UtilityF}
	\vspace{-0.2cm}
	\begin{tabular}{|c|c|}
		\hline
		\textbf{Device index} &
		\textbf{Utility Functions} \\ \hline
		1 &  $f_1(x_1) = -(x_1+9)^2 - x_1^3 + 900$ \\ \hline
		2 & $f_2(x_2) = -(x_2-4)^2 + 500$ \\ \hline
		3 & $f_3(x_3) = -(2x_3+3)^2 - x_3^3 + 110$ \\ \hline
		
	\end{tabular}
	
\end{table}

\subsection{Allocation with sufficient resources}

\begin{figure}[ht]
	\vspace{-0.1in}
	\centering
	\includegraphics[width=0.8\textwidth]{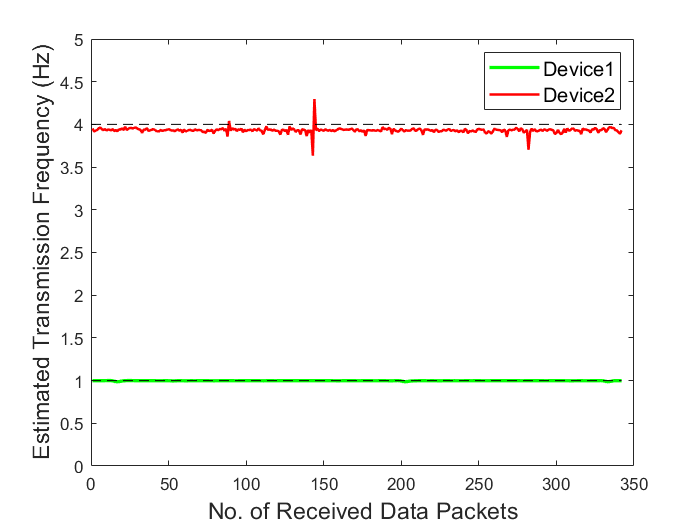}
	\caption{Decentralised optimisation process of transmission frequencies for Device 1 and Device 2.}
	\label{fig2a}
	\vspace{-0.1in}
\end{figure}

\begin{figure}[ht]
	\vspace{-0.1in}
	\centering
	\includegraphics[width=0.8\textwidth]{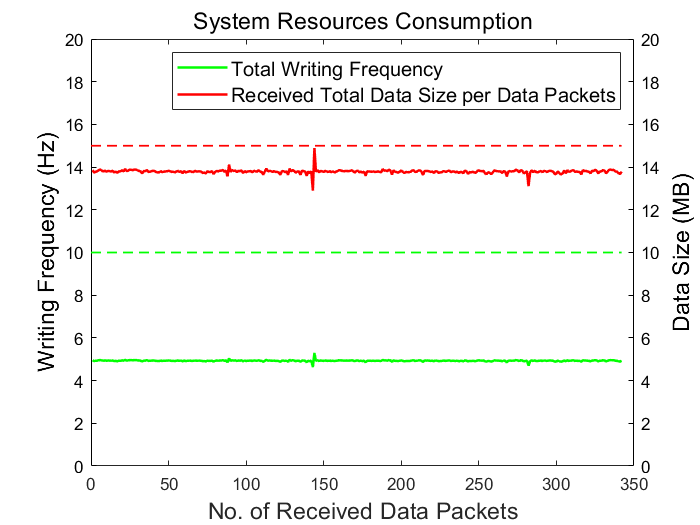}
	\caption{System resources consumption.}
	\label{fig2b}
	\vspace{-0.1in}
\end{figure}

In this scenario, only device $1$ and device $2$ are connected to the gateway (i.e., parameter $N = 2$) and all other system parameters are kept as $c = 10$, $d = 15$, $a_1 = 2$, $a_2 = 3$ with the associated utility functions $f_1(x_1)$ and $f_2(x_2)$ shown in Table \ref{UtilityF}. With these parameters, the theoretical optimal results of the ADMM implementation are $x_1^{*} = 1$ and $x_2^{*} = 4$ for the optimisation problem \ref{eq}. This result implies that the gateway expects to receive $1$ and $4$ data packet(s) per second from device $1$ and $2$ on average.  In this setup, the capacity provided by the system is sufficient since $x_1^{*} + x_2^{*} < c$ and $a_1 * x_1^{*} + a_2 * x_2^{*} <d$. With the decentralised ADMM implemented using the simulation setup, the optimisation results and resource consumption of the system are illustrated in Fig. \ref{fig2a} and Fig. \ref{fig2b}, respectively. In particular, Fig. \ref{fig2a} shows the evolution of the calculated DFWF for both devices as estimated by the gateway. The DFWFs are estimated along with the number of received data packets, indicated by the red and green lines for device $1$ and device $2$, respectively. Concretely, our results show that the estimated DFWFs are $0.9984\ Hz$ and $3.9318\ Hz$ for device $1$ and $2$, respectively, as shown in Table \ref{simulationA}, which result in a $1.6\ ms$ delay for device $1$ (i.e., calculated by $ \frac{1}{Actual DFWF} - \frac{1}{Theoretical DFWF}$) and a $4.3\ ms$ delay for device $2$. The estimated DFWFs are just slightly below the theoretical optimal DFWFs, indicated by the dotted line in Fig. \ref{fig2a}. The decrease of the DFWF may be accounted for by the internet speed, while the communication between the gateway and the devices is based on a router. Meanwhile, we find that the fluctuation of the estimated DFWFs is caused by the data jamming when the gateway is receiving data packets from IoT devices with high writing frequency. Fig. \ref{fig2b} shows the sum of DFWFs as well as the size of total data packets of all connected devices per second transmitted to the gateway. The dotted line indicates the maximum total DFWF (in red) and received data size (in green) for each data packet. Since the system can provide sufficient resources, the total DFWF and the writing data size have not reached the resource boundary after the transmission frequencies are optimised, indicating that the proposed system is robust as long as the system resources are sufficient for this specific data transmission task.

\begin{table}[!h]
	\centering
	\caption{Simulation results (average) with sufficient system resource}\label{simulationA}
	\vspace{-0.2cm}
	\begin{tabular}{|c|c|c|}
		\hline
		\textbf{ } & \textbf{DFWF (Hz)} & \textbf{DFWF (Hz)}  \\ \hline
		\textbf{ } & \textbf{Device 1} & \textbf{Device 2} \\ \hline
		\textbf{Theoretical} & 1.0000 & 4.0000   \\ \hline
		\textbf{Actual} & 0.9984 & 3.9318   \\ \hline
		\textbf{Absolute Error} & 0.0016 & 0.0682   \\ \hline
	\end{tabular}
\end{table}

\subsection{Allocation with insufficient resources}

In this scenario, after device $1$ and device $2$ have connected to the gateway and the optimised transmission frequencies have been calculated, a new device, device $3$, connects to the gateway and the timing of connection is recorded. Given $N = 3$, $c = 10$, $d = 15$, $a_1 = 2$, $a_2 = 3$, $a_3 = 5$ and the corresponding utility functions $f_1(x_1)$, $f_2(x_2)$, $f_3(x_3)$ reported in Table \ref{UtilityF}, the theoretical optimal results of the ADMM implementation are calculated as  $x_1^{*} = 1.00$, $x_2^{*} = 2.66$ and $x_3^{*} = 1.00$ for optimisation problem \ref{eq}. This result implies that, on average, the gateway expects to receive $1$, $2.66$ and $1$ data packet(s) per second from devices $1$, $2$, and $3$ respectively.

\begin{figure}[ht]
	\vspace{-0.1in}
	\centering
	\includegraphics[width=0.8\textwidth]{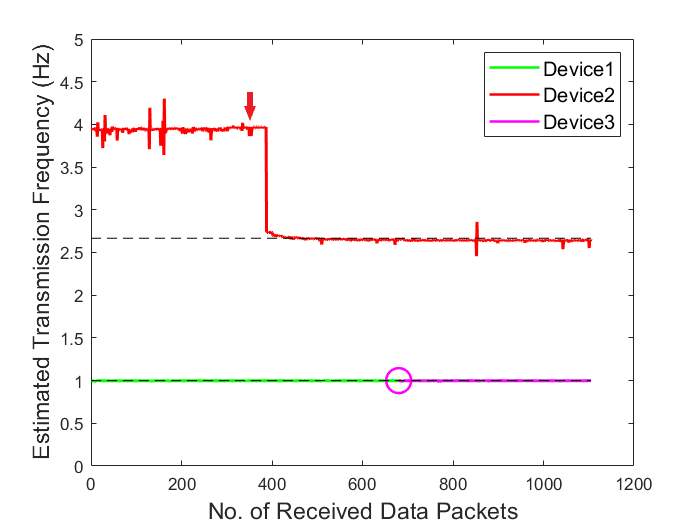}
	\caption{Decentralised optimisation process of transmission frequencies for Device 1, Device 2 and Device 3.}
	\label{fig4}
	\vspace{-0.1in}
\end{figure}
\begin{figure}[ht]
	\vspace{-0.1in}
	\centering
	\includegraphics[width=0.8\textwidth]{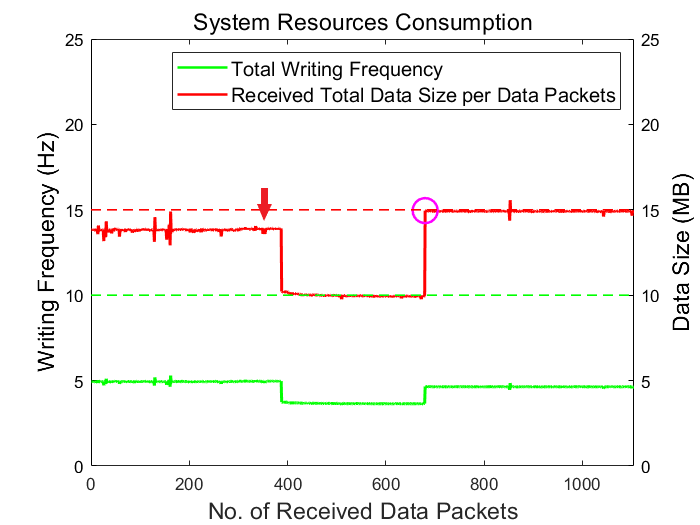}
	\caption{System resource consumption in boundary conditions. The magenta cycle shows the connection point of device 3.}
	\label{fig5}
	\vspace{-0.1in}
\end{figure}

Based on the simulation platform, the decentralised optimisation process and system resource usage are shown in Fig. \ref{fig4} and Fig. \ref{fig5} in the scenario of insufficient resources. We note that before the connection of device $3$, device $1$ and device $2$ transmit their data packets under the corresponding optimised transmission frequencies exactly as described in the first scenario with sufficient resources. As shown in Fig. \ref{fig4}, after the device $3$ connects to the system (indicated by the red arrow), the DFWF of device $2$ is readjusted and converges to a new optimal value. The DFWF of device $1$ remains unchanged since the recalculated optimal result equals the previously assigned DFWF before the connection of device $3$. After the decentralised ADMM solution is found for device $3$ (indicated by the magenta circle), device $3$ pushes data packets to the gateway using its optimal DFWF. After all three devices are transmitting data steadily (i.e., after the magenta circle), our results show that the estimated DFWFs are $0.9984\ Hz$, $2.6410\ Hz$ and $0.9984\ Hz$ for device $1$, $2$, and $3$, respectively, which are reported in Table \ref{simulationB}. Again, these estimated DFWFs are slightly below the theoretical optimal DFWFs, indicated by dotted lines, reflecting time delays of $1.6\ ms$, $3.7\ ms$ and $1.6\ ms$ (i.e., calculated by $ \frac{1}{Actual DFWF} - \frac{1}{Theoretical DFWF}$) for devices $1$, $2$, $3$, respectively during their transmissions.

After the optimal transmission frequencies are established, as shown in Fig. \ref{fig5}, device $3$ starts to push data (marked by the magenta circle) and the total writing data size reaches the level of the system resource boundary immediately. This indicates that the proposed system is able to reallocate the system resources to finish the data transmission task effectively using the ADMM approach. Finally, for comparison purposes, we evaluate the overall utility under the ADMM-optimised DFWFs, with non-optimised average distributed DFWFs (i.e., $x_i = c/N$), and non-optimised proportionally distributed DFWFs (i.e., $x_i = (a_i*c)/\sum a_i$) as two baselines given the same MWF $c$. The results shown in Table \ref{UtilityValue} find that the utility under ADMM-optimised DFWFs achieves the largest value, which demonstrates that the proposed system obtains the best result compared to other trivial system setups that have not undergone any optimisation process. 

\begin{table}[!h]
	\centering
	\caption{Simulation results (average) with insufficient system resource}\label{simulationB}
	\vspace{-0.2cm}
	\begin{tabular}{|c|c|c|c|}
		\hline
		\textbf{ } & \textbf{DFWF (Hz)} & \textbf{DFWF (Hz)} & \textbf{DFWF (Hz)}  \\ \hline
		\textbf{ } & \textbf{Device 1} & \textbf{Device 2} & \textbf{Device 3}  \\ \hline
		\textbf{Theoretical} & 1.0000 & 2.6667 & 1.0000  \\ \hline
		\textbf{Actual} & 0.9984 & 2.6410 & 0.9984  \\ \hline
		\textbf{Absolute Error} & 0.0016 & 0.0257 & 0.0016 \\ \hline
	\end{tabular}
\end{table}

\begin{table}[ht!]
	\centering
	\caption{Utility Evaluation}\label{UtilityValue}
	\vspace{-0.2cm}
	\begin{tabular}{|c|c|}
		\hline
		\textbf{DFWFs} &
		\textbf{Utility Value} \\ \hline
		ADMM optimised &  1381.22 \\ \hline
		Average distributed & 1190.35 \\ \hline
		Proportionably distributed & 1086.00 \\ \hline	
	\end{tabular}
\end{table}

\begin{figure}[ht]
	\vspace{-0.1in}
	\centering
	\includegraphics[width=0.8\textwidth]{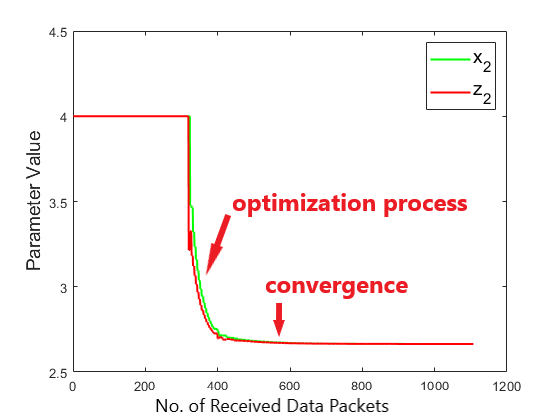}
	\caption{Monitoring of $z_{2}$ and $x_{2}$ during the decentralised optimisation process for Device 2.}
	\label{fig6}
\end{figure}

\section{Anomaly detection for changes of transmission frequency } \label{anomalydefine}

While the transmission frequencies are determined and allocated by the system, all the devices push data steadily with their specified DFWF. However, the transmission frequencies can be tampered with both explicitly and implicitly. In other words, a malicious attack to the device can not only manipulate the DFWF explicitly, but also can modify the utility function (i.e., the input or function type), the system transmission data size and the system resource, which leads to a change of DFWF implicitly. In this section, the above manipulations are discussed for the examination of abnormal transmission frequency detection at the gateway side.

We first consider the scenario of manipulating the DFWF explicitly. According to the fundamental mechanism of the ADMM algorithm, the gateway only has access to $\textbf{z}$. Since $\textbf{x}$ achieves convergence to $\textbf{z}$ eventually, as a specific example (i.e., $z_{2}$ and $x_{2}$) shown in Fig. \ref{fig6}, we argue that the gateway is able to detect the anomaly of $\textbf{x}$ during the whole transmission process based on its knowledge of the latest value of $\textbf{z}$. Specifically, this detection process can be described in the following three steps: 

\begin{itemize}
	\item[S1:] Gateway accesses the value of $z_{i}$ for each device.
	\item[S2:] Gateway estimates the DFWF (i.e., the converged value of $x_{i}$) for each device according to the received time-stamped data flow.
	\item[S3:] If the estimated DFWF is significantly different to the reference value of $z_{i}$ (i.e., $|z_{i}-x_{i}| \geq \delta$, where $\delta$ is a threshold depending on the network delay), the optimal transmission frequency can regard as anomalous and as being manipulated.
	
\end{itemize}

However, the above detection process is not able to apply in some scenarios. Given the transmission frequency management system described in Fig. \ref{fig:architecture1} and problem \eqref{eq}, there are other types of manipulations on the edge (i.e. including edge devices and gateway) that can also lead to the changes in transmission frequencies. Specifically, these manipulations can happen by changing the utility function input, function type, data size requested per writing request (i.e. defined on edge devices), maximum writing frequency and data storage (i.e. system resource allocated to the gateway), leading to a new ADMM optimisation process with $\textbf{x}$ value converging on $\textbf{z}$ value. In general, when manipulations happen on the device $j$ in the network, a new optimisation process needs to be reactivated by solving the following problem: 

\begin{equation} \label{eqAnomaly}
	\begin{gathered}
		\underset{x_1, x_2, \ldots, x_{N}}{\max} \quad
		\sum\limits_{i = 1, i \neq j}^{N} h_{i}\left(x_i \right) + h_{j}^*\left(x_j \right),\\
		{\text{subject to}} ~
		\sum\limits_{i = 1}^{N} x_i \leq c, \sum\limits_{i = 1, i \neq j}^{N} a_i x_i  + a_j^* x_j  \leq d, ~ x_i \geq 0  \\
	\end{gathered}
\end{equation}
where $h_{j}^*$ and $a_j^*$ denote the new utility function and new data packet size after tampering, respectively. 



Clearly, there are many ways that an optimal transmission frequency $x_j$ can be implicitly tampered. In our context, we consider the following specific definitions:

\begin{itemize}
	\item[1.] \textbf{Manipulation on utility function input only}: The independent variable of the utility function is manipulated by adding an input factor with a small given range, $h_j(x_j) \Rightarrow h_j(x_j + \textit{input factor} ) $. 
	\item[2.] \textbf{Manipulation on utility function type and input}: The utility function can be totally changed to anther type of concave function specified by the utility function set of the system, i.e., $h_j(x_j) \Rightarrow h_j^*(x_j + \textit{input factor} )$.
	\item[3.] \textbf{Manipulation on transmission data size}: The data size $a_j$ required for the $j$’th device per writing request is manipulated by adding a size factor with a small given range, $a_j \Rightarrow a_j + \textit{size factor}$.
\end{itemize}

\noindent\textbf{Comment:} It is also possible to affect the optimal transmission frequency $x_j$ and $x_i, i \neq j$ by manipulating system resource in a small given range, such as $c \Rightarrow c + \textit{MWF factor}$ and $d \Rightarrow d + \textit{storage factor}$. In our definition, such manipulations are regarded as a systematic adjustment as it is not directly related to any user-specific property, e.g., $h_j$, and thus it will be regarded as normal scenarios in our anomaly detection analysis.


In addition, we also have the following assumptions in our problem \ref{eqAnomaly}. 

\begin{figure}[ht]
	\vspace{-0.1in}
	\centering
	\includegraphics[width=0.6\textwidth]{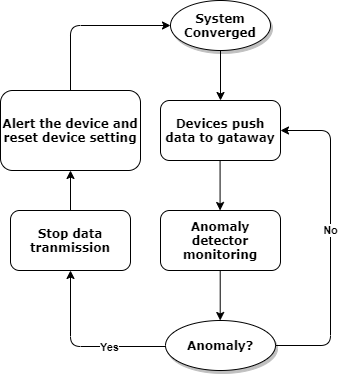}
	\caption{Anomaly detection and response process. }
	\label{fig:detectionProcess}
\end{figure}

\begin{itemize}
	\item[1.] We assume that at every given time only one edge device is manipulated, which is the fundamental basis for detecting an anomaly when multiple devices are manipulated in our system.
	\item[2.] We assume that the anomaly detector is a separate process running on the gateway, and it can only access limited information on the gateway but not all. More specifically, we assume that the anomaly detector can only access the value of $\textbf{\textit{z}}$ and the sum of $\textbf{\textit{x}}$ and $\textbf{\textit{u}}$, denoted by $\textbf{\textit{v}}$, from the ADMM iterative process at the gateway. It will never access the exact transmission frequency $\textbf{\textit{x}}$ directly from the local devices and other resources/parameters shared between devices and the gateway. 
	\item[3.] We assume that the anomaly detector starts to monitor anomalies in real-time once the ADMM algorithm converges and local devices start pushing data to the gateway. The device setting will be reset when any anomalies are detected, and the optimisation process will be reactivated to reset the optimal solutions for fair resource allocation as per the normal situation. To further illustrate this point, the process of anomaly detection is shown in Fig. \ref{fig:detectionProcess}. 
\end{itemize}

We now introduce two approaches to address the anomaly detection problem, namely a rule-based approach and a deep learning approach. The rule-based approach detects system anomalies based on the mathematical deduction, and the deep learning approach solves the detection problem using collected experimental datasets of the system. The rule-based approach is proposed as a baseline method as we shall see it has some drawbacks in detecting system anomalies in detail.

\subsection{Rule-based anomaly detection} \label{detect_math}

Our objective is to investigate the behaviour of the optimised system before and after manipulation. To this end, we borrow some fundamental concepts from the optimisation theory, i.e. the Karush-Kuhn-Tucker (KKT) conditions \cite{gordon2012karush} for the optimisation problem \eqref{eq} under study. For mathematical conventions, we now rewrite the original optimisation problem \eqref{eq} in the following format:

\begin{equation} \label{eq2}
	\begin{gathered}
		\underset{x_1, x_2, \dots, x_N} {\min} \quad
		\sum\limits_{i = 1}^{N} f_i(x_i) ,\\
		{\text{s.t.}} ~
		\sum\limits_{i = 1}^{N} x_i \leq c,  ~ \sum\limits_{i = 1}^{N} a_i x_i \leq d, ~ x_i \geq 0  \\
	\end{gathered}
\end{equation}
where $f_{i}\left(x_i \right) := - h_{i}\left(x_i \right)$ is a convex function. The Lagrange equation of \eqref{eq2} is presented as follows:

\begin{equation} \label{eqL}
	\begin{gathered}
		\textbf{L}(\textbf{\textit{x}}, \lambda _1, \lambda _2) = \sum\limits_{i = 1}^{N} f_i(x_i) + \lambda _1 g_1(\textbf{\textit{x}}) + \lambda _2 g_2(\textbf{\textit{x}})\\
	\end{gathered}
\end{equation}
and the KKT conditions require the following to be held for optimality: 

\begin{equation} \label{eq3}
	\begin{gathered}
		{\frac{\partial \textbf{L}}{\partial x_i} = {\frac{\partial  f_i (x_i)}{\partial x_i}
				+ \lambda _1 \frac{\partial  g_1 (\textbf{\textit{x}})}{{\partial x_i}}	
				+ \lambda _2 \frac{\partial  g_2 (\textbf{\textit{x}})}{{\partial x_i}}} = 0  },\\
		{\lambda _1 , \lambda _2 \geq 0}, \\ 
		\lambda _1  g_1(\textbf{\textit{x}}), \lambda _2  g_2(\textbf{\textit{x}}) = 0\\
	\end{gathered}
\end{equation}
where $\partial$ is the operation of partial derivative (i.e., gradient), $\lambda _1$, $\lambda _2$ are Lagrange coefficients for $g_1(\textbf{\textit{x}})$, $g_2(\textbf{\textit{x}})$ and $\textbf{\textit{x}} = (x_1, x_2, \cdots, x_N)$. 

Specifically, $g_1(\textbf{\textit{x}}) = \sum\limits_{i = 1}^{N} x_i - c$ and $g_2(\textbf{\textit{x}}) = \sum\limits_{i = 1}^{N} a_i x_i - d$, which represents the constraints in problem \eqref{eq2} with:

\begin{equation} \label{eq4}
	\begin{gathered}
		{\frac{\partial  g_1 (\textbf{\textit{x}})}{\partial x_i} = 1 },\\
		{ \frac{\partial  g_2 (\textbf{\textit{x}})}{\partial x_i} = a_i}. \\
	\end{gathered}
\end{equation}

Clearly, the converged optimal solution will fall into one of the following situations with reference to system constraints.\\

\subsubsection{Situation $g_1 (\textbf{x}) = 0$, $g_2 (\textbf{x}) < 0$} \label{constraint1}

Given $g_2 (\textbf{\textit{x}}) < 0$, we have $\lambda _2 = 0$ according to equation \eqref{eq3}. The system is running under $\sum\limits_{i = 1}^{N} x_i = c$. Thus, for each device $i$, we have 

\begin{equation} \label{eq5}
	\begin{gathered}
		{\frac{\partial \textbf{L}}{\partial x_i} = {\frac{\partial  f_i (x_i)}{\partial x_i}
				+ \lambda _1 \frac{\partial  g_1 (\textbf{\textit{x}})}{{\partial x_i}}} }\\
		{  = \frac{\partial  f_i (x_i)}{\partial x_i} + \lambda _1 = 0}, \\
	\end{gathered}
\end{equation}
That is 

\begin{equation} \label{eq6}
	\begin{gathered}
		{{\frac{\partial  f_1 (x_1)}{\partial x_1} = { \frac{\partial  f_2 (x_2)}{\partial x_2} = \cdots 
					= \frac{\partial  f_N (x_N)}{\partial x_N} }} } \\
	\end{gathered}
\end{equation}
for problem \eqref{eq2}.

Considering the constraint $\sum\limits_{i = 1}^{N} x_i = c$, when a manipulation results in an increase of DFWF for device $j$ (i.e. $x_j$), at least one of $x_i$ ($i \neq j $) decreases. Considering that $\frac{\partial  f_i (x_i)}{\partial x_i}$ is monotonously increasing with respect to an increased $x_i$ (i.e. with convexity), the decrease of $x_i$ will also decrease $\frac{\partial f_i (x_i)}{\partial x_i}$. Consequently, $\frac{\partial f_i (x_i)}{\partial x_i}, \forall i \neq j$, will decrease as per \eqref{eq6}, which indicates decrease of $x_i, \forall i \neq j$. Therefore, an increase of $x_j$ results in the decrease of transmission frequencies of all other devices $x_i, \forall i \neq j $. \\

\subsubsection{Situation $g_1 (\textbf{x}) < 0$, $g_2 (\textbf{x}) = 0$} 
Given $g_1 (\textbf{\textit{x}}) < 0$, we have $\lambda _1 = 0$ according to equation \eqref{eq3}. The system is running under $\sum\limits_{i = 1}^{N} a_i x_i = d$. For each $i$, we have 

\begin{equation} \label{eq7}
	\begin{gathered}
		{\frac{\partial \textbf{L}}{\partial x_i} = {\frac{\partial  f_i (x_i)}{\partial x_i}
				+ \lambda _2 \frac{\partial  g_2 (\textbf{\textit{x}})}{\partial x_i}} }\\
		{  = \frac{\partial  f_i (x_i)}{\partial x_i} + \lambda _2 a_i = 0}, \\
	\end{gathered}
\end{equation}
where $a_i \geq 0$ since $a_i$ is the required data size. This implies 
\begin{equation} \label{eq8}
	\begin{gathered}
		{{\frac{\partial  f_1 (x_1)}{\partial x_1} = {\frac{a_1}{a_2} \frac{\partial  f_2 (x_2)}{\partial x_2}
					= \cdots 
					= \frac{a_1}{a_N} \frac{\partial  f_N (x_N)}{\partial x_N} }} } \\
	\end{gathered}
\end{equation}
for problem \eqref{eq2}.

Similar to the first situation, without loss of generality, an increase of DFWF for device $j$, $x_j$, after a manipulation will lead to a decrease of at least one $x_i, i \neq j$ due to the equality constraint $g_2 (\textbf{\textit{x}}) = 0$. Since $ f_i (x_i)$ is convex, the decreases of $x_i$ indicates a decrease of $\frac{\partial f_i (x_i)}{\partial x_i}$. Given formula \eqref{eq8}, we have that $\frac{\partial f_i (x_i)}{\partial x_i}, \forall i \neq j$ decreases proportionally followed by the increase of $x_j$, resulting a reduced $x_i, \forall i \neq j$.\\

\subsubsection{Situation $g_1 (\textbf{x}) < 0$, $g_2 (\textbf{x}) < 0$} 
Given $g_1 (\textbf{\textit{x}}) < 0$ and $g_2 (\textbf{\textit{x}}) < 0$, we have $\lambda _1 = 0$ and $\lambda _2 = 0$ according to equation \eqref{eq3}. Thus, the system is running within the boundary of system resources. For each device $i$, we have

\begin{equation} \label{eq9}
	\begin{gathered}
		{\frac{\partial \textbf{L}}{\partial x_i} = {\frac{\partial  f_i (x_i)}{\partial x_i} } }, \\			 
	\end{gathered}
\end{equation}

Considering that the system is running within the boundary of system resources, manipulation on any device will not affect other devices. That is, for instance, when a manipulation results in an increase of DFWF for device $j$ (i.e. $x_j$), other $x_i, \forall i \neq j $, remain unchanged since they were already optimised and the system resource is sufficient to cover the extra needs for device $j$. 

\begin{figure}[ht]
	\vspace{-0.1in}
	\centering
	\includegraphics[width=0.6\textwidth]{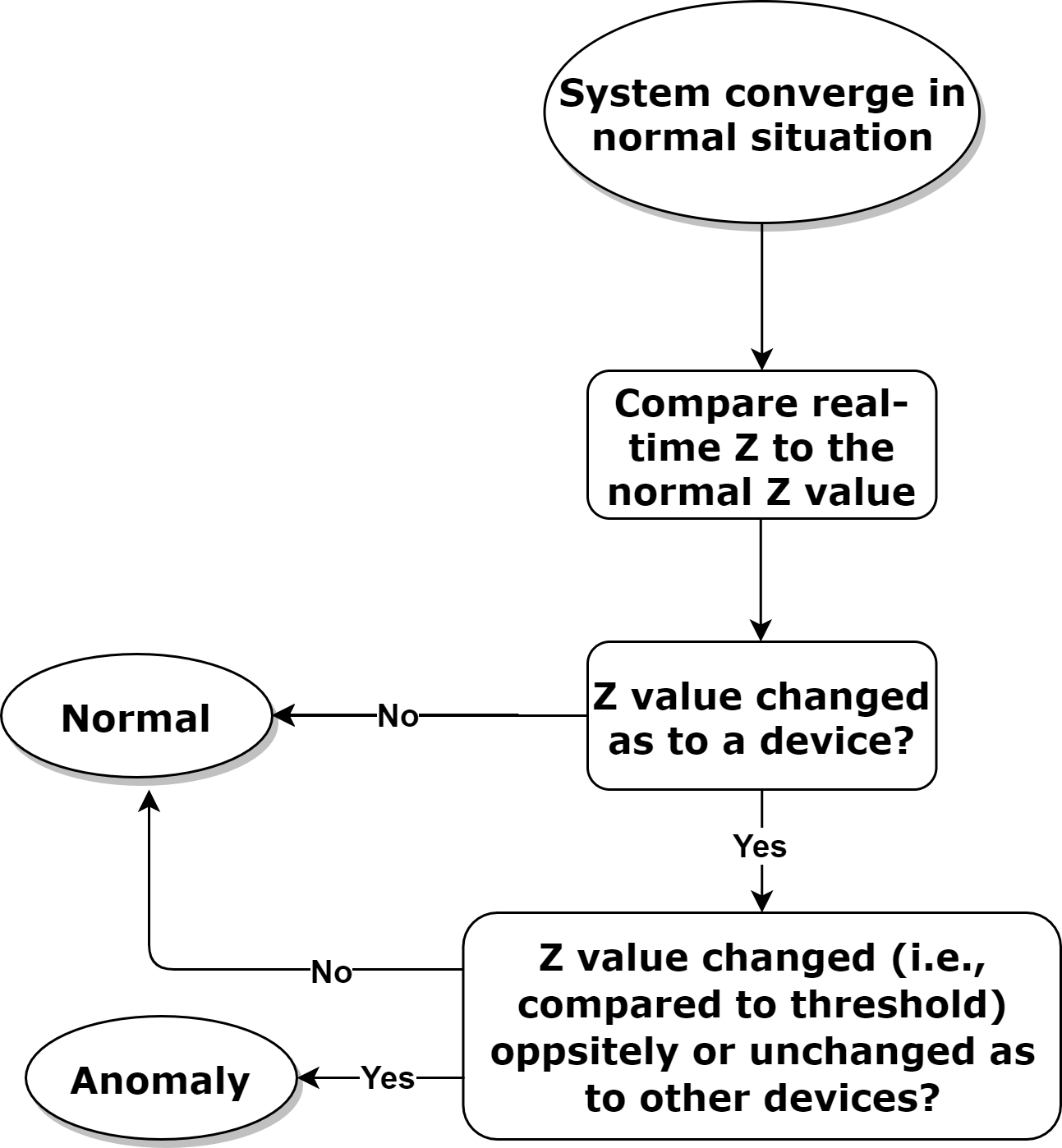}
	\caption{The proposed rule-based anomaly detection process.}
	\label{fig:rule}
\end{figure}

Given the above discussion, we have observed that once the manipulation accounts for a change of DFWF on a given device, DFWF of other devices will either change oppositely or remain unchanged. Accordingly, we can devise a simple rule-based mechanism for anomaly detection, and the flow chart is shown in Fig. \ref{fig:rule}. It operates as follows. When the system starts to operate and converges to optimality normally, the anomaly detector keeps a record of the normal $\textbf{\textit{z}}$ value while keeping monitoring the $\textbf{\textit{z}}$ value from the algorithm iteration in real-time. Once the absolute difference between the observed $\textbf{\textit{z}}$ value and the normal $\textbf{\textit{z}}$ value becomes greater than a preset threshold (component-wise), the anomaly for the corresponding device is recorded. In this work, the thresholds are defined as $1\%$, $5\%$, $10\%$, $15\%$, $30\%$, $50\%$ to the change of the recorded normal $\textbf{\textit{z}}$ value so that the performance of the approach can be evaluated comprehensively. 


\begin{figure}[ht]
	\vspace{-0.1in}
	\centering
	\includegraphics[width=0.7\textwidth]{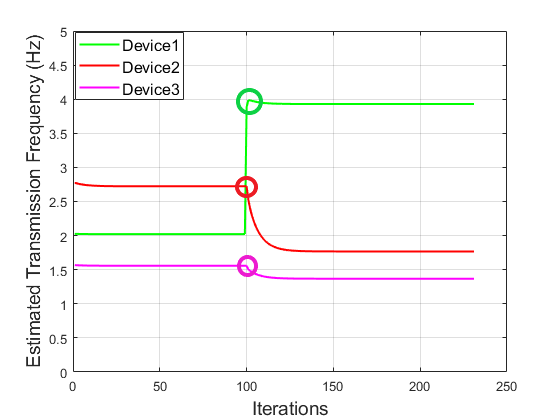}
	\caption{An example to illustrate the change of transmission pattern due to the manipulation of device 1.}
	\label{fig7}
\end{figure}

To further demonstrate how we can apply the rule-based approach for anomaly detection, a simple simulation is conducted on the IoT system consisting of three devices. The utility functions for all three devices are reported in Table \ref{simultionBFunctions}, where we assumed that the first device, i.e., device 1 was manipulated by only adding an $\textit{input factor} = -2$ at a given point during our experiment. Our results are shown in Fig. \ref{fig7}. It can be seen that device 1 was manipulated at the 100$^{th}$ iteration, indicated by different cycles highlighted in Fig. \ref{fig7}, leading to an increase by $94\%$ (i.e., from 2.02 to 3.93) in DFWF, while device 2 and device 3 reduced their transmission frequencies by $35\%$ and $12\%$ correspondingly. Therefore, by applying a threshold less than $12\%$ to the change of recorded normal $z$ values, the rule-based detector can detect the increase of transmission frequency in device 1 and the decrease of transmission frequencies in device 2 and device 3 successfully. Given this, an anomaly will be spotted in this case.



\begin{table}[ht!]
	\centering
	\caption{Utility Functions for anomaly detection experiment}\label{simultionBFunctions}
	\begin{tabular}{|c|c|}
		\hline
		\textbf{Index} &
		\textbf{Utility Functions} \\ \hline
		1 &  $f_1(x_1) = (x_1-9)^2 + x_1^3$ \\ \hline
		2 & $f_2(x_2) = (x_2-4)^2 $ \\ \hline
		3 & $f_3(x_3) = (2x_3-6)^2 + x_3^3 $ \\ \hline
		
	\end{tabular}
	
\end{table}

\subsection{Limitations of the Rule-based Anomaly Detection}

Our results in Section \ref{detect_math} show that a rule-based approach has potential for anomaly detection as long as the manipulation leads to a change of transmission frequency. However, such an approach also has certain limitations when deployed in the real world, which is summarised as follows:

\begin{itemize}
	\item[A.] The rule-based approach mainly relies on the optimality criteria without fully leveraging information from the iterative process, and as a result it cannot further distinguish different types of anomalies when a manipulation happens on the edge device. 
	\item[B.] As we shall see, system parameters, i.e., $\textbf{\textit{z}}$, may fluctuate during the optimisation process and that can easily result in misjudgements when using the rule-based approach.
	\item[C.] Furthermore, when there are network delays in the IoT network, transmission frequencies of the devices may not change simultaneously, which can also lead to misjudgements when using the rule-based approach.
\end{itemize}

Due to the uncertainty of a practical running IoT environment as well as the depth of information that can be leveraged from the collected data for anomaly detection, we are also interested in exploring a data-driven based solution to address the limitations exposed by the rule-based approach which is introduced in the following section. 

\subsection{IoT anomaly detection with LSTM-based approaches}

In this section, deep learning-based approaches are proposed for anomaly detection on the gateway, covering all categories of the anomalies defined in Section \ref{anomalydefine}. Our starting point is the observation that an anomaly detector can only access the value of $\textbf{\textit{z}}$ and the sum of $\textbf{\textit{x}}$ and $\textbf{\textit{u}}$, i.e., $\textbf{\textit{v}}$ at every given time point of interest, i.e., a sequential data. Inspired by this, we aim to leverage a prevalent sequence-based model, long short-term memory (LSTM) \cite{schmidhuber1997long} in our work,which becomes one of the popular architectures in anomaly detection \cite{ergen2019unsupervised,lindemann2021survey} and can leverage the collected information during the optimisation process.

Specifically, we apply a basic one-layer LSTM architecture in our model design and compare the detection performance with different complicated variants which have been applied in anomaly detection, e.g., bidirectional LSTM (bi-LSTM) \cite{aljbali2020anomaly}, stacked-LSTM \cite{thill2019anomaly}, LSTM with attention mechanism \cite{xia2021new} (LSTM-attention) and LSTM with encoder techniques \cite{nguyen2021forecasting} (LSTM-encoder), considering that the extra deep learning architectures may improve the detection performance. Let $X_t$ denote the input feature at step t (i.e., the $t^{th}$ iteration of the ADMM algorithm), then the LSTM network essentially extracts hidden information at each step, $t$, and feeds this in as the input of the next step, $t+1$. A standard LSTM unit includes a cell, a forget gate, an input gate and an output gate to jointly manage the information flow from input to output. The input feature $X_t$ can be either a scalar, vector or matrix. In our case, the input feature $X_t$ is represented as a matrix consisting of system parameters of each device, $i$, from iteration $t$ to $t+n$. Here, the input features contain $[v_i^t, \cdots, v_i^{t+n}]$, $[z_i^t, \cdots, z_i^{t+n}]$ where $v_i := x_i + u_i$. The output of the LSTM model is the categorical label for the anomaly corresponding to the manipulation types as per our definition.

\section{Experimental setup for anomaly detection} \label{setup} 

In this section, we first introduce several different types of manipulations, then we discuss the IoT system setup and data generation process. Finally, we present the LSTM network for anomaly detection. The IoT system is setup to transmit the data stream, under the circumstance where the transmission frequency may be manipulated implicitly. During the process of data stream transmission, the ADMM parameters which are able to reflect the system behaviours are recorded, to generate a dataset for LSTM model training for detecting the manipulations.

\subsection{Setup for manipulations} \label{manipulationsetup}

Utility functions defined on IoT devices may indicate user's preference in real-world IoT application. It is worth noting that how to define a user's preference using a utility function is an open issue \cite{5990666} as different users may end up having totally different utility values with respect to a given source, i.e., DFWF in our case. However, in our context, we shall make the assumption that such a function is concave as it generally reflects the fact that a user's satisfaction level is increased when the allocated DFWF is also increased. With this in mind, we have the following settings:

\begin{itemize}
	\item \textbf{Manipulation on utility function type and input}: The utility function is changed from $f_j(x_j)$ to $f_j^*(x_j)$ (i.e., see Table \ref{FuncSet}) with $\textit{input factor}$, resulting in manipulation $f_j(x_j) \Rightarrow f_j^*(x_j + \textit{input factor} ) $, labelled as type $1$.
	
	\begin{table}[ht!]
		\centering
		\caption{Utility Functions set}\label{FuncSet}
		\begin{tabular}{|c|}
			\hline
			\textbf{Utility Functions} \\ \hline
			$f_j(x_j) = (x_j-9)^2 + x_j^3$ \\ \hline
			$f_j^*(x_j) = \exp(x_j - 9) $ \\ \hline
			$f_j^*(x_j) = 1/(x_j - 9) $ \\ \hline
			$f_j^*(x_j) = \log (1 + \exp(x_j - 9))$ \\ \hline
		\end{tabular}
	\end{table}
	
	\item  \textbf{Manipulation on transmission data size}: The data size factor is set as a random value from the set of $[-1,1]$ and the $a_j$ is manipulated as $a_j \Rightarrow a_j + \textit{size factor}$, labelled as type $2$.
	
	\item \textbf{Manipulation on utility function input only}: In this case the $\textit{input factor}$ is set as a random value from the set of $[-3,3]$ for the manipulation $f_j(x_j) \Rightarrow f_j(x_j + \textit{input factor} )$, which is labelled as type $3$.
	
\end{itemize}

\noindent\textbf{Comment:} As mentioned in Section \ref{anomalydefine}, manipulating system resources can also affect the optimal transmission frequencies for edge devices, but it will be treated as a normal systematic adjustment. Regarding manipulation of system resources, the MWF, $c$, and data storage amount, $d$, are manipulated by adding an MWF factor and storage factor. The factors $c$ and $d$ are attributed a value from the set of $[-3,3]$ and $[-5,5]$ respectively, ensuring that the manipulated $c$ and $d$ are positive. Here we have manipulation $c \Rightarrow c + \textit{MWF factor}$ and $d \Rightarrow d + \textit{storage factor}$ which are labelled as normal (type $0$).

\subsection{System setup}

In general, we consider two different system setups in experiments. One simulates the ideal IoT scenario including an arbitrary number of devices, without considering the effects of network delay. The second simulates a practical IoT environment involving real IoT devices with network delay. In order to compare the performance of the two setups, we manually trigger manipulations and record the manipulation count/type for both systems. However, considering that in a real-world environment it is impossible for a gateway to know the ground truth, thus we deploy the pre-trained model based on the ideal scenario and evaluate the performance of the model in the practical IoT environment.

More specifically, the simulation system (SS) and real world system (RS) are introduced to validate the performance of the anomaly detector. The SS simulates the ideal scenario that all devices transmit data to the gateway without network delay. In our experiment, SS includes the virtual edge devices and gateway on a local computer where the data streams can be exchanged even if there is no network environment. The RS simulates the practical IoT application that all devices transmitting data to the gateway with network delay effect being considered. In our real-world implementation, this consists of three edge devices (i.e. Raspberry Pis) and a laptop acting as the gateway. Edge devices communicate with the gateway through a wireless router as shown in Fig. \ref{fig:devices}. The key system properties for this practical system are set as $N = 3, c = 10, d = 20, a_1 = 2, a_2 = 3,$ and $a_3 = 5$.  
	
It is worth noting that in SS we simulate the ideal scenario and generate data for the purpose of training the anomaly detector. Therefore the data is labelled corresponding to anomalies when devices are manipulated. In RS, we simulate the scenario that edge devices are implemented in a real-world IoT network for daily service. In this context, the data collected from RS is without labels and is used for anomaly detection in real-world applications. 

\subsection{Data generation}

The process of data generation can be summarised as follows: during the daily service of the IoT network, the system suffers attacks and transmits a data flow containing unexpected transmission frequencies, the system returns to its normal state after the end of the attack.
Specifically, at the beginning, SS and RS are running under the normal state. After the ADMM algorithm has converged for the duration of several ADMM iterations, a type of manipulation happens on the IoT devices and the system reacts, calculating new transmission frequency values. 
After the anomaly happens and the ADMM algorithm converges under the anomaly, the edge devices return to the normal state and the system repeats the process. The duration of normal states varies between 100 and 120 iterations, while the anomalies last for duration between 50 and 70 iterations.
During this cycle, the normal situation is labelled as type $0$ and anomalies are labelled as different numeric types.  Data (i.e. ADMM parameter) $\textbf{\textit{z}}$ and $\textbf{\textit{v}}$ generated from the ADMM algorithm are recorded along with each iteration during the interaction between the gateway and the edge devices. Data is fully labelled as either normal (type 0) or anomalous (type 1, 2, 3) and attributed to either SS or RS. Data generated from SS is called simulation set while that generated from RS is called practical set.

\begin{center}
	\begin{figure*}[ht]
		\centering
		\includegraphics[width=1.1 \textwidth]{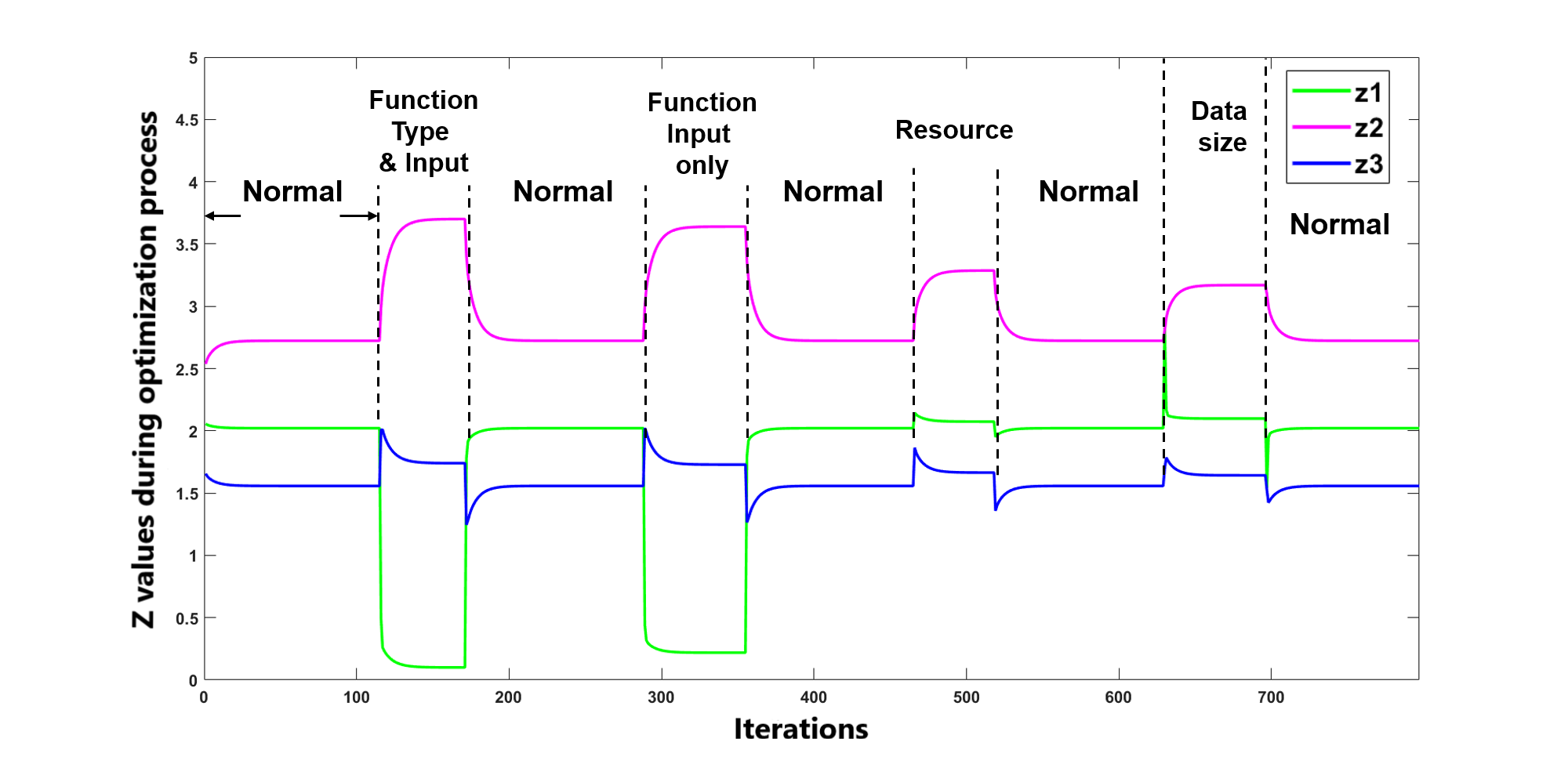}
		\caption{$z$ values of device 1, 2 and 3 under different anomalies in the SS.}
		\label{figAnomalies}
	\end{figure*}
\end{center}

Note that anomalies can happen on any device and in this section, we evaluate the anomaly detection based on anomalies occurring on device number one. This considers a reasonable scenario in a real-world IoT network, where a small number of devices (i.e. one device in our system) are attacked while the majority of devices (i.e. the other two devices) are maintained as normal. Fig. \ref{figAnomalies} demonstrates the real-time change of parameter $\textit{z}$ when anomalies happen on device one. A decrease in the $z$ of device one ($z_1$) is accompanied by an increase in the $z$ of device two and three ($z_2$ and $z_3$) when the function type and function input are manipulated in SS.

\subsection{Setup for LSTM-based networks}

For the one-layer LSTM architecture, the settings include an input feature length of $10$, resulting in an input size of 6$\times$10, which consists of  $[v_i^1, \cdots, v_i^{10}; z_i^1, \cdots, z_i^{10}]$, where $i = 1, 2, 3$ indicates the IoT device number. The step size is set as 5 and the hidden size of LSTM is set as 100. The bi-LSTM model is established based on this one-layer LSTM architecture with bidirectional mechanism. The stacked-LSTM is composed by stacking two one-layer LSTM architectures. For LSTM-attention, a multi-head attention mechanism with two heads follows the one-layer LSTM architecture. The number of input units of attention is set as 100, as the same as the hidden size of LSTM. For LSTM-encoder, an encoder-decoder based on the one-layer LSTM is established and trained at the first stage. Then the encoder part is used for extracting the hidden feature for detection. The simulation data set is split as follows: $60\%$ for training, $20\%$ for model validation and $20\%$ for simulation testing. Finally, the LSTM-based models are tested using the practical data set. Experiments are repeated ten times for each anomaly type and the mean and standard deviation of prediction accuracy are presented in Table \ref{tabresult} for the simulation test and practical prediction.

For the purpose of clear observation, we first investigate the performance of one-layer LSTM separately for each anomaly type then combine all anomaly types to assess general detection ability. Finally, we compare the performances of different variants of LSTM in detecting all anomaly types.

\section{Detection results and discussion}\label{results} 
\subsection{Anomaly detection on SS}


In this section, different anomaly types are detected on SS and the model performance is evaluated. Firstly, when generating data (i.e. ADMM parameters $\textbf{\textit{z}}$ and $\textbf{\textit{v}}$) from the SS, we investigate the scenario that only one specific type of anomaly (manipulation of function input alone) happens repeatedly. Here we should note that different one-layer LSTM models are trained for different scenarios that only consist of a specific type of anomaly, with $60\%$ of the data specified as the training set, $20\%$ of the data for validation and $20\%$ of the data for testing.



As shown in Table \ref{tabresult}, anomalies caused by manipulating the utility function input only are detected with an accuracy of $98.14\% $. Similarly, we investigated the detection performances for the other two anomaly types ``Function Type and Input" and ``Data size". Our results show that both anomalies can be detected with relatively high accuracy ($99.82\%$ and $93.91\%$) for manipulations of utility function type \& input and transmission data size respectively. These separated detection accuracies for specific manipulations reveal that the deep learning based approach is able to extract the individual pattern of each type of manipulation with very high accuracy. We note that the detection accuracy for ``Data size" is slightly lower than the detection accuracy for other types. The reason might be that the chosen data size factor (in section \ref{manipulationsetup}) leads the manipulated data size close to the correct data size and the change harder to detect.

\begin{table}[htbp]
	\caption{Accuracy of LSTM-based anomaly detection} 
	\begin{center}
		\begin{tabular}{|l|c|c|c| }
			\hline
			\textbf{Anomaly types} & \textbf{Simulation} & \textbf{Real world system} \\
			\hline
			
			Function input only & 98.14\% $\pm$ 0.52\% & 82.84\% $\pm$ 3.81\%  \\
			
			Function type and input &  99.82\% $\pm$ 0.01\% & 93.90\% $\pm$ 1.52\%  \\
			
			Data size  & 93.91\% $\pm$ 1.00\% & 92.65\% $\pm$ 0.85\%  \\

			General (two-class) & 98.81\% $\pm$ 0.38\% & 96.28\% $\pm$ 0.89\%   \\

			General (four-class) & 92.35\% $\pm$ 0.84\% & 78.88\% $\pm$ 3.80\%  \\
			
			\hline
		\end{tabular}
	\end{center}
	\label{tabresult}
\end{table}

Furthermore, when generating data ($\textbf{\textit{z}}$ and $\textbf{\textit{v}}$) from the SS, we also investigated the scenario that three types of anomalies appear randomly (only one anomaly happens each time but can be any one of the different anomaly types). Here, only one LSTM model is trained for detecting different anomalies using data from the SS, with $60\%$ of the data used as the training set, $20\%$ of the data for validation and $20\%$ of the data for testing, which is consistent with the previous setups.

Both four-class detection (with labels $0, 1, 2, 3$ for situations including normality and the different anomalies, respectively) and two-class detection (here, normality and manipulation of system resources are labelled as $0$, and other manipulations are labelled as $1$) are investigated. The prediction accuracy was found as $92.35\%$ for four-class anomaly detection and $98.81\%$ for two-class anomaly detection.

The rule-based detection in the SS is based on thresholds by identifying to which extent the $z$ value is changed. Here, the threshold was assumed to be $1\%$ of the optimal transmission frequencies of the IoT devices. Given this setting, Table \ref{tabresult_math} demonstrates the detection results obtained using this approach. Specifically, comparing with Table \ref{tabresult}, the general (two-class) results show that the LSTM-based detection can easily outperform the rule-based detection method.

\begin{table}[htbp]
	\caption{Accuracy of rule-based anomaly detection} 
	\begin{center}
		\begin{tabular}{|l|c|c|c| }
			\hline
			\textbf{Anomaly types} & \textbf{Simulation} & \textbf{Real world system} \\
			\hline
			
			Function input only & 97.48\% & 86.53\%  \\
			
			Function type and input &  99.65\% & 80.21\%  \\
			
			Data size  & 65.26\% & 66.59\%  \\

			General (two-class) & 91.78\% & 83.34\%   \\

			\hline
		\end{tabular}
	\end{center}
	\label{tabresult_math}
\end{table}



\subsection{Anomaly detection on RS} \label{Gen_anomaly}

In order to better represent detection of anomalies in a real-world IoT environment, different types of anomalies are detected using the RS in this section. We recall that the LSTM model is trained based on the simulated data from the SS and will be tested using the data from the RS in this setup. 

\begin{center}
	\begin{figure*}[ht]
		\centering
		\includegraphics[width=1.0\textwidth]{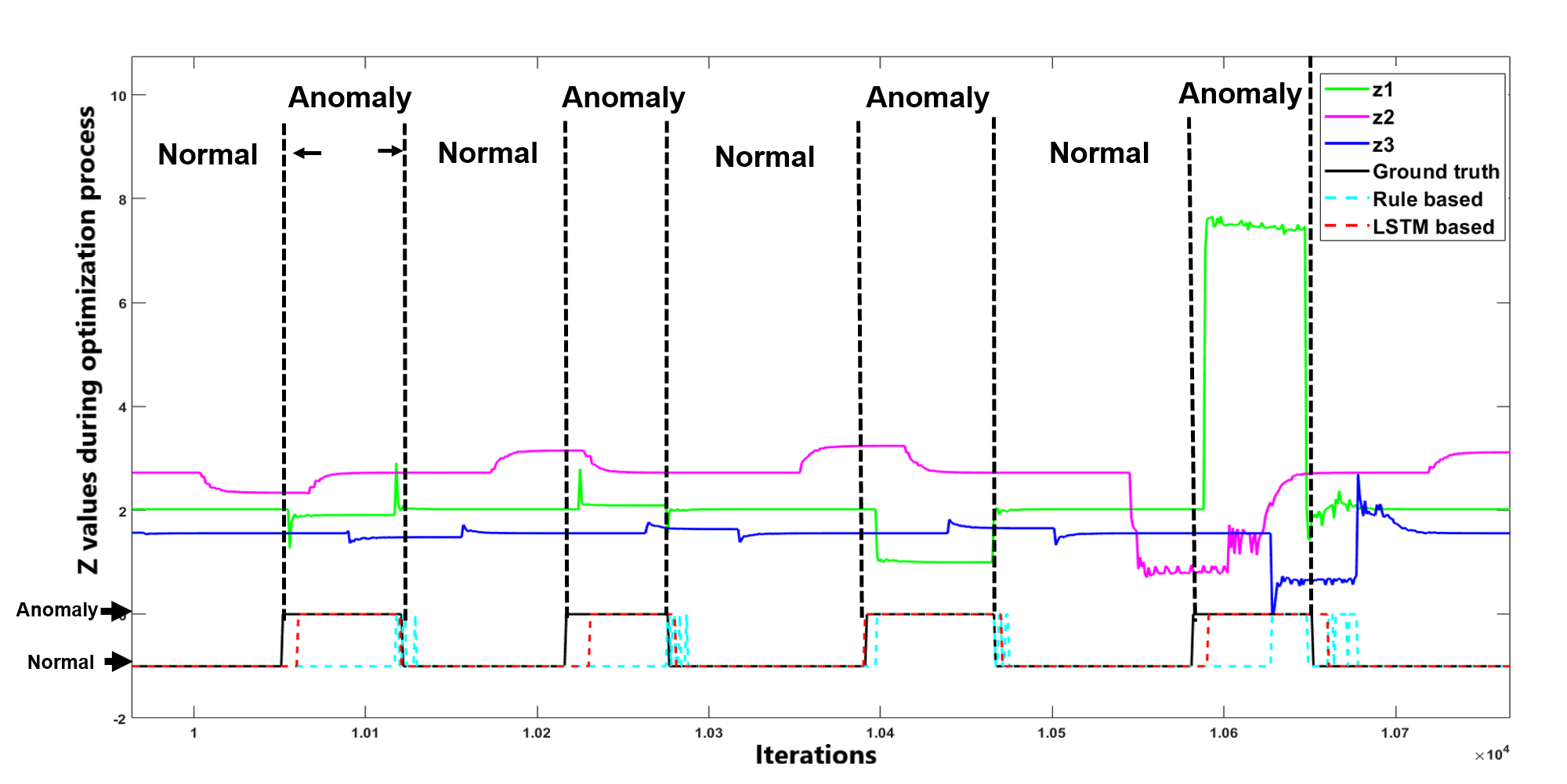}
		\caption{$z$ values of device 1, 2 and 3 in the RS, including both normal and abnormal situations. General two-class detection results from the rule-based method (turquoise dotted-line) and LSTM (red dotted-line) are compared with the ground truth (black line).}\label{figAnomalies_practical}
	\end{figure*}	
\end{center}

\begin{center}
	\begin{figure*}[ht]
		\centering
		\includegraphics[width=1.0\textwidth]{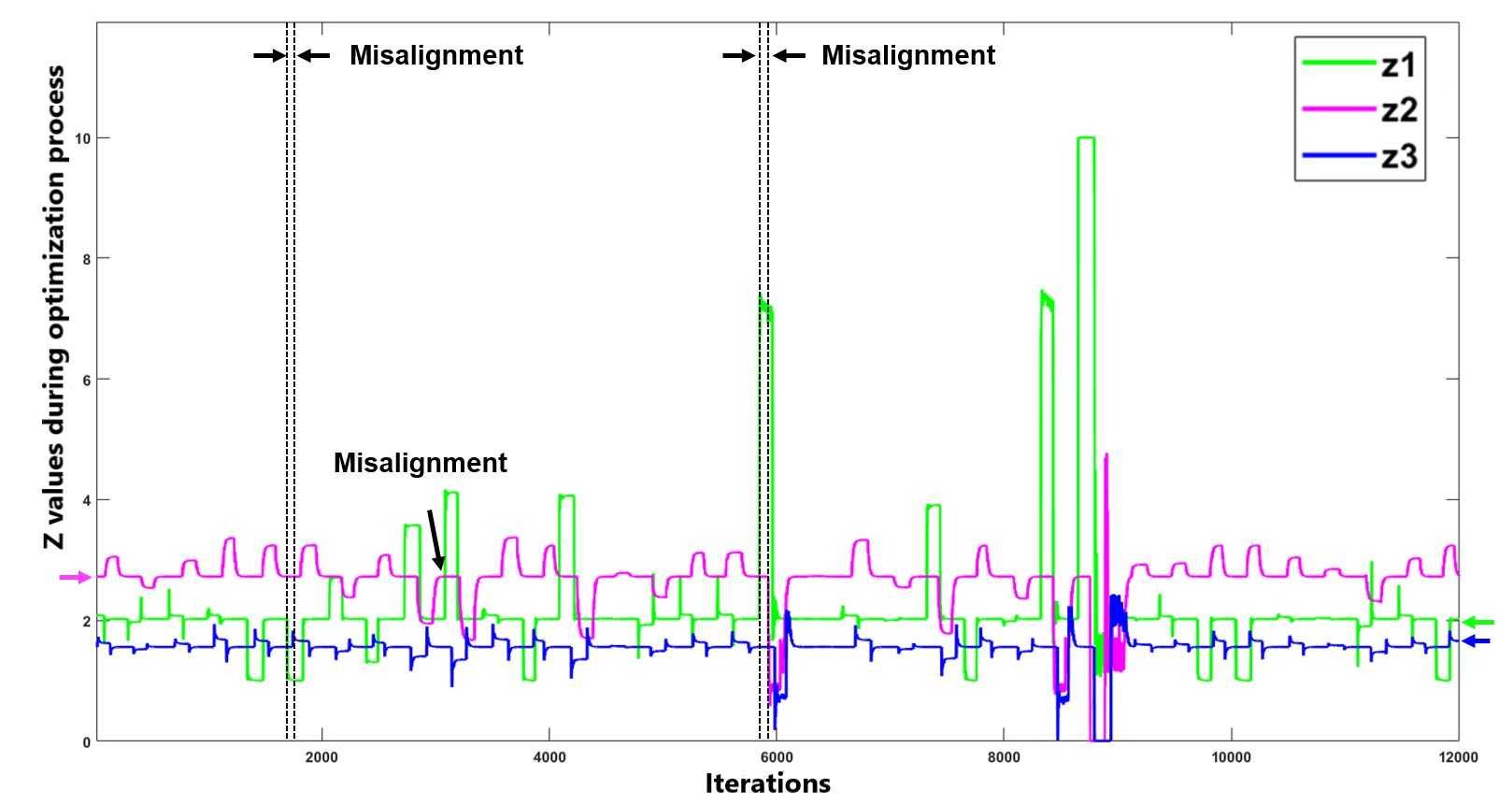}
		\caption{$z$ values of device 1, 2 and 3 during the update process in the PS. The misalignments, fluctuations and jumps can affect the performance of detection when detecting anomalies in real-world applications. }\label{figZ}
	\end{figure*}	
\end{center}

Our results in Fig. \ref{figAnomalies_practical} indicate the value of parameter $z$ for devices 1, 2 and 3 (green, magenta and blue lines, respectively) when the RS system is running normally, and in scenarios when three types of anomalies occur. In comparison to Fig. \ref{figAnomalies}, in this case, when an anomaly occurs, the value of parameter $z$ for devices 1, 2 and 3 do not change at the same time, which causes the observed misalignments with respect to iterations of the ADMM algorithm. Fig. \ref{figZ} shows the variation of parameter $z$ for device 1, 2 and 3 on RS on a long-time scale with misalignments, fluctuations and jumps.

Comparing these results to those obtained for the SS experiments (Table \ref{tabresult}), it is evident that the accuracy of anomaly detection for ``Function Input Only" from the RS ($82.84\% $) is lower than that from the SS ($98.14\% $) because of the misalignments between the $z$ values of different devices. Similarly, detection of ``Function Type and Input" and ``Data Size" anomalies in the RS ($93.90\%$ and $92.65\%$) had accuracies slightly lower than those presented in the SS simulation results. In addition, four-class detection and two-class detection were also investigated in the RS. As shown in Table \ref{tabresult}, the general two-class detection achieved the highest accuracy of $96.28\%$ in the RS, which indicates that the proposed LSTM-based method is promising for real-world IoT networks. However, with misalignments between parameter $z$ for different devices, performances from the RS for four-class detection (i.e. an accuracy of $78.88\% $) and for two-class detection (i.e. an accuracy of $96.28\% $) are reduced compared to the performances from the SS (i.e., accuracies of $92.35\% $ and $98.81\% $ for four-class and two-class prediction respectively).

Performance on the RS was poorer for the rule-based anomaly detection approach (Table \ref{tabresult_math}), which may be due to the misalignments between parameter $z$ between different devices. Since the rule-based approach leverages the simultaneous relationship between different transmission frequencies, it can be expected that a larger misalignment leads to poorer performance for the rule-based approach. The general two-class detection results from rule-based and LSTM methods are compared against ground truth in Fig. \ref{figAnomalies_practical} in the RS. The detection results from the LSTM method better match the ground truth, while the rule-based method claims the anomalies incorrectly when there are misalignments and fluctuations in the data flow. 

\vspace{-1.0cm}

In order to provide more details for comparing the performance of rule-based and LSTM methods, precision, specificity and recall metrics are calculated and shown in Table. \ref{confusion_matrix}. Note that we calculate the metrics for LSTM every $10$ time steps as the input length of the LSTM model is taken as $10$ in the model settings presented in Section \ref{setup}, while the metrics for the rule-based method are computed in each time step. When the system is running normally, both methods have a high specificity value ($0.98$ for the LSTM method and $0.95$ for the rule-based method), which means that most of the time both anomaly detectors will not alarm when the RS is running normally. However, the LSTM method obtains a higher recall value than the rule-based method for anomaly detection ($0.93$ for the LSTM method and $0.45$ for the rule-based method), indicating that the LSTM method can alarm promptly when most malicious manipulations occur, but the rule-based method fails to detect most anomalies. Given the precision values ($0.95$ for the LSTM method and $0.76$ for the rule-based method), the majority of anomalies identified by the LSTM method are real anomalies and therefore the LSTM method is more acceptable for use in real-world applications.

%
%


\begin{table}[!ht]
	\centering
	\caption{Confusion matrix on general two-class detection in the RS}
	\label{confusion_matrix}
	\begin{tabular}{c}

		\includegraphics[width=1.0\textwidth]{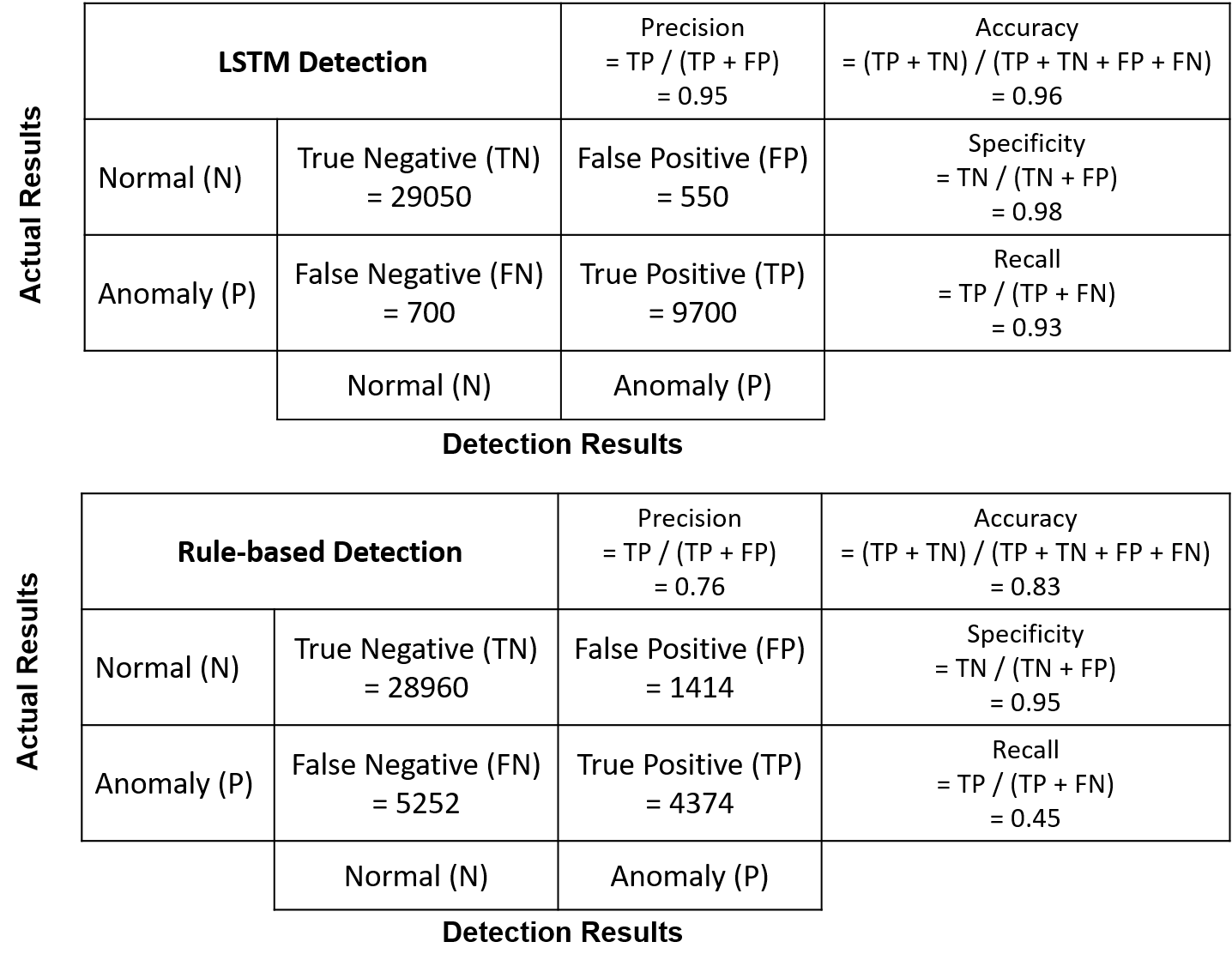}\\

	\end{tabular}
\end{table}


\section{Discussion}\label{discussion}

The results presented in Table \ref{tabresult} indicates that the one-layer LSTM detects anomalies more effectively in both the SS and the RS. In Table \ref{tabresult}, the standard deviations of the detection results reveal that LSTM-based anomaly detection is robust, including the real-world system (RS). Although the accuracy decreases to $78.88\%$ with some uncertainty (standard deviation of $3.80\%$) for four-class anomaly detection in the RS, the LSTM method can still obtain stable high performance (accuracy of  $96.28\%$ with standard deviation of $0.89\%$) for two-class anomaly detection. However, when detecting the ``Function Input Only" anomaly, the LSTM method has worse performance than the rule-based method. One possible reason for this is that the fluctuations and jumps in data flow shown in Fig. \ref{figZ} cause uncertainty during the training process of LSTM models.

\begin{center}
	\begin{figure*}[ht]
		\centering
		\includegraphics[width=1.0\textwidth]{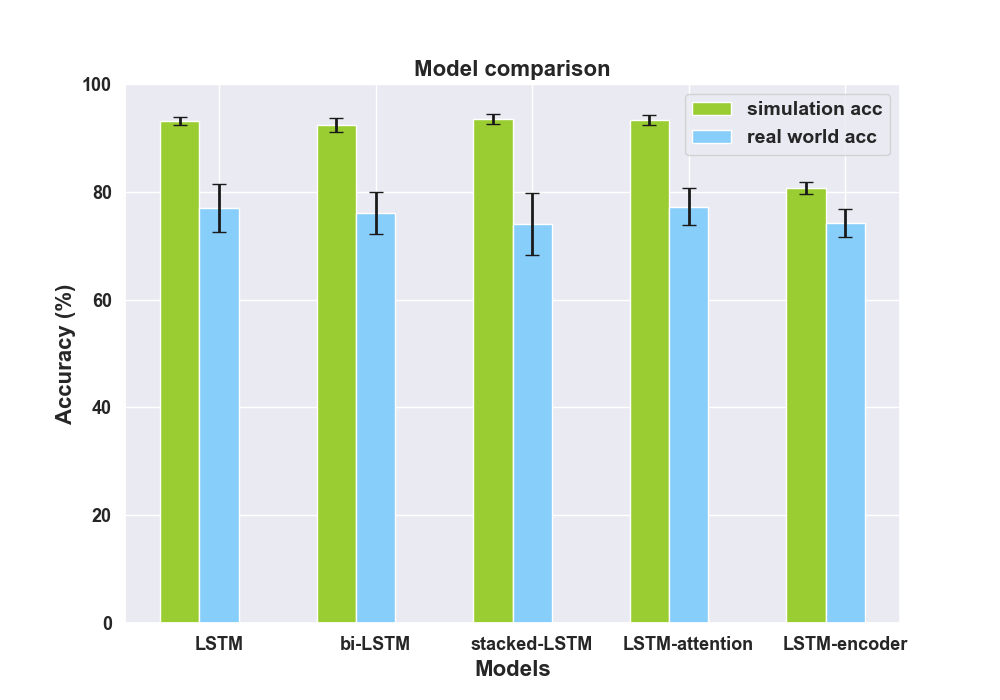}
		
		\caption{Four-class anomaly detection results from different deep learning architectures.}\label{model_comparison}
		
	\end{figure*}	
\end{center}

As applying the extra deep learning architecture may enhance the detection in complicated environment, the one-layer LSTM, bi-LSTM, stacked-LSTM, LSTM-attention and LSTM-encoder architecture are compared in four-class anomaly detection. Fig. \ref{model_comparison} demonstrates the four-class detection accuracy from one-layer LSTM, bi-LSTM, stacked-LSTM, LSTM-attention and LSTM-encoder. Each model is trained 10 times with different parameter initializations and the average detection accuracy and standard deviation are calculated. For the detection in real world environment, applying the extra deep learning techniques is not able to improve the detection accuracy apparently. By contrary, the encoder-decoder mechanism degenerates the anomaly detector in simulation environment. Table \ref{tabcomputation} shows that the complexity of different architectures. With the comparable inference time consumption, one-layer LSTM has the minimum number of parameters which means that one-layer LSTM can detect anomalies effectively with less computational resource.

\begin{table}[htbp]
	\caption{The complexity comparison of different architectures} 
	\begin{center}
		\hspace{-0.5cm}
		\begin{tabular}{|c|c|c|c|c|c| }
			\hline
			\textbf{Complexity} & LSTM & bi-LSTM & stacked-LSTM & LSTM-att. & LSTM-en. \\
			\hline
			
			No. of model parameters & 43204 & 86404 & 123604 & 73204 & 124210 \\
			
			Simulation inference time (s) &  0.66 & 1.05 & 0.93 & 0.56 & 0.36  \\
			
			Real world inference time (s)  & 0.60 & 1.13 & 0.93 & 0.52 & 0.33  \\

			\hline
		\end{tabular}
	\end{center}
	\label{tabcomputation}
\end{table}

Table \ref{tabresult_math} indicates the detection results from rule-based model. Detection of anomaly “Data Size” using the rule-based method has the lowest accuracy when compared to the other types of anomalies. The reason is that a change of transmission frequency for the manipulated device may lead to identical-trend changes of transmission frequencies for other devices, which prevents the rule-based detection working effectively. Interestingly, the detection accuracy for “Data Size” in the RS is very comparable to that in the SS which may be largely caused by the misalignments and fluctuations in the RS data flow. However, as expected, the detection accuracies for other types of anomalies in the SS are greater than those in the RS. Finally, we also investigated the impact of thresholds in rule-based anomaly detection (Table \ref{tabresult_math_threshold}). The threshold plays an important role in anomaly detection, but simply increasing or decreasing the threshold can not obtain a better performance on anomaly detection. One the one hand, a small network disturbance will trigger the anomaly alarming incorrectly if a small threshold applies. On the other hand, anomaly will be ignored because the change of transmission frequency can not trigger the detector if the threshold is too high. Therefore, a problem arises on how to select an optimal threshold for anomaly detection in a practical IoT application (i.e., trial \& error), which is another drawback of rule-based method compared to the LSTM-based approach.

\begin{table}[htbp]
	\caption{Accuracy of rule-based anomaly detection for utility function input under different thresholds.} 
	\begin{center}
		\begin{tabular}{|l|c|c|c| }
			\hline
			\textbf{Thresholds}  & \textbf{Real world system} \\
			\hline
			
			1\% optimal frequency & 86.53\%   \\
			
			5\% optimal frequency & 87.90\%   \\
			
			10\% optimal frequency &  87.02\%   \\
			
			15\% optimal frequency &  86.62\%   \\
			
			30\% optimal frequency  & 83.23\%   \\

			50\% optimal frequency & 77.10\%   \\

			\hline
		\end{tabular}
	\end{center}
	\label{tabresult_math_threshold}
\end{table}

\section{Conclusion}\label{sec:Conclusion}

In this chapter, we propose a novel transmission frequency management system for IoT edge devices. This innovative system is able to assign the optimal transmission frequency for each IoT device in the network dynamically and recalculate the new optimal transmission frequencies adaptively, when there is a new connection of a new device. Furthermore, we also devise mechanisms for anomaly detection of the system when transmission frequencies may be manipulated in different settings.

Our simulation results show that the proposed system is effective in real-world scenarios, with high accuracy for estimation of transmission frequency in a low-latency ($5\ ms$) router-based experimental IoT network. Considering that IoT edge devices may suffer attacks which manipulate their transmission frequency and transmit data streams with an incorrect cadence, we use both a mathematical rule-based and LSTM-based approach to detect the potential anomalies in transmission frequency. The rule-based approach demonstrates the internal process during an anomaly event but can not reliably detect the anomaly in a practical environment. In contrast, the LSTM-based approach indicates greater potential for implementation in both simulations and real-world environments for the detection of abnormal transmission frequency.

\chapter{Sharing-bike availability prediction} \label{bike}
\graphicspath{ {Chapter03/} }

%

\textbf{Abstract:} In this chapter, we propose novel graph neural network for sharing-bike availability prediction and discussed the impacts of different modelling methods of adjacency matrices on the proposed architecture. The work presented in this chapter has been published in \cite{chen2021comparative}.


\section{Introduction}

Most recently, there has been an increasing interest in adopting sharing bike schemes globally as these schemes can be seen as effective tools in combating global challenges such as improving sustainability (e.g., reduce the commuting cost and air pollution \cite{otero2018health}) in transportation. One of the key requirements to facilitate the bike-sharing system is whether the supply and demand can have a good balance in a bike-sharing network \cite{raviv2013optimal}. In general, the relocation of bikes ensures the balance between supply and demand, but the uncertainty of departure and arrival among different bike stations has been making the bike relocation harder to execute precisely. Therefore, accurately forecasting the availability of bike at a given time and station becomes increasingly important.

Recently, convolutional neural networks (CNN) have been applied to extract the relationship between adjacent traffic networks whilst the recurrent neural networks (RNN) were used to arrest the temporal information. For short-term traffic prediction, fully connected long short-term memory (LSTM) \cite{shi2015convolutional} and CLTFP \cite{wu2016short}, two architectures mixed the long short-term memory networks with convolutional operation, were proposed in order to catch both temporal and spatial cues. However, LSTM or other networks with recurrent architecture are computationally intensive. Also, it is harder for the network parameters to converge to global optimal values, since the recursive training process accumulates the error. On the other hand, CNN-based methods also have their limitation since the convolution process the data in 2-D form restrictively, which may not be the natural structure of traffic data.

These above issues of CNN and RNN-based methods were investigated and addressed by the spatial-temporal graph convolutional networks (ST-GCN) \cite{DBLP:conf/ijcai/YuYZ18}, a variant of a graph neural network (GNN) for utilizing spatial information. Spatial-temporal convolutional blocks were introduced and applied repeatedly in this architecture, combining several graph convolutional layers \cite{DBLP:conf/nips/DefferrardBV16} with sequential convolution in order to  represent the spatial-temporal relations. Subsequent to this approach, STG2Seq \cite{bai2019stg2seq}, a sequence-to-sequence variant of STGCN, is proposed with more reference on historical data and an attention module, for multi-step passenger demand forecasting. However, there are still some important issues to be solved in the ST-GCN architecture. For instance, how effective a specific adjacency matrix scheme can contribute to traffic demand prediction. Also, to what extent an attention-based mechanism can be applied to further improve the accuracy for a given demand prediction model. 

To answer these questions, our key objective in this chapter is to investigate how ST-GCN, supplemented with an attention-based mechanism, can further enhance the performance of bike availability prediction across different bike stations in cities. From an application/service perspective, we believe the proposed method can help cyclists make their personalised travel plan more appropriately by finding the best bike station nearby with high confidence in availability. Thus, the contribution of this chapter can be summarised as follows: 

\begin{itemize}
	\item[1.] We combine an attention mechanism with the ST-GCN, namely AST-GCN, to improve the ability of extracting spatial-temporal features for the prediction task. In comparison with the existing methods, our model shows a promising performance. 
	\item[2.] We review related works in the recent literature and summarise four categories for modelling adjacency matrices, namely spatial based, temporal based, spatial-temporal based and adaptive based adjacency matrix.
	\item[3.] Given our findings in 1 and 2, we evaluate our proposed AST-GCN model with the adjacency matrices of interest using a real-world dataset, Dublinbike, for bike sharing availability prediction. Our results show that: (a) adaptive spatial-temporal adjacency matrix can achieve the best performance; (b) spatial-temporal based adjacency matrix can achieve better results than that only using spatial-based or temporal-based adjacency matrix; (c) spatial-based adjacency matrix achieves similar performance as the temporal-based one.
\end{itemize}

The rest of the chapter is organised as follows. We introduce some previous research related to traffic demand prediction in section \ref{RW} and formulate our problem in section \ref{Method}. Experimental setups are demonstrated in section \ref{experimentch3} and the results are discussed in section \ref{result}. Finally, we summarise our work in section \ref{concl}.

\section{Related Work} \label{RW}

\subsection{Existing Methods}
In general, forecasting traffic demand is difficult, when a traffic demand depends not only on the historical demand pattern of the target area (e.g., suburb) but also on the pattern of other areas (e.g., urban). To meet this challenge, many studies using deep learning such as CNN, RNN, and GNN have been proposed.

As the traditional convolutional operation in CNN process the data with a 2D approach, the layout of a city is geographically divided into square blocks in order to extract spatial relationships from all regions \cite{zhang2017deep}, nearest regions \cite{yao2018deep} or in other 2D forms \cite{chu2020passenger}. RNN based methods and their variants \cite{yao2019revisiting} are applied to catch temporal correlation, for instance, structuring the historical traffic demand sequence for each region \cite{shi2015convolutional} and presented as a 1D feature-level fused architecture \cite{wu2016short}. GNN based methods, with natural advantages in utilizing spatial information, model the traffic network by a general graph instead of treating the traffic data arbitrarily (e.g., grids and segments) in CNN and RNN methods. GCN, as a variant of GNN, which is able to combine spatial and temporal information, is widely used in the scenario of traffic demand prediction as seen in many recent works \cite{DBLP:conf/nips/DefferrardBV16} \cite{bai2019stg2seq} \cite{DBLP:conf/ijcai/YuYZ18}.

Attention is a popular technique in deep learning that mimics physiological cognitive attention. The effect enhances the importance of small parts of the input data and de-emphasising the rest. This technique has been used to enhance the prediction performance for many sequence-based tasks of GNNs, i.e. Graph attention networks \cite{velivckovic2017graph}. In traffic demand prediction, the importance of each previous step to target demand is different, and this influence changes with time. For instance, a temporal attention mechanism \cite{bai2019stg2seq} is able to add an importance score for each historical time step to measure the influence and this strategy can effectively improve the prediction accuracy.

\subsection{Adjacency Matrices} \label{AM}
An adjacency matrix is used to indicate whether a pair of vertices is connected by edge or not in graph data. For a traffic network, it is important to understand how an adjacency matrix can be used to best capture the interconnectivity between different nodes in the graph. To the best of our knowledge, four types of adjacency matrices have been investigated in  previous research works, namely spatial (S), temporal (T), spatial-temporal (ST) and adaptive (A). A spatial adjacency matrix is usually distance-based. Euclidean distances between different stations (i.e., nodes in graph) \cite{DBLP:conf/ijcai/YuYZ18}  \cite{chen2020multitask} or the natural geographical distance \cite{kim2019graph} are usually used as weights for its entries. For instance, a shorter geographical distance between two stations may indicate a stronger connection in the graph.
A temporal adjacency matrix can be defined based on the similarity score \cite{bai2019stg2seq} (i.e., Pearson correlation coefficient) between the temporal information (i.e., historical traffic demand sequence) of each pair of nodes/stations. For example, a larger value of Pearson coefficient calculated from the time sequential data for the number of available bikes between two stations, may indicate a stronger connection in graph between these two stations compared to other pairs. To combine the benefits of both spatial and temporal features, an spatial-temporal embedding (ST embedding) can be generated for each node in a graph \cite{ye2020coupled}. However, in such a scenario, it can be hard to describe the adjacency matrix intuitively with the high dimension embedding features and thus the adjacency matrix needs to be adaptively defined along with the training process of GCN \cite{wu2019graph} \cite{chiang2019cluster}.

\section{Methodology}\label{Method}

\subsection{Notations and Problem Statement}

We consider a scenario where $N$ bikes stations are included as part of a bike-sharing system. Let $\underline{\textbf{N}}:= \left\lbrace 1,2,\ldots, N \right\rbrace$ be the set for indexing the bike stations in the system. For a given bike station $i \in \underline{\textbf{N}}$, let $A^i_t \in \R$ be the number of available bikes at the station $i$ at time $t$. We denote $\textbf{A}_t \in \R^{N}$ the vector consisting of the number of available bikes across all stations $N$ at time $t$. In addition, each bike station $i$ is associated with a set of features for model training, e.g. weather condition, day of week, etc, and let  $F^i_t \in \R^{d}$ represent the values of its features at time $t$, where $d$ is the number of features used. Similarly, we let $\textbf{F}_t \in \R^{N \times d}$ be the feature set values of all bike stations at time $t$. Given the notation above, our learning objective is to find a function $\textbf{H}(.)$ which is able to address the following problem:
\begin{equation*}
	\textbf{A}_{t+1:t+n} = \textbf{H}(\textbf{A}_{t-m+1:t}; \textbf{F}_{t-m+1:t})
\end{equation*}

\noindent where $m, n$ denotes the input and output length for the model respectively. Also, the notation ${t+1:t+n}$ presents the output as a sequence of vectors from steps $t+1$ to $t+n$.

\subsection{Attention-based ST-GCN}

In this section, we introduce the attention-based ST-GCN architecture that  used for solving our bike sharing availability prediction problem. We note that the ST-GCN architecture has been presented in \cite{DBLP:conf/ijcai/YuYZ18}, and the architecture consists of two identical ST-Conv-Blocks and a fully connected output layer. Specifically, an ST-Conv-Block consists of two temporal gated convolutional (TGC) layers and one spatial graph convolutional (SGC) layer, which are the essential modules of ST-GCN. In general, TGC is in charge of extracting temporal features and SGC is able to extract spatial features from the data. However, since there is no attention on the temporal channel of ST-GCN, this significantly degrades the performance for sequence to sequence based learning tasks. As such,  the model's learning capability may be significantly reduced due to ``lost of focus''. To deal with this issue, we introduce a temporal-attention module (TAM) in each ST-Conv-Block, as shown in Fig. \ref{fig:stgcn} where the temporal-attention module is depicted in green.  \\

\noindent \textbf{Remark:} An attention mechanism was introduced in \cite{shiraki2020spatial} and \cite{zhang2020sta} to extract both spatial and temporal information from ST-GCN networks. The architectures proposed in both works applied attention operation to extract spatial and temporal information separately. In particular, the model in \cite{shiraki2020spatial} consisted of 15 ST-Conv blocks in total with two attentions matrices calculated from them, while the model in \cite{zhang2020sta} was stacked by 10 ST-Conv blocks with two attention matrices computed from each ST-Conv block. With increased model complexity and computation cost, stacking multiple ST-Conv blocks with attention matrices calculated separately may be of less interest since the spatial and temporal information may not be combined towards an effective spatial-temporal embedding in such a case. Instead, our model only consists of 2 ST-Conv blocks and the proposed AST-GCN architecture lightly merges spatial-temporal information with attention by calculating the attention matrix only once in each ST-Conv Block, which reduces the computation costs during the model training process. Specifically, the first TGC module generates original temporal information and the last TGC module generates spatial-temporal information (as it takes account of the output of the preceding SGC layer as its input). Passing through two average 3D pooling layers, both temporal and spatial-temporal information are combined before a Relu activation function is applied. A sigmoid function is connected here to generate probabilistic weights (attention matrix) with values between 0 and 1. With this matrix in place, the attention-based temporal information is generated by using a dot product with the output of the first TGC layer and then concatenated as input to the subsequent ST-Conv Block. Both spatial and temporal information in the data flow are fully captured before passing to the dense layer for sequential output prediction. 
\vspace{-0.3cm}

\begin{center}
	\begin{figure*}[htbp]
		\centering
		\includegraphics[width=5.5in, height=2.6in]{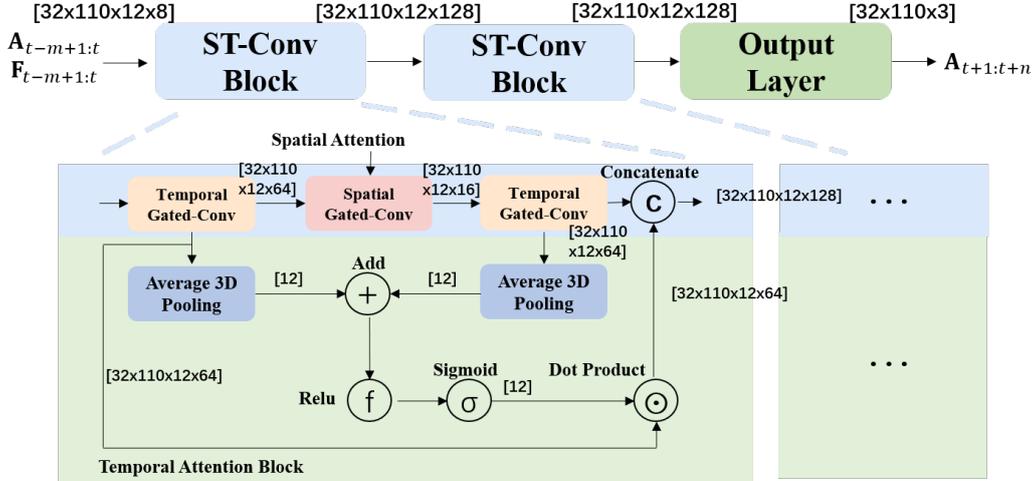}
		\caption{The proposed ST-GCN with a temporal-attention module (TAM).}\label{fig:stgcn}
	\end{figure*} \vspace{-0.1cm}
\end{center}

\section{Algorithms and Experiments} \label{experimentch3}

In this section, we discuss the different configurations investigated for comparative studies. 

\subsection{Experimental Datasets}
\begin{figure}[htp]
	\centering
	\includegraphics[width=11cm]{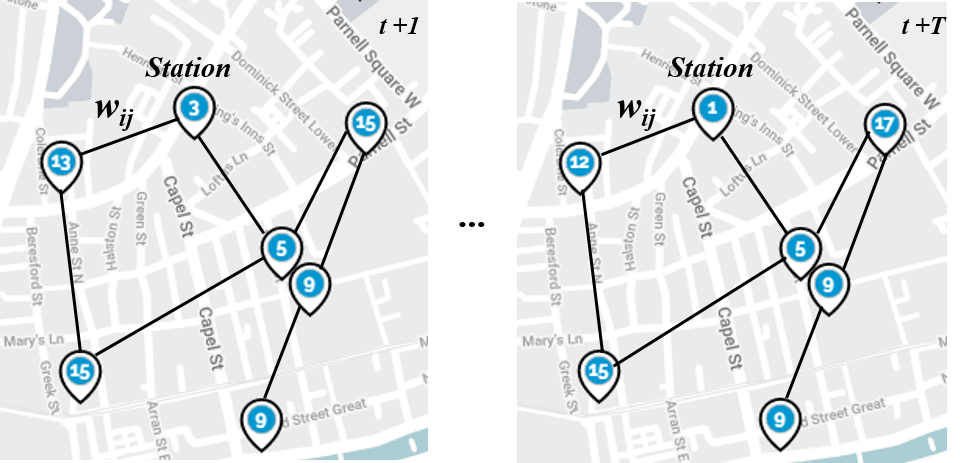}
	\caption{A subset of bike stations of the Dublin bike sharing system operates in real-time. The edges are added for the illustration of the inherent graph signals.}
	\label{fig:bike}
\end{figure}

\begin{itemize}
	\item Dublinbike:
	DublinBikes is a bike-sharing scheme in operation in Dublin City, Ireland. The system is illustrated in Fig. \ref{fig:bike}, where each node is a bike station and each blue number in the circle indicates the number of available bikes in real-time. Real-time data is accessible using an API and we also have access to historic data, recorded every five minutes, which includes timestamps, station states, number of available bikes and station locations, etc. We choose the data\footnote[1]{https://data.smartdublin.ie/dataset/analyze/33ec9fe2-4957-4e9a-ab55-c5e917c7a9ab} from 01/07/2020 to 01/10/2020 for our studies.
	\item NYC-Bikes\cite{2016DNN}: This dataset includes the NYC Citi daily bike orders of people  using the bike sharing scheme. We choose the transaction records from April 1st, 2016 to June 30th, 2016 (91 days). This contains the following information: bike pickup station, bike drop-off station, bike pick-up time, bike drop-off time and trip duration.
	\item Visualcrossing Weather Data\footnote[2]{https://www.visualcrossing.com/weather-data}: This dataset provides weather conditions at different locations at different historical time points, including temperature, humidity and wind speed, etc. This weather dataset has been integrated for experiments that use  the Dublinbikes dataset.
\end{itemize}

\subsection{Experimental Setup}
\begin{itemize}
	\item Dublinbike: 
	For this scenario, we use the number of available bikes at each bike station in the first 3 hours to predict the number of available bikes at each bicycle station 45 minutes later, where each data point is the averaged number of available bikes in 15 minutes. This implies that we take the past 12 consecutive observation points to predict the following 3 points of our interest. The dataset consists of 110 bike stations in total. The data is then separated into a training set (60$\%$), a validation set (20$\%$) and a testing set (20$\%$) in a sequential manner.
	\item NYC-Bikes:
	NYC Citi Bike is dock-based and every depot of bikes is considered as a station. Following the same experiment setup as in CCRCN \cite{ye2020coupled}, we filter out the stations with fewer orders and keep the 250 stations with the most orders. The time step is set to half an hour. Among the last four weeks considered, the first two are used for validation, and the last two are for testing.
\end{itemize}

To evaluate the performance across different models, Mean Absolute Error (MAE) has been selected as the performance metric, indicating an intuitive margin between the predicted and the true amount of available bikes at each station.

\subsection{Baseline Algorithms}
\begin{itemize}
	\item Dublinbike:
	To the best of our knowledge, there has been no GNN based methods implemented for the Dublinbike dataset. In particular, there has also been no ST-GCN based methods applied for solving the prediction for this dataset. For comparative studies, we conduct the experiments and use ST-GCN \cite{DBLP:conf/ijcai/YuYZ18} as our baseline.
	\item NYC-Bikes:
	A lot of methods have been reported using this dataset to predict traffic demand. The state of the art work is presented in CCRCN \cite{ye2020coupled}. Based on this,  we compare the performance of different methods, including our proposed model, in a similar experimental setting. Specifically, the following methods are compared: (a) HA \footnote[3]{The average of historical values at previous time steps of a fixed length.}; (b) XGBoost\cite{xgboost2016}; (c) FC-LSTM\cite{lstm1997}; (d) DCRNN\cite{2017arXiv170701926L}; (e) ST-GCN\cite{DBLP:conf/ijcai/YuYZ18}; (f) STG2Seq\cite{bai2019stg2seq}; (g) GraphWaveNet\cite{wu2019graph} and (h) CCRNN\cite{ye2020coupled}.
\end{itemize}

\subsection{Network Setup}\label{net_set}

The historical data length used for both  the Dublinbikes dataset and the NYC-citi dataset is set to 12, the prediction length is set to 3 in Dublinbikes and 12 in NYC-citi respectively. The feature dimension used in NYC-citi is 2 representing the pick-up and drop-off demand. The feature dimension used for the  Dublinbikes dataset is 8, details of the feature selection will be discussed in the results section.  All models are optimised by the Adam algorithm \cite{kingma2017adam}. Other setting of parameters  are presented in Table \ref{tab:Expri_settings}. The dimensions of the data flow during the training process of the proposed model are overlapped in Fig. \ref{fig:stgcn} for illustration purposes. It is worth noting that the input of the first temporal gated-Conv is strictly the same as the input of the corresponding ST-Conv block while the input of the second temporal gated-Conv is the output of previous spatial gated-Conv block. The concatenate operation concatenates the output of the first and the second temporal gated-Conv block. 
\begin{table}[htbp]
	\caption{Experiment settings for two datasets } 
	\begin{center}
		\begin{tabular}{|l|c|c|c| }
			\hline
			\textbf{Setup} & \textbf{Dublinbikes} & \textbf{NYC-citi}\\
			\hline
			Station amount  & 110 &  250\\
			Historical data length  & 12 &  12\\
			Prediction length & 3 & 12\\
			Feature dimension & 8 & 2\\
			Batch size & 32 & 32\\
			Initial learning rates & 0.001 & 0.0001\\
			Optimiser & Adam algorithm & Adam algorithm\\
			Weight decay & 0.001 & N/A\\
			LR adjustment strategy & cosine annealing & adjust at equal intervals\\
			\hline
		\end{tabular}
	\end{center}
	\label{tab:Expri_settings}
\end{table}

\subsection{Adjacency Matrix Setup}
The adjacency matrix in the original ST-GCN architecture is not adjustable/trainable. As a result, this fixed adjacency matrix may not fully capture the spatial relationship between nodes in the graph. To improve it, we adapt the fixed adjacency matrix to a trainable adjacency matrix and then initialise the matrix using meaningful contextual information, e.g. distance between nodes, similarity between stations’ historical time-series data. Further, an adaptive adjacency matrix (AAM) is able to extract spatial attention information from the graph adaptively, and thus it makes the AST-GCN effective in capturing both spatial and temporal attention information. For our comparative studies, different setups of adjacency matrices are investigated as follows:


\begin{itemize}
	\item For the implementation of the adjacency matrix proposed in ST-GCN \cite{DBLP:conf/ijcai/YuYZ18}, the sigma is set to 0.2 and the epsilon is set to 0.368;
	\item For the implementation of the adjacency matrix proposed in STG2Seq\cite{bai2019stg2seq}, the sigma is set to 0.05;
	\item For the implementation of the adjacency matrix proposed in CCRCN\cite{ye2020coupled}, the dimension of station feature is set to 20 and the sigma is set to 1.
	\item Other adjacency matrices do not need parameters to be set. In other words, these adjacency matrices are calculated directly without parameters or are purely adaptive. 
\end{itemize}

\section{Results and Discussion} \label{result}
\subsection{Feature Selection for Dublinbikes Dataset}

In order to select the best features for our experiments, an ablation study has been carried out for a set of features which model temporal, spatial as well as weather characteristics. Specifically, we adopt ST-GCN as our basic setting for evaluation of different feature combinations. Our full feature sets are as follows: (1) number of available bikes (AB); (2) time of day (TD); (3) day of week (WD); (4) weather condition description (WCD); (5) temperature (T); (6) wind speed (WS); (7) cloud coverage (CC) and (8) Humidity (H). 

Results of the ablation study are reported in the Table \ref{tab:abl_result}, from which we easily conclude that the following feature combination gives the best performance: number of available bikes (AB), time of day (TD), day of week (WD) and weather conditions description (WCD). 

\begin{table}[htbp]
	\caption{Results of the ablation study of feature combinations}
	\begin{center}
		\begin{tabular}{|l|c|c|c| }
			\hline
			\textbf{Feature combination} & \textbf{MAE}\\
			\hline
			AB  & 3.24\\
			AB+TD & 3.19\\
			AB+TD+WD & 3.21\\
			\textbf{AB+TD+WD+WCD} & \textbf{3.16}\\
			AB+TD+WD+WCD+T & 3.36\\
			AB+TD+WD+WCD+WS & 3.40\\
			AB+TD+WD+WCD+CC & 3.30\\
			AB+TD+WD+WCD+H & 3.40\\
			All Together & 3.56\\
			\hline
		\end{tabular}
	\end{center}
	\label{tab:abl_result}
\end{table}

\subsection{Results Discussion}
\begin{center}
	\begin{figure*}[ht]
		\centering
		\includegraphics[width=6.0in]{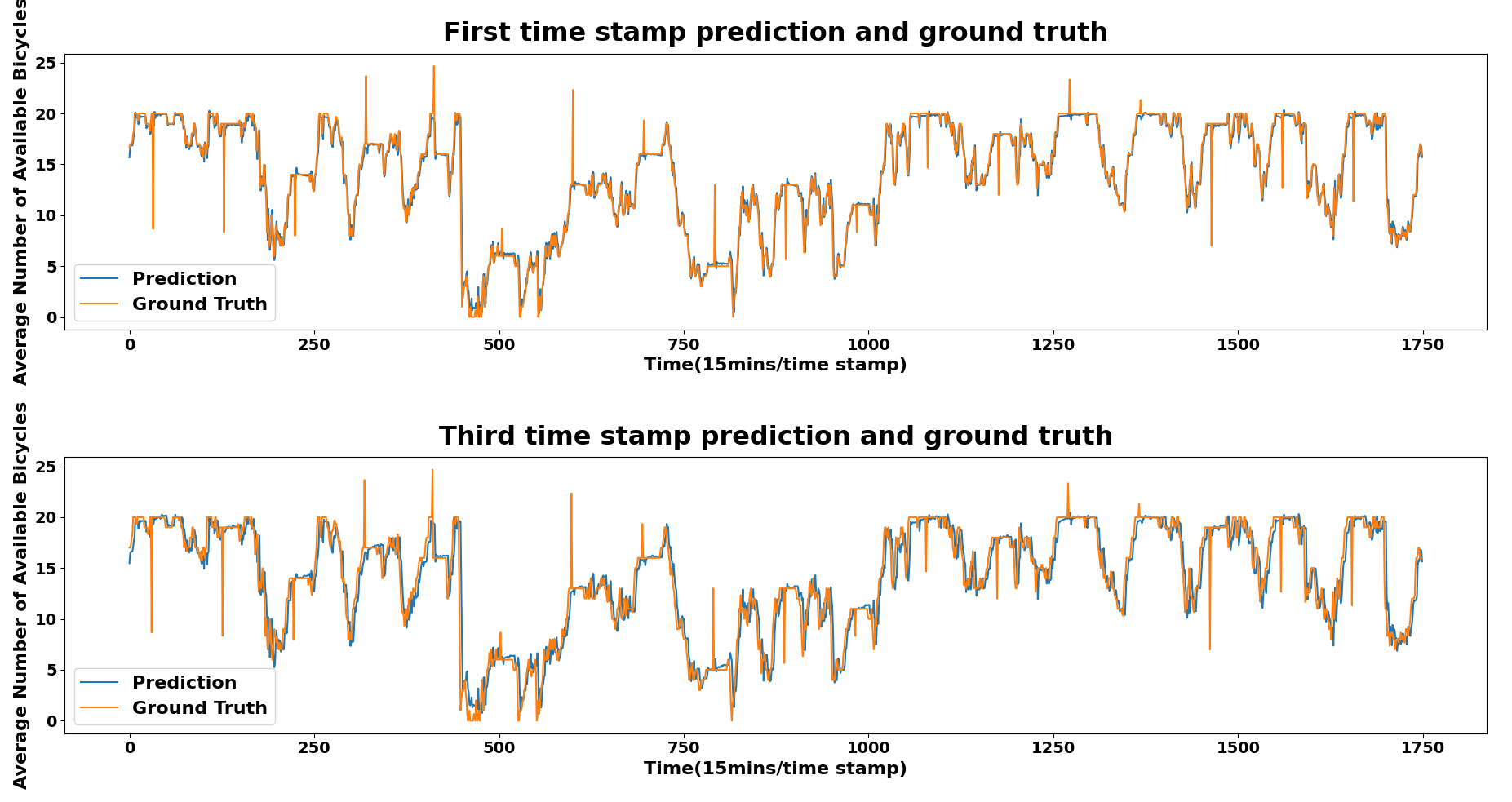}
		\caption{Comparison between ground truth and prediction.}\label{fig:GTP}
	\end{figure*}	
\end{center}

\begin{itemize}
	\item NYC-Bikes:
	The results on the NYC dataset are compared between the proposed AST-GCN and the existing algorithms reported in \cite{ye2020coupled} as shown in Table \ref{tab:nyc_stagcn}. It is shown that the AST-GCN algorithm outperforms the existing graph based architectures (i.e., ST-GCN and STG2Seq) with 24.67\% improvement in MAE, from 2.4976 to 1.8815. Also, although CCRNN beats all of its competitors, the AST-GCN shows minor difference in performance, and it still demonstrates comparable metrics compared to other sequence based models including Graph WaveNet, DCRNN.
	
	\begin{table}[htbp]
		\caption{Experiment result of AST-GCN on NYC-citi \cite{ye2020coupled}} 
		\begin{center}
			\begin{tabular}{|l|c|c|c| }
				\hline
				\textbf{Model} & \textbf{MAE}\\
				\hline
				HA & 3.4617\\
				ST-GCN & 2.7605\\
				STG2Seq & 2.4976\\
				XGBoost & 2.4690\\
				FC-LSTM & 2.3026\\
				Graph WaveNet & 1.9911\\
				DCRNN & 1.8954\\
				\textbf{AST-GCN + EAAM} & \textbf{1.8815}\\
				CCRNN & 1.7404\\
				\hline
			\end{tabular}
		\end{center}
		\label{tab:nyc_stagcn}
	\end{table} 
	
	\item Dublinbikes:
	As shown in Table \ref{tab:stagcn}, after applying distance initialised AAM (DIAAM) on ST-GCN, the prediction achieve better results with MAE equals 1.27. By replacing ST-GCN with AST-GCN, the MAE result has been significantly improved from 1.27 to 1.04. Among others, the embedding AAM (EAAM) makes the best performance which leads to the MAE equals 1. Results in Fig. \ref{fig:GTP} further highlight this key finding. Specifically, the biases between the ground truth and the first timestamp (i.e. NAB prediction for the first 15 minutes) as well as the third timestamp (i.e. NAB prediction for the 45 minutes) are both negligible showing that our proposed model can achieve impressive prediction performance for both short-term (15 mins) and long-term (45 mins) for the best case scenario.
\end{itemize}

\begin{table}[htbp]
	\caption{Experiment result of AST-GCN on Dublinbikes} 
	\begin{center}
		\begin{tabular}{|l|c|c|c| }
			\hline
			\textbf{Model} & \textbf{Categories} \tablefootnote{The abbreviations in this column have been presented in Section \ref{AM}.} & \textbf{MAE (\%)} \\
			\hline
			ST-GCN + Euclidean distance& S & 1.36 (0\%) \\
			
			ST-GCN + DIAAM & S + A & 1.27 (-6.67\%) \\
			
			AST-GCN + DIAAM & S + A & 1.04 (-23.5\%) \\
			
			\textbf{AST-GCN + EAAM \cite{wu2019graph}} & ST + A & \textbf{1.00 (-26.5\%)}\\
			
			AST-GCN + Euclidean distance \cite{DBLP:conf/ijcai/YuYZ18}& S  & 1.06 (-22.0\%) \\
			
			AST-GCN + Geographical distance \cite{kim2019graph}& S & 1.09 (-19.8\%) \\
			
			AST-GCN + Temporal correlation \cite{bai2019stg2seq} & T & 1.07 (-21.3\%) \\
			
			AST-GCN + ST embedding \cite{ye2020coupled} & ST & 1.01 (-25.7\%) \\
			
			\hline
		\end{tabular}
	\end{center}
	\label{tab:stagcn}
\end{table}

\subsection{Performance Evaluation w.r.t. Adjacency Matrices}

In this section, we discuss how different adjacency matrices can affect the learning performance for our proposed AST-GCN architecture. Our results are illustrated in Table \ref{tab:stagcn} where the percentage in parenthesis shows the difference of the achieved MAE in comparison to the basic setting: ST-GCN + Euclidean distance. Unsurprisingly, our results show that those fixed adjacency matrices, including both spatial based and temporal based, achieve the worst results among all other settings. In contrast, the adaptive-based settings can generally achieve better results compared to the fixed types, but with one exception for the spatial-temporal based setting, i.e. AST-GCN + ST embedding, which also shows a competitive result. For the adaptive-based settings, the embedding AAM, i.e. AST-GCN + EAAM, achieves the best result compared to the other AAM setting initialised by distance, i.e. AST-GCN + DIAAM.

\subsection{Performance Evaluation w.r.t. Different Bike Stations} 

\begin{center}
	\begin{figure*}[htbp]
		\centering
		\includegraphics[width=6.0in]{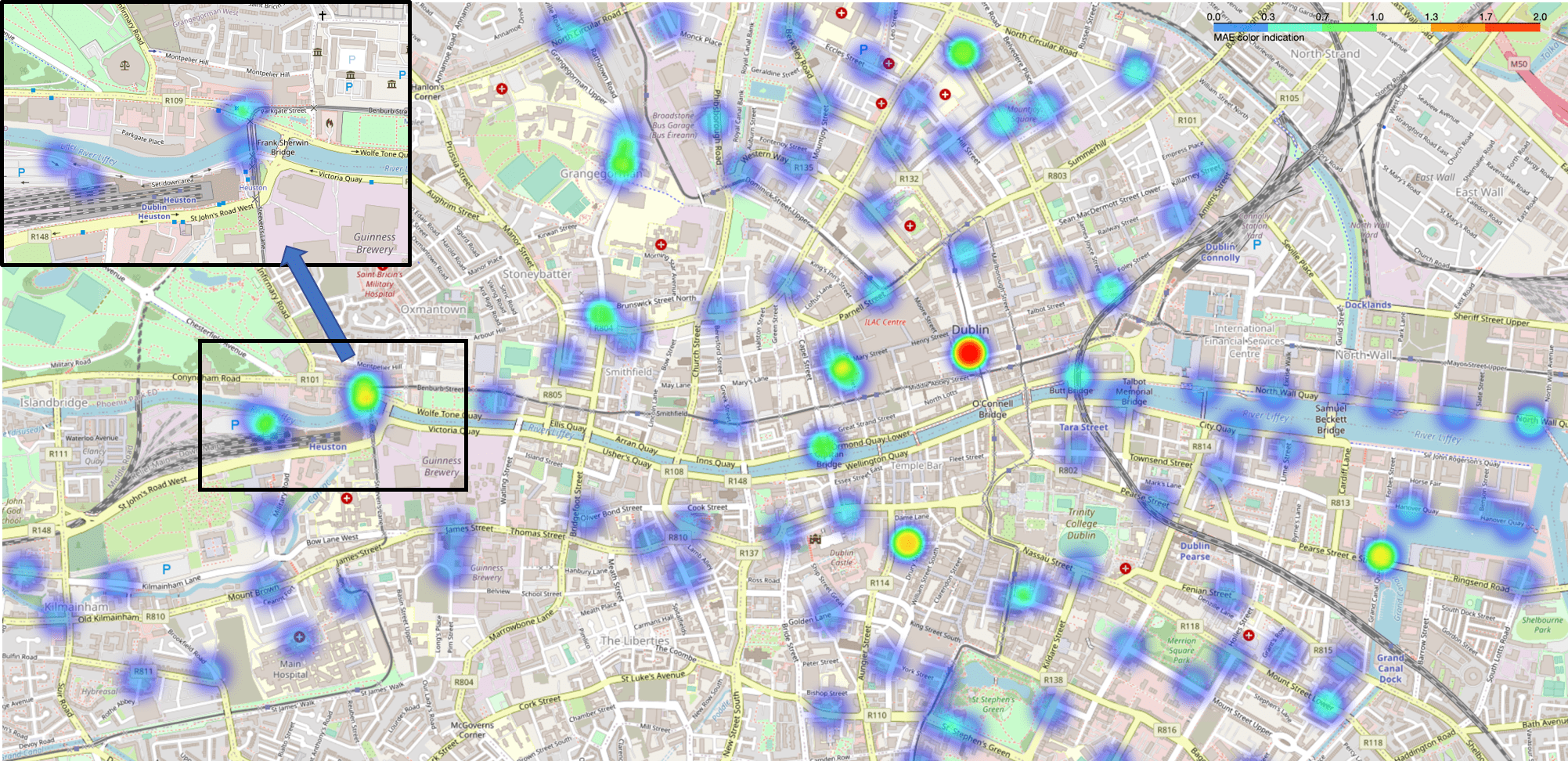}
		\caption{Heat-map of the station-wise MAE-based prediction error in Dublin city. A warm-toned color (red) indicates a higher MAE and a cool tone color (green-blue) indicates a lower MAE.}
		\label{fig:heatmap}
	\end{figure*}	
\end{center}

\begin{figure}[htbp]
	\centering
	\includegraphics[width=6.0in]{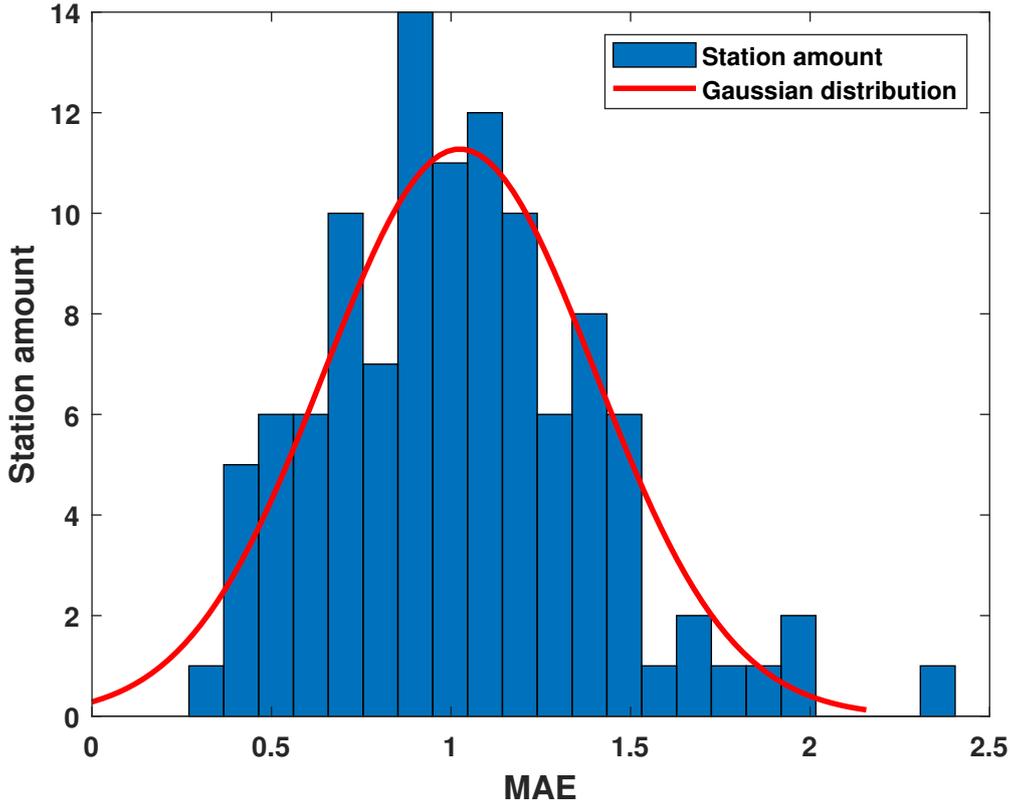}
	\caption{Histogram of station-wise MAE.}
	\label{fig:histogram}
\end{figure}

In this section, we present the prediction results for each bike station in the Dublinbike dataset using the best trained model (AST-GCN + EAAM). Our objective here is to illustrate the confidence with which a user can rely on our proposed prediction model to make a decision when he/she decides to get access to a bike from his/her nearby area. Our station-wise results are illustrated in Fig. \ref{fig:heatmap} and Fig. \ref{fig:histogram}. Specifically, Fig. \ref{fig:heatmap} shows the heat-map of station-wise MAE over the geographical map of Dublin city where the bike stations are facilitated. The red marks indicate a higher MAE and blue-green marks indicate a lower MAE in the corresponding area. Generally speaking, the results demonstrate that the prediction is more accurate (low-MAE values) outside of the city center showing that users can collect bikes with high confidence in the availability of bikes. The highest prediction error occurs in the heart of city centre, i.e. the bike station located at the ``Princes Street/O'Connell Street'', with the MAE equalling to 2.4. This may be caused by a frequent access and return of bikes by users in this central commuting area, leading to a relatively higher uncertainty in bike availability. The second highest prediction error appears in the western part of the city, i.e. the green-blue region indicated in the rectangular box in Fig. \ref{fig:heatmap}. However, this is mainly due to the aggregated effect where a few bike stations are very close to each other in the ``Benburb Street'' area. An in-depth view of the region, shown in the upper left corner of the rectangular box in Fig. \ref{fig:heatmap}, further validates that the prediction error of each bike station therein is low. Another reason causing the relative high prediction error in ``Benburb Street'' area may be the train arrivals in Heuston Station. The frequent access and return of the sharing bikes by travellers travelling by train may be challenging for the model to predict the availability of bikes. Finally, the statistical histogram of the station-wise MAE is illustrated in Fig. \ref{fig:histogram} showing that most bike stations have an MAE-based prediction error less than 1.5 bikes, which indicates that our proposed forecasting system is very robust and accurate for a number of bike stations in the Dublin city.


\section{Conclusion} \label{concl}

In this chapter, we propose a spatial-temporal graph convolutional network architecture embedded with a temporal-attention module (AST-GCN) to predict the number of available bikes in bike-sharing systems  using realistic datasets. The temporal attention module is able to extract temporal attention information which aims to enhance the prediction accuracy compared to that of the original ST-GCN architecture reported in \cite{DBLP:conf/ijcai/YuYZ18}. Our experimental results show that the proposed AST-GCN can perform better than most of existing methods in the NYC-Citi dataset. As for the Dublinbikes dataset, our proposed model has demonstrated a very promising result of 1.00 MAE as the selected performance metric. In addition, we have thoroughly investigated how different modelling of the adjacency matrices can affect the overall model performance through a comprehensive comparative study on the DublinBikes dataset. Current results have shown that embedding AAM can achieve the best results compared to many other settings. 

To conclude, we believe that the work presented in this chapter is an important step towards making  bike sharing systems more efficient thanks to the ST-GCN enabled techniques. A deep exploration on different adjaceny matrices reveals that embedding adaptive adjacency matrix can achieve the best performance in this work.

\chapter{Lane changing intention detection}  \label{lane}
\graphicspath{ {Chapter04/} }

\textbf{Abstract:} In this chapter, we detect traffic flow caused by lane changing intentions on the highway traffic networks, with graph modelling on the data generated by a popular mobility simulator. This chapter is related to the work \cite{wu2022lane}, which has been accepted by the 25th IEEE International Conference on Intelligent Transportation Systems (IEEE ITSC 2022) and will be published.

\section{Introduction}

\begin{figure}[htp]
	\centering
	\includegraphics[width=10.0 cm]{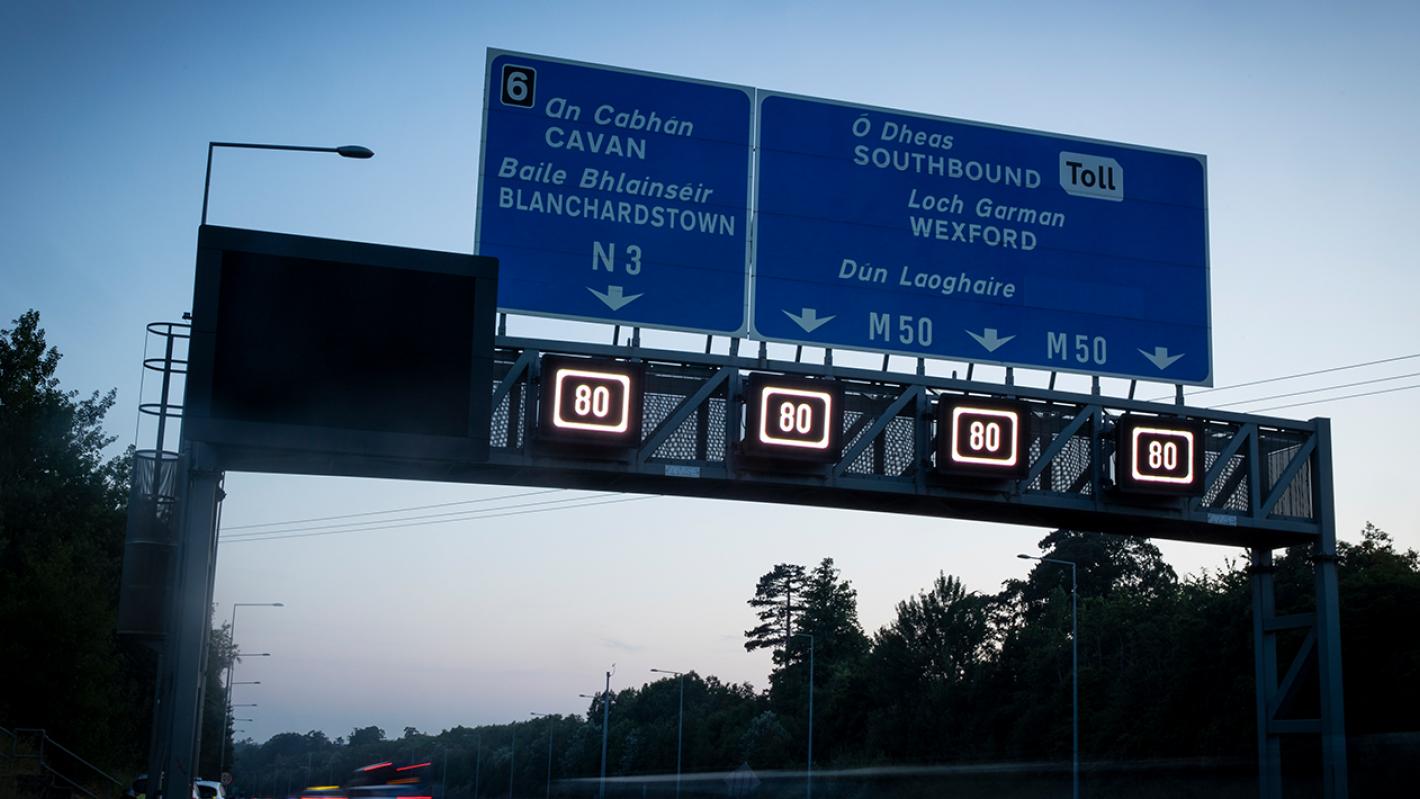}
	\caption{Speed advisory system on M50 highway traffic network in Dublin city. Optimal speeds are advised and shown on the screen for each lane on the highway traffic network\protect\footnote[1]{https://www.rod.ie/projects/enhancing-motorway-operation-services}.}
	\label{fig_sas}
\end{figure}

With the growing population in modern cities, traffic and transportation systems are becoming the most important infrastructure, supporting citizens for their daily commuting and travelling. Among the components of the traffic system, highway traffic networks provide the most efficient way to commute between different parts of cities, with a lower chance of traffic jams. The expanding use of highway traffic networks inevitably introduces new challenging problems of traffic management, such as the concern of safe driving, to avoid severe collisions by sudden unexpected accelerations, braking and lane change when the surrounding vehicles can not react  promptly. Given this background, Intelligent Transportation Systems (ITS) play an important role in solving traffic problems and ensuring traffic safety with fewer fatal traffic accidents \cite{calibaba2017road}. For instance, with the successful application of computer vision and network communication, such as a camera monitoring system, it is easy to track the moving vehicles with the image processing techniques and then various information (e.g., speed, number of vehicles on the road) are accessible via the appropriate application programming interface (APIs) \cite{mejia2021vehicle} \cite{nam2020deep}. With such information, variable speed limits and real-time speed advisory systems have been proposed to alleviate traffic congestion and maximise the utility of highway traffic networks in various aspects \cite{kuvsic2020extended, 7350149, liu2021mpc}. For instance, Fig \ref{fig_sas} demonstrates the speed advisory system (SAS) with variable speed limits deployed in Dublin city which is able to recommend optimal speeds for each lane on the M50 highway traffic network in Dublin city\footnote[1]{https://www.rod.ie/projects/enhancing-motorway-operation-services}.

\begin{figure}[htp]
	\centering
	\includegraphics[width=10.0 cm]{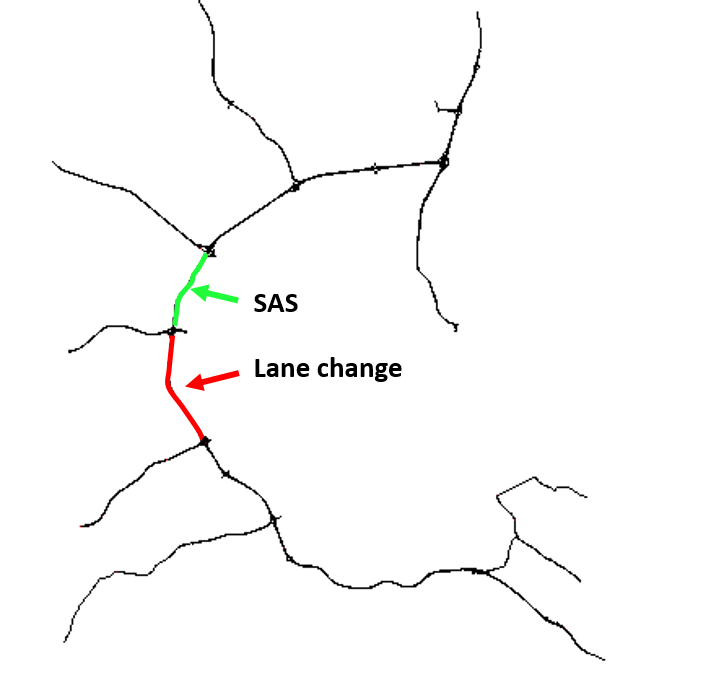}
	\caption{The simulated framework of the M50 highway network in Dublin city using SUMO. In the real world, The segment in green is implemented with a speed advisory system and the segment in red indicates where frequent lane changing behaviours may happen. }
	\label{fig_highway}
\end{figure}

However, even if the SAS system has been applied to govern driving speed and reduce the chances of traffic accidents \cite{li2013impacts}, drivers may drive with different driving intentions (i.e., acceleration; lane changing) unconsciously if they drive freely without traffic restriction \cite{jeon2014effects}. For instance, Fig. \ref{fig_highway} illustrates the framework of the M50 highway traffic network, where the SAS is implemented on the segment marked in green. Once vehicles leave the segment with SAS system (in green), it is more likely that vehicles may change lane freely and accidents may happen (in red). 


Although current work has shown the effectiveness of detecting the lane change in transportation systems using HMM \cite{li2016lane} and LSTM based methods \cite{tang2020driver, 8813987}, these methods can not leverage the natural geographical information (e.g, the connection between lanes) sufficiently.
The key difference between our work and theirs is that we detect the intention of lane changing based on GNN, in which the graph modelling can extract the spatial information between lanes and boost the detection performance. Existing works related to detecting the lane changing behaviours focus on vehicle-level detection \cite{mandalia2005using} \cite{7835731}. These works forecast whether a specific vehicle has an intention to change the lane while driving on the road, in order to avoid potential collisions. In this work, given the background that vehicles are driving at recommended speed on the highway traffic network, we detect the lane changing intentions using information collected from road-level rather than from individual vehicles, to indicate the chaotic level of the current road network such that different levels of traffic intervention may need to be applied. Regarding traffic network modelling, the previous works model the highway traffic network as a graph with the junctions as nodes and the roads as edges. We model the highway traffic network as a graph with the lanes as nodes and connectivity between lanes as edges, to extract the graph features with lane changing information, which will be discussed in section \ref{graph_modeling}. 
The main contributions of this chapter include:

\begin{itemize}
	
	\item [1.] We evaluate the performance of lane changing detection against different temporal segments, to investigate the efficiency of the detection algorithm. Results show that our method can detect lane changing intention in 90 seconds with higher accuracy comparing to HMM-based \cite{li2016lane} and LSTM-based method \cite{tang2020driver}, which can raise an alarm promptly in real-world applications.
	
	\item [2.] We apply temporal graph convolutional network with attention mechanism, to leverage the temporal information for accurate detection. In comparison with temporal convolutional neural network (TCNN), attention temporal graph convolutional network (ATGCN) shows the advantages in real-world application.
	
	\item [3.] Finally, for the purpose of interpreting our trained model, we calculate the standard deviation and spectral information divergence for the input features, to evaluate the contributions that the features make to the model. 
	
\end{itemize}

The remaining parts of the chapter are organised as follows. We introduce speed advisory system (SAS) on highway traffic networks as the background of this work and review some deep learning based detection for traffic flow in section \ref{relatework}. The experiment design, data processing and neural network architecture are demonstrated in section \ref{methodology}. Experimental results and further details regarding the results are discussed in section \ref{experimentch4}. Finally, we summarise this work in section \ref{conclusion}.

\section{Related works} \label{relatework}

\subsection{Speed advisory system}

With the development of ITS and vehicle-to-vehicle/infrastructure
(V2X) technologies, Intelligent Speed Advisory (ISA) systems have shown the capability in improving driving safety in various traffic scenarios \cite{hounsell2009review, tradisauskas2009map, gu2018design, chen2021intelligent, liu2015topics}. In highway traffic networks, in addition to driving safety, driving vehicles at the suggested speed has the benefits such as minimizing the emission, energy consumption and health risks \cite{7350149, gu2018design}. With this in mind, road operators and transportation departments can always monitor the speed of vehicles with the help of an intelligent camera-based platform \cite{mejia2021vehicle} to ensure that drivers follow the recommendation of the speed advisory system.

\subsection{Deep learning based traffic flow analysis}

A large body of work in the literature has been found using deep learning methods for traffic flow analysis. Most recently, deep belief networks \cite{huang2014deep}, autoencoder\cite{lv2014traffic} and recurrent neural network (RNN) based approaches \cite{tian2018lstm} have been implemented to analyse the sequential traffic flow data leveraging the long term temporal dependencies. Jointly working with sequential deep learning models, by segmenting the city into multiple areas and grids, CNN architectures with temporal units have been devised to access both spatial and temporal information where the traffic flow is processed into sequential 2-D data \cite{8526506} \cite{ma2017learning}. However, the above methods meet with common limitations for traffic flow analysis since they neglect the natural non-Euclidean property (e.g., graph) in road networks.

In general, traffic networks are naturally represented in graph format, where the roads are natural edges and connections between roads act as nodes. In order to overcome the significant limitation of the previous mentioned deep learning methods in traffic flow analysis, graph neural networks (GNNs) are applied as an ideal approach to data analysis on traffic networks since spatial dependencies between different nodes have been represented in graph structure. With the input of graph features, variants of GNN architectures have been proposed as the state-of-the-art approaches and obtained promising performances in various scenarios \cite{wu2020comprehensive} for detection problem. For instance, Diffusion Convolutional Recurrent Neural Network (DCRNN) \cite{li2018dcrnn_traffic}, Graph Wavenet \cite{wu2019graph} and spatial-temporal Graph convolutional network (STGCN) \cite{DBLP:conf/ijcai/YuYZ18} have been designed to leverage the spatial-temporal information and improve the traditional GNN architecture, which can boost the performance of data analysis in highway traffic networks. Tanwi et al. \cite{9413270} refined the DCRNN to transfer the common spatial-temporal information between cities with similar geographical structure to improve the detection performance. Yu et al. \cite{DBLP:conf/ijcai/YuYZ18} proposed STGCN to leverage the spatial and temporal dependencies between different areas of a city, to improve the performance of traffic demand forecasting.

\section{Methodology} \label{methodology}

\subsection{Simulation \& Experiment Design} 

In this section, the traffic flow influenced by different driving intentions is simulated using SUMO \cite{behrisch2011sumo}. SUMO is open-source software for the simulation of urban mobility, which is prevalent for the purposes of proposing and validating research ideas related to the intelligent transportation. In this work, we select a segment of highway traffic networks in Dublin city (i.e., the M50 highway network) as the scenario where different driving intentions may happen in the real world. As shown in red in Fig. \ref{fig_highway}, as the vehicles leave the green segment where the driving speed is guided by SAS, the drivers may drive with frequent lane changing intentions (in red segment) which endanger the traffic safety. There are four lanes in this segment of the M50 highway network and the data on traffic flow is collected while vehicles are running on this highway traffic network segment.

\begin{figure}[htp]
	\centering
	\includegraphics[width=7.0cm]{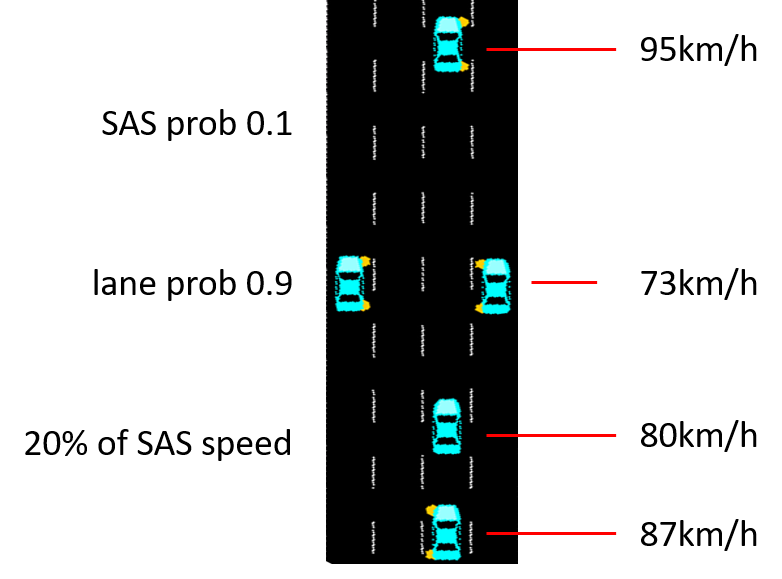}
	\caption{An example of the setting of driving intention, with setting SAS probability = 0.1, lane probability = 0.9 and $20\%$ of SAS speed. }
	\label{fig_setup}
\end{figure}

In this experiment, a new vehicle is generated per simulation step (i.e., 1 second) on the lane recommended by SUMO. In a normal situation, all vehicles are driving at SAS speed without frequent lane changes on the highway traffic network, where the SAS speed is set as 80 km/h. However, considering that different driving intentions could happen in the real world, we consider the possibility of violating SAS speed and frequent lane changing, when generating the traffic flow data. Violating SAS speed is defined as driving at a speed that is different from SAS speed in a given range (e.g, $5\%$, $10\%$, $20\%$ of SAS speed) and lane changing means that the vehicle randomly switches to any lane (i.e., four lanes including the current lane where the vehicle is currently driving on) of the highway traffic network. With this in mind, each vehicle has the possibility (i.e., SAS probability 0.1, 0.5 and 0.9) driving at SAS speed and the possibility (i.e., lane probability 0.1, 0.5 and 0.9) driving at the same lane at each simulation step while staying in the highway traffic network. The higher probability indicates that the vehicle has a higher chance to follow SAS speed and drive in the same lane. For instance, SAS probability = 1.0 and lane probability = 1.0 mean that the vehicle will drive at the SAS-recommended speed and will not change lane for the whole journey. Fig. \ref{fig_setup} demonstrates an example with setting SAS probability = 0.1, lane probability = 0.9 and $20\%$ of SAS speed. It indicates that each vehicle conducts uniform motion for the whole journey where the speed has 0.1 probability to set as SAS speed (i.e., 80 km/h) or has 0.9 probability to set as the speed from 64 km/h to 96 km/h (i.e., violating $20\%$ of SAS speed). Once the speeds are set, the vehicles are not able to change the speed. Each vehicle has 0.9 probability to drive at the current lane (i.e., 0.1 probability change the lane). In a real-world application, multiple cameras can be set to monitor the vehicles in each lane respectively. Since the driving speed of vehicles and the possibility that vehicles driving at SAS speed can be estimated, we detect the different lane changing intentions (i.e., different lane changing possibilities) in this chapter.

\subsection{Feature selection and model training} 
For feature selection, while vehicles are driving with intentions of speed and lane changing, the average driving speed and vehicle number on each lane are collected and estimated by the camera.  With this information, the road sector management unit can estimate not only the possibility that vehicles driving at SAS speed, but also the range of speed changes (e.g, $5\%$, $10\%$, $20\%$ of SAS speed) for the vehicles that do not follow the SAS speed. Therefore, the different models for lane changing detection will be trained on different SAS probability and range of speed changes. We label the traffic flow based on different probabilities of lane changing (i.e., lane probability). The traffic flow data used for model training, validation and testing are generated for 3600, 1800 and 3600 simulation steps, corresponding to monitoring the traffic flow for a period of one hour, half-an-hour and one hour respectively in real world.

\subsection{Traffic Flow on Graph} \label{graph_modeling}

\begin{figure}[htp]
	\centering
	\includegraphics[width=7.0cm]{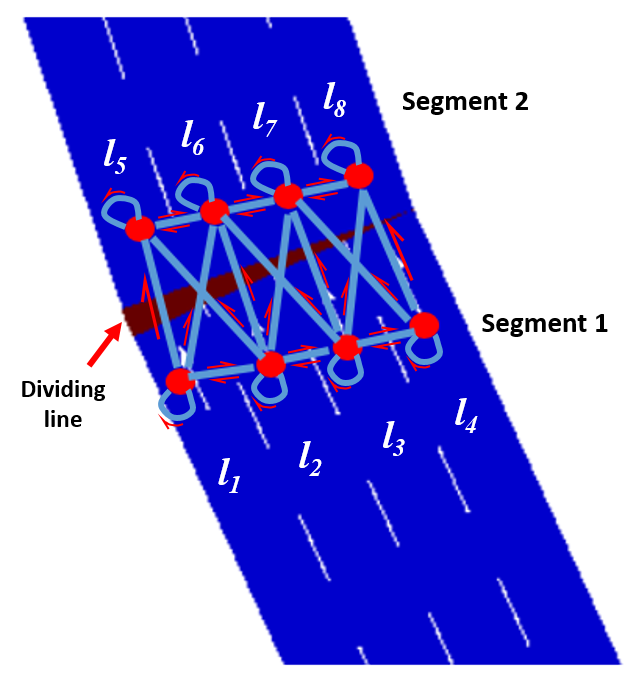}
	\caption{Directed graph modelling of a segment of the M50 highway network. Each node (red point) denotes a lane segment and the edges (blue lines) between nodes indicate adjacent lane segment. The vehicles can change to an adjacent lane segment or remain in the current lane segment.}
	\label{fig_graph}
\end{figure}

In this section, we introduce the processing of highway traffic flow with graph modelling. In previous works, the traffic flow data is collected at the junctions between different roads. However, there are multiple lanes on each road and we collect the lane-wise traffic flow data. With this setting, we treat the highway network as a graph $ G = (V,E) $, where $V$ denotes the nodes which is the set of lane segments $V = \{l_i|i = 1,2,\dots, N\}$, $E$ denotes the edges which is the connections between nodes. The adjacency matrix derived
from a graph is denoted by $A$. The connectivity of the graph is set as fully connected as the vehicle may change lanes from one to any other while driving with lane changing intentions, indicating $A_{i,j} = 1$ for $i,j = 1,2,\dots, N$. Specifically, as shown in Fig.\ref{fig_graph}, the highway network is divided into two segments therefore we have 8 lane segments (i.e., $N = 8$) and graph signal $X(t) = (S(t),D(t))$ is collected at each simulation step $t$ among different nodes, where $S$ denotes the averaged vehicle speed and $D$ the number vehicle (i.e., density) on the lane. Finally, $X(t)$ for $t = 1,2,\dots, T$ is processed as a sample of graph data, where $T$ indicates the length of the temporal segment when processing the graph data.


\subsection{Network architecture}

\begin{figure}[htp]
	\centering
	\includegraphics[width=10.0 cm]{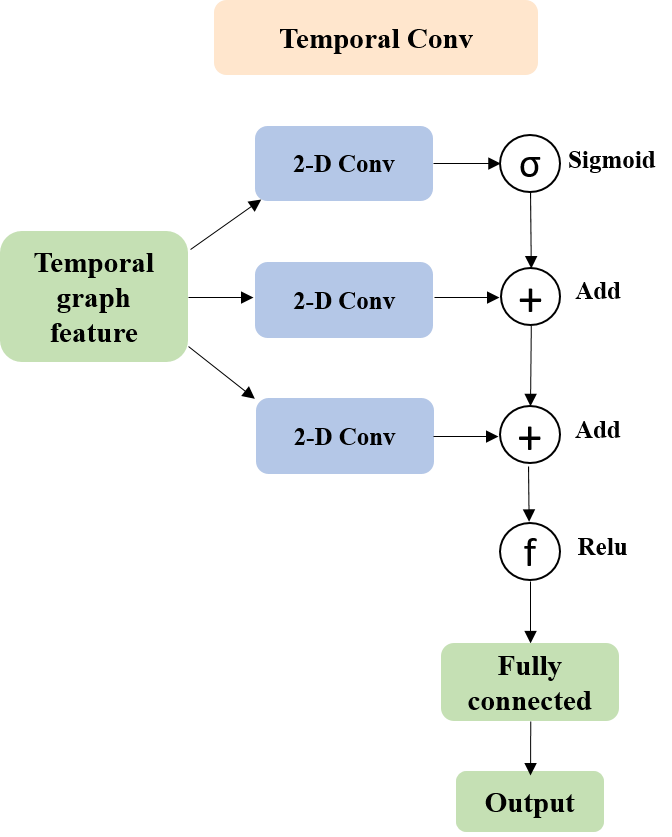}
	\caption{An architecture of a temporal convolutional networks based on \cite{DBLP:conf/ijcai/YuYZ18}. }
	\label{fig_tcnn_network}
\end{figure}

\begin{figure}[htp]
	\centering
	\includegraphics[width=10.0 cm]{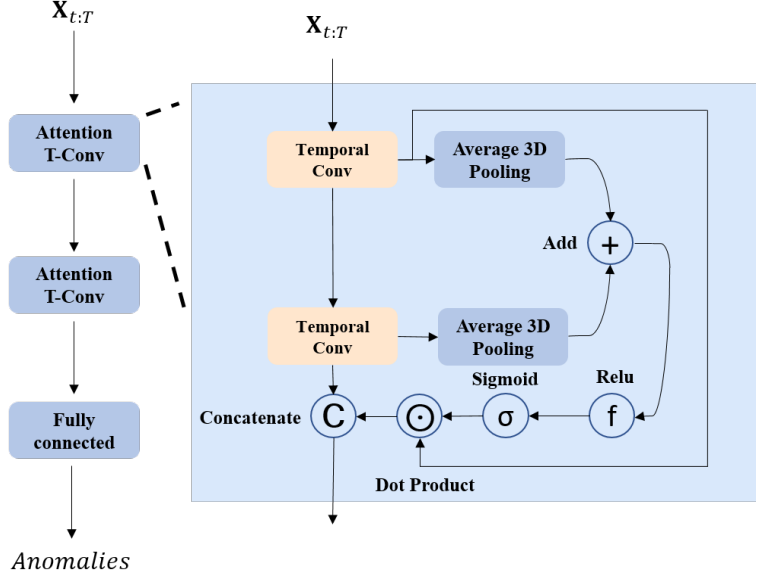}
	\caption{Architecture of attention temporal graph convolutional networks. }
	\label{fig_gcn_network}
\end{figure}

\begin{itemize}
	
	\item Temporal convolutional networks (TCNN). With the graph modelling in highway traffic networks, TCNN is designed as a baseline, to evaluate the ability of CNN in detecting the intentions given graph-traffic data flow. As simple as possible, the architecture of TCNN is refined from \cite{DBLP:conf/ijcai/YuYZ18} and demonstrated in Fig. \ref{fig_tcnn_network}. Graph features extracted from each temporal segment are conveyed to three identical 2-D convolutional layers. The output from the first convolutional layer is activated by a sigmoid function to have normalised values between 0 and 1. Output from the other two convolutional layers is added with normalised values and then activated by a Relu function, followed by a fully-connection layer. This setting considers that two convolutional layers without sigmoid activation tune the model parameters in general, converging to the optimal values faster, while normalised values from the first convolutional layer with sigmoid activation can help to adjust the parameters precisely.

	\item Temporal graph convolutional networks with attention mechanism (ATGCN). Based on the temporal convolutional networks proposed above, we extend the network architecture to graph convolutional networks with  an attention component. Referring to the work presenting the ST-GCN architecture \cite{DBLP:conf/ijcai/YuYZ18}, we introduce TGCN with attention mechanism, consisting of two attention temporal convolution blocks (ATCs) and a fully-connected output layer. Each ATC consists of two temporal convolution blocks used in TCNN, with attention mechanism applied to process temporal information, as demonstrated in Fig. \ref{fig_gcn_network}. Note that ATGCN has the latent static spatial information since the nodes are fully connected to each other as discussed in section \ref{graph_modeling}.

\end{itemize}

%
%
%
%

\subsection{Network setup}

Referring to the processing traffic flow into graph format in section \ref{graph_modeling}, the number of nodes is set as 8, indicating 8 lane segments in highway traffic network. The number of features is set as 2, corresponding to the average speed and number of vehicles on each lane segment collected as the graph features. The length of temporal segment $T$ is set to 30, 60 and 90 respectively, which will be examined later by our algorithm.

For TCNN architecture, each convolution layer has 2 input channels (e.g., corresponding to the number of features) and 64 output channels, with the kernel size set to 3.  In each input channel, a 2-D traffic data slice with a dimension of [$T$, 8], indicating the specific feature from the 8 lanes in a given temporal segment, is used for the model training. Therefore, the fully-connected layer receives the input size as [$T$ x 8 x 64] and the output size as 3, corresponding to the 3 categories of anomalies that will be discussed in section \ref{anomalies}.  We set the batch size as 32 indicating there are 32 2-D traffic data slices for each training iteration. 

The ATGCN architecture shares the same setting with TCNN architecture related to the part of the temporal convolution layer and fully-connected layer. The averaged 3-D pooling operator processes the data along with the dimension of $T$, with the output vector with a size of [1,$T$]. This vector conducts dot product operation with the output of the temporal convolution module, realizing the attention effect on temporal information. Table \ref{tabsettings} lists the common settings when training the ATGCN and TCNN architecture.

\begin{table}[htbp]
	\caption{Network settings for ATGCN and TCNN} 
	\begin{center}
		\begin{tabular}{|l|c|c| }
			\hline
			\textbf{Parameters} & \textbf{Values}\\
			\hline
			Nodes  & 8 (only for ATGCN)  \\
			Length of temporal segment  & 30, 60, 90\\
			Intention categories & 3 \\
			Feature dimension & 2 \\
			Batch size & 32 \\
			Initial learning rates & 0.001 \\
			Optimiser & Adam algorithm \\
			Weight decay & 0.001 \\
			
			\hline
		\end{tabular}
	\end{center}
	\label{tabsettings}
\end{table}


\section{Experimental Results and Discussion} \label{experimentch4}
In this section, we analyse and discuss the results of detection for lane changing intention when the vehicles were driven under irregular speeds.

%
%
%
%

\subsection{Lane changing detection} \label{anomalies}

\begin{table}[h]
	\caption{Accuracy of detection for lane changing intention from ATGCN}
	\label{table_gcn_lane}
	\begin{center}
		\begin{tabular}{|c|c|c|c|c|}
			\hline
			Speed change & Conditions & T=30 & T=60 & T=90\\
			\hline
			5\% of SAS speed & SAS prob = 0.1 & 90.85\%& 96.00\%& 97.39\%\\
			& SAS prob = 0.5 & 86.62\%& 90.57\%& 94.78\%\\	
			& SAS prob = 0.9 & 98.17\%& 98.86\%& 100.00\%\\
			& Average & 91.88\%& 95.14\%& \textbf{97.39}\%\\
			\hline
			
			10\% of SAS speed& SAS prob = 0.1 & 91.69\%& 95.71\%& 97.39\%\\
			& SAS prob = 0.5& 90.42\%& 97.14\%& 98.26\%\\
			& SAS prob = 0.9 & 97.61\%& 98.86\%& 97.83\%\\
			& Average & 93.24\%& 97.24\%& \textbf{97.83}\%\\
			\hline
			
			20\% of SAS speed& SAS prob = 0.1 & 96.06\%& 98.00\%& 99.13\%\\
			& SAS prob = 0.5 & 95.07\%& 98.29\%& 100.00\%\\
			& SAS prob = 0.9 & 96.20\%& 98.86\%& 99.13\%\\
			& Average & 95.78\%& 98.38\%& \textbf{99.42}\%\\

			\hline
		\end{tabular}
	\end{center}
\end{table}

\begin{table}[h]
	\caption{Accuracy of detection for lane changing intention from TCNN}
	\label{table_cnn_lane}
	\begin{center}
		\begin{tabular}{|c|c|c|c|c|}
			\hline
			Speed change & Conditions & T=30 & T=60 & T=90\\
			\hline
			5\% of SAS speed & SAS prob = 0.1 & 91.27\%& 96.29\%& 90.43\%\\
			& SAS prob = 0.5 & 86.06\%& 91.71\%& 89.57\%\\	
			& SAS prob = 0.9 & 98.59\%& 98.86\%& 97.83\%\\
			& Average & 91.97\%& \textbf{95.62}\%& 92.61\%\\
			\hline
			
			10\% of SAS speed& SAS prob = 0.1 & 92.54\%& 97.14\%& 94.35\%\\
			& SAS prob = 0.5& 93.52\%& 96.86\%& 97.39\%\\
			& SAS prob = 0.9 & 97.75\%& 98.00\%& 96.96\%\\
			& Average & 94.60\%& \textbf{97.33}\%& 96.23\%\\
			\hline
			
			20\% of SAS speed& SAS prob = 0.1 & 96.48\%& 98.00\%& 98.26\%\\
			& SAS prob = 0.5 & 95.49\%& 98.00\%& 96.96\%\\
			& SAS prob = 0.9 & 96.76\%& 96.29\%& 99.13\%\\
			& Average & 96.24\%& 97.43\%& \textbf{98.12}\%\\

			\hline
		\end{tabular}
	\end{center}
\end{table}

Here we evaluate the deep learning algorithms for detecting lane changing intentions. In order to exclude the effect of speed violation when detecting the traffic flow caused by lane changing, the data is divided under three conditions, that is data generated under possibilities (i.e., 0.1, 0.5 and 0.9) of speed violation. Detection for intentions of lane changing is investigated in these conditions separately.

The detection also considers the effect of temporal segments when processing the graph data. We select three temporal segments with different lengths (i.e., $T = 30, 60, 90$) when generating the sample of graph data. With these settings, the algorithm detects the lane changing intention every 30, 60 and 90 seconds respectively in a real-world application. Every two contiguous samples have an overlap time of $T/2$ given the specific length of temporal segment $T$.

Table \ref{table_gcn_lane} and Table \ref{table_cnn_lane} demonstrate the results of lane changing detection using ATGCN and TCNN respectively. On the one hand, the averaged accuracies based on ATGCN are better than that based on TCNN for different ranges of speed change. For each category of speed change, the detection based on ATGCN obtains the highest averaged accuracy given the length of temporal segment $T = 90$, which outperforms the performances of TCNN. For instance, ATGCN achieves the highest accuracy $99.42\%$ and TCNN obtains accuracy $98.12\%$ given $T = 90$.  On the other hand, the length of a temporal segment when processing the graph data has an important impact on detecting the traffic flow caused by lane changing. For ATGCN, as the length of temporal segments become longer, the performance of detecting traffic flow caused by lane-changing behaviours gets better for most conditions when the speed is changed in different ranges. For instance, when most vehicles violate the SAS speed (i.e., when vehicles have only a 0.1 probability to follow SAS speed), expanding the length of the temporal segment from $T=30$ to $T=90$ increases the averaged detection accuracy from $96.06\%$ to $99.13\%$ when the speed is changed within the range of $20\%$ of SAS speed. The averaged accuracy among all conditions also increase in line with the increase of the length of temporal segments. For TCNN, extending the length of the temporal segment can enhance the algorithm's chances of detecting the lane-changing behaviour.
The performances under $T = 60$ and $T = 90$ are better than that of under $T = 30$ and the detection obtains the best performance (i.e., accuracy $98.12\%$) under $T = 90$ when the speed is changed within the range of $20\%$ of SAS speed.

The average accuracies from Table \ref{table_gcn_lane} and Table \ref{table_cnn_lane} also indicate that the detection is getting more accurate as the range of speed change gets larger for both ATGCN and TCNN. When the vehicles are driving at large different speeds, the lane changing behaviours can cause a disturbance in traffic flow and lead to higher risks of traffic accidents. Therefore, given the larger range of speed change, the detection algorithm can catch the information representing the lane changing intentions easier. Even if the vehicles are driving at the most similar speed (i.e., only violate SAS speed within a range of $5\%$) where the chance of traffic accident is smaller, the ATGCN can also detect the corresponding lane changing behaviour with an accuracy of $97.39\%$.

\begin{center}
	\begin{figure*}[ht]
		\centering
		\includegraphics[width= 6.0in]{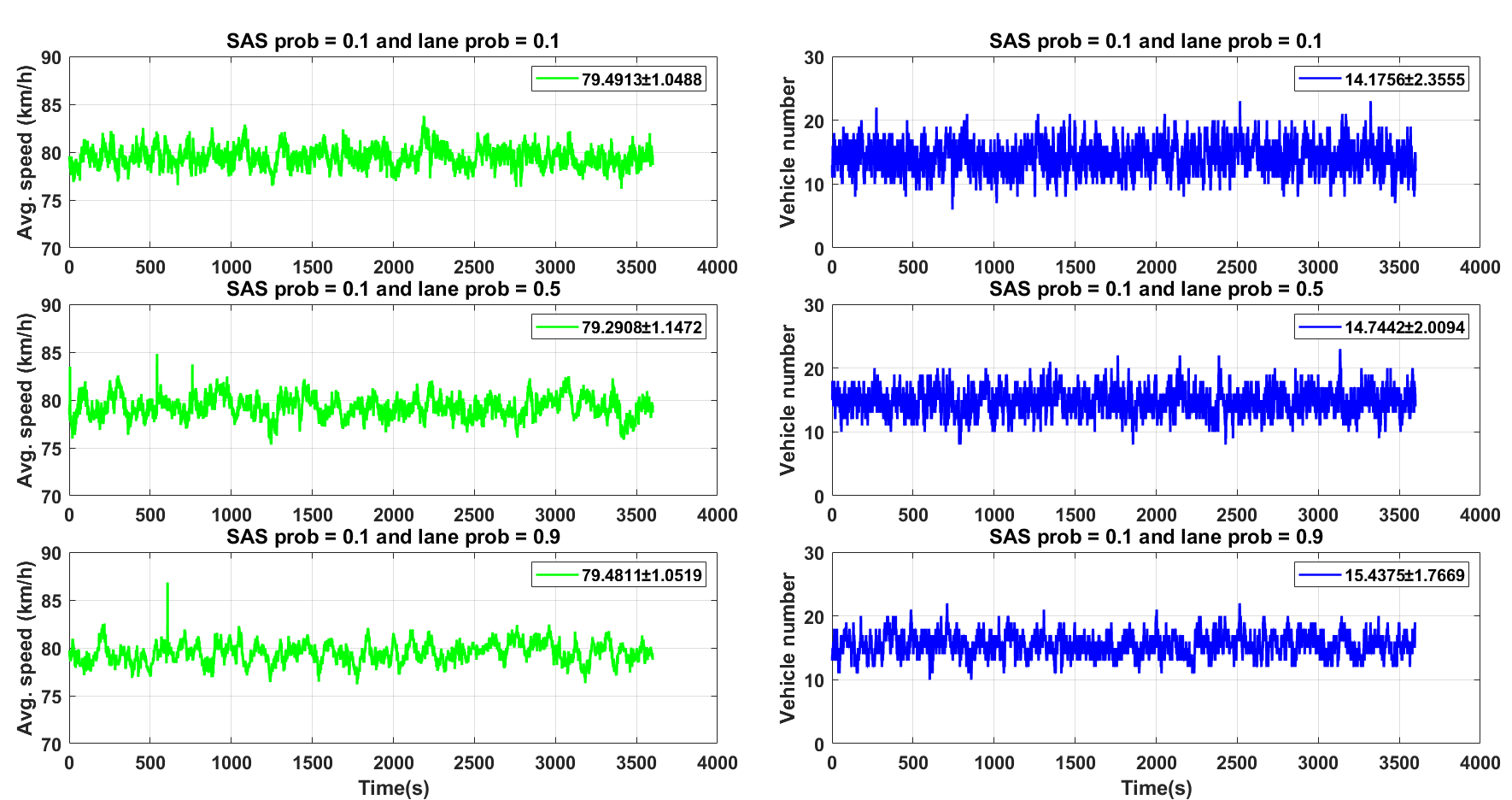}
		\caption{The average speed and vehicle number in lane 4 (node $l_4$) among different lane changing probabilities under conditions SAS prob = 0.1 and $10\%$ of SAS speed change. The legends indicate the mean and standard deviation for the corresponding feature.}
		\label{fig_SAS01}
	\end{figure*}	
\end{center}

\begin{center}
	\begin{figure*}[ht]
		\centering
		\includegraphics[width= 6.0in]{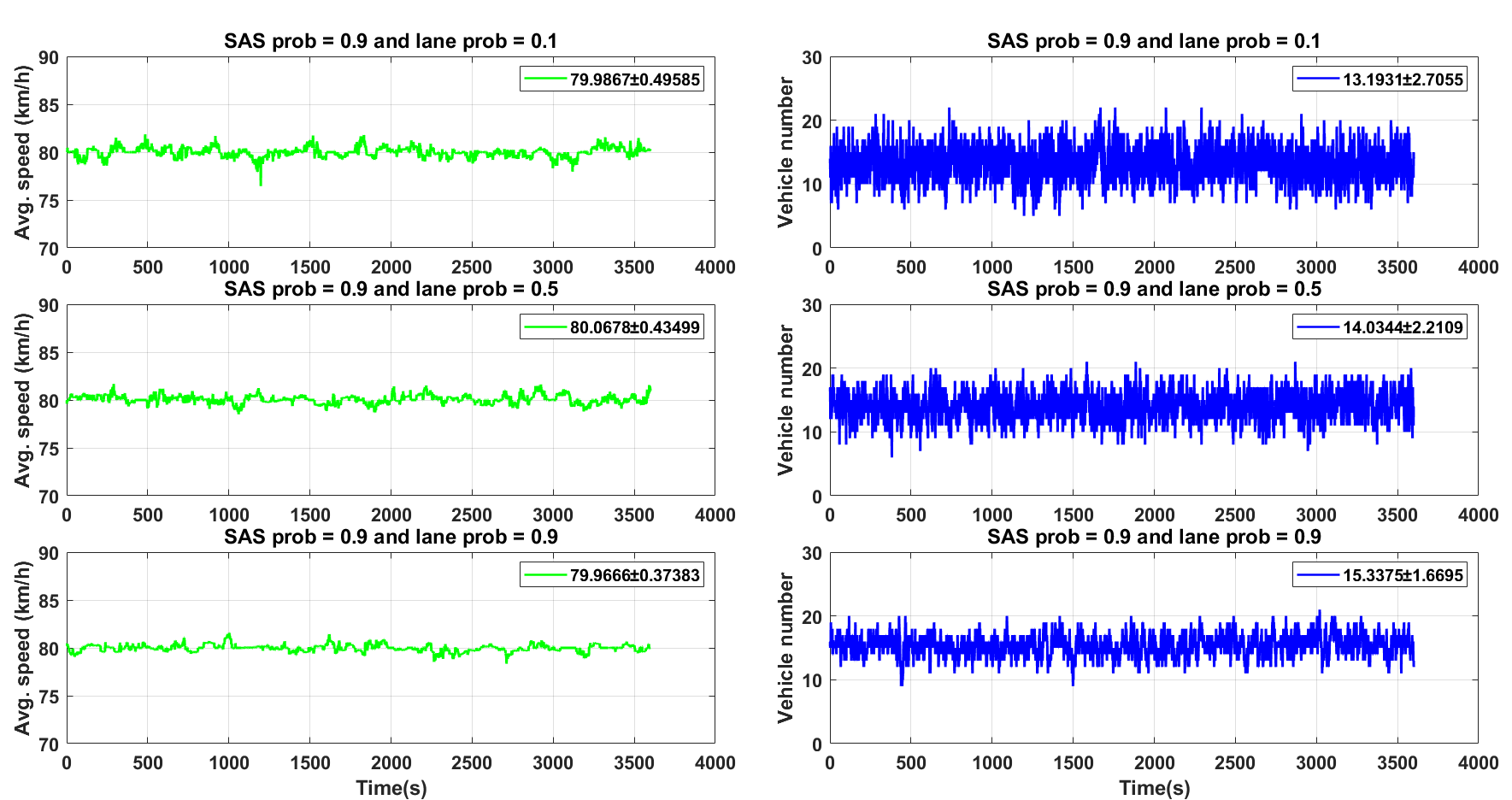}
		\caption{The average speed and vehicle number in lane 4 (node $l_4$) among different lane changing probabilities under conditions SAS prob = 0.9 and $10\%$ of SAS speed change. The legends indicate the mean and standard deviation for the corresponding feature.}
		\label{fig_SAS09}
	\end{figure*}	
\end{center}



\subsection{Feature visualisation and analysis} \label{data_visualization}

In this section, the features (i.e., average speed and vehicle number on the lane) are visualised and the importance of these features are discussed for lane changing detection. 

Fig. \ref{fig_SAS01} and Fig. \ref{fig_SAS09} demonstrate the average speed and vehicle number in lane 4 for an hour, among different lane changing intention, where the vehicles have the probability of 0.1 and 0.9 to follow the SAS speed respectively. Table \ref{Std} shows the statistical mean and standard deviation of the corresponding features. For both situations where SAS prob = 0.1 and SAS prob = 0.9, as vehicles tend to drive without lane change intentions (i.e., the lane probability increases), the standard deviations of vehicle number get smaller, which indicates the sequential feature of vehicle number tend to be more stable and this pattern can be caught by the prediction models. As for the average speed on the lane, the standard deviation changes slightly without a clear trend and the values of standard deviations are close to each other, while the lane probability increases. This pattern indicates that the vehicle number plays a dominant role in lane change detection even if the average speed has reflections to lane changing intentions. 

For the purpose of comparing the similarity of the features between different lane probabilities in the frequency domain, the spectral information divergence (SID) measurements \cite{773549} are calculated between average speeds and between vehicle numbers. A higher value of SID indicates the two signals are more different with respect to the spectrum pattern. Table \ref{SID} shows the SID measurements for average speeds and vehicle numbers between different lane probabilities under conditions SAS prob = 0.9 and $10\%$ of SAS speed change. The SID values for the feature of vehicle numbers are tremendously larger than that for average speeds, indicating that the patterns shown in vehicle numbers are more specific given the corresponding lane probability and provide crucial information for lane changing detection. 

\begin{table}[h]
	\caption{Standard deviation for features in different lane probability (LP)}
	\label{Std}
	\begin{center}
		\begin{tabular}{|c|c|c|c|c|}
			\hline
			Conditions & Features & LP = 0.1 & LP = 0.5 & LP = 0.9\\
			\hline
			SAS prob = 0.1 & Avg. Speed & 1.04 & 1.14& 1.05\\
			&Vehicle Num.	& 2.35 & 2.00& 1.76\\	
			\hline
			SAS prob = 0.9 & Avg. Speed & 0.49 & 0.43& 0.37\\
			&Vehicle Num.	& 2.71 & 2.21& 1.67\\	
			\hline

		\end{tabular}
	\end{center}
\end{table}

\begin{table}[h]
	\caption{Spectral information divergence for features between different lane probability (LP)}
	\label{SID}
	\begin{center}
		\begin{tabular}{|c|c|c|c|}
			\hline
			Features & LP=0.1 vs. LP=0.5 & LP=0.1 vs. LP=0.9 & LP=0.5 vs. LP=0.9\\
			\hline
			Avg. Speed & 46.35 & 49.31& 43.97\\
			Vehicle Num.	& 217.79 & 224.72& 284.41\\	
			\hline

		\end{tabular}
	\end{center}
\end{table}

\section{Conclusions} \label{conclusion}

In this chapter, we model the traffic flow data on highway traffic networks using graph and leverage temporal graph convolutional network architecture embedded with attention mechanism to detect vehicles lane changing intentions. The experiments compare the detection performance of ATGCN with that of TCNN. Comparison results indicate that the attention mechanism enhances the ability in capturing the key temporal information and improves the detection accuracy. With ATGCN, anomalies can be detected within 90 seconds with the highest accuracy and this prompt detection is important for traffic condition monitoring. In addition, with graph data modelling for highway traffic networks, a simpler TCNN architecture can also detect vehicles lane changing intentions accurately if sufficient information is provided by a larger temporal segment. In fact, TCNN is also a promising alternative with shorter time window (e.g., 30- and 60-second window) calculation, but a longer time window with ATGCN can achieve better performance/accuracy.

To conclude, we believe that the chapter releases implications for intelligent transportation: 1) Graph modelling on traffic flow suits the nature of highway networks and helps to enhance the knowledge representation. 2) The length of the temporal segment affects the performance of anomaly detection. When anomalies are required to be detected accurately and rapidly in important segments of highway traffic networks, delicate models (e.g., ATGCN) deserve more consideration. On the contrary, if anomalies will not cause severe threats to the driving safety (e.g., the driving speed varies within a small range) and can be monitored infrequently, the simpler model (TCNN) can be applied to reduce the computation cost.

\chapter{Conclusion}
In this chapter, we conclude the thesis by highlighting our contributions, noting any limitations of the work and suggesting areas for future investigation. The chapter is organised as follows: section \ref{Thesissummary} summarises the key findings of the thesis with respect to each technical chapter; section \ref{Limitationandfuture} points out limitations and potential future work.

\section{Thesis summary}\label{Thesissummary}

In this thesis, we discuss three topics related to IoT and smart transportation. In chapter 2, we investigate the problem on maximising the overall utility of IoT networks in a secure, privacy-aware and plug-and-play manner. For achieving this objective, we assume that there are different priority levels when different IoT devices transmit data to the central node in a decentralised setup with limited system resources. We propose a transmission frequency management system with anomaly detection mechanisms to better manage the IoT networks. Also, we introduce the system architecture including four key components: IoT devices, Gateway, Cloud platform and Users. Each IoT device is associated with a utility function with certain assumptions, and our objective is maximise the overall utility for the group of devices in the network by solving a mathematical optimization problem. Applying decentralised ADMM optimisation, the transmission frequency management is able to allocate the optimal transmission frequency to each IoT device in a privacy-protected manner. We also discuss anomaly detection in different scenarios using both mathematical rule-based and an LSTM-based approaches. In real-world experiments, the optimal transmission frequencies are calculated and set locally on each IoT device, without the allocation from central node. Meanwhile, manipulations that lead the IoT devices to transmit data deviating the set transmission frequencies can be detected by the proposed anomaly detector with high accuracy.

In chapter 3, we investigate the problem of sharing bike availability. Based on the current research related to traffic demand prediction, we leverage the state-of-the-art spatial-temporal graph convolutional network (ST-GCN) as the foundation to approach the research objective, to predict the number of available sharing bikes using realistic datasets. To enhance the prediction accuracy, we embed ST-GCN architecture with an attention module (AST-GCN) to leverage spatial and temporal information with different focuses. Furthermore, we also discuss the impacts of different modelling methods of adjacency matrices on the proposed architecture. Experimental results show that our proposed method using AST-GCN with the embedded adaptive adjacency matrix outperforms the majority of existing approaches in two real-world datasets.

In chapter 4, we consider the problem of detecting the lane changing intention on highway traffic networks for improving driving safety. We define the lane changing intention as lane changing probability and then simulate the traffic flow with a group of vehicles drive at different lane changing probabilities using a popular mobility simulator (i.e., SUMO). Given the simulation scenario, we leverage temporal graph convolutional network with attention module (ATGCN) to detect the lane changing intentions and compare the performance with another concise algorithm, i.e., temporal convolutional neural network (TCNN). Experiment results show that ATGCN can detect the lane changing intentions within 90 seconds with higher accuracy, while the TCNN can also detect the lane changing intentions quite accurately with just $1.3 \%$ lower accuracy compared to ATGCN. In a word, there is a trade-off between detection performance and the computation resources. If the computational resource is limited in the IoT network, TCNN can play as a computation-economic role to ensure the driving safety; While there is enough resource for computation, ATGCN is the better option for detecting the lane changing intention.

In general, the thesis investigates how the advanced optimisation theories and novel machine learning methods can be applied to deal with real-world challenges arising in several research areas of IoT and smart transportation.

%
%
%

\section{Limitations and Future works}\label{Limitationandfuture}

The thesis discusses different topics related to the deep learning and optimisation algorithm applied in IoT and smart transportation. During the research carried out in this thesis, some limitations which merit further improvement arise and we now revisit these in our future work.

In chapter 2, the transmission frequency management system allocates the optimal transmission frequencies in order to maximise the overall utility of a group of IoT devices. The utility functions defined on IoT devices are strictly concave and smooth. However, in some scenarios where the utility functions are non-smooth and non-concave, the system behaviours become different and we will investigate the dynamics of system behaviours given the non-smooth and non-concave utility functions defined on IoT devices. Regarding the anomaly detection in transmission frequency management system, we only employ the LSTM architecture and there is a lack of investigations using other deep learning methods. The future work will experiment with other deep learning algorithms (e.g., graph neural network \cite{deng2021graph, zhao2020multivariate} and Transformer based architecture \cite{huang2020hitanomaly}). We will also experiment with different topologies (e.g., partially connected topology \cite{jun2010partial}) that will be applied to the IoT network to model the connection relationship between devices and the gateway. As for anomaly detection, we only simulate the scenario when only one device suffers one type of malicious manipulation at the same time. In future work, we will investigate cases when a device suffers attacks by multiple manipulations at the same time and we will also consider the scenario where there are new devices connecting to or disconnecting from the gateway.

In chapter 3, although the overall accuracy of bike availability prediction is low for all stations, the prediction errors are relatively higher for the stations in the city centre area. In order to improve the prediction performance in the city centre, the network structure, adjacency matrices and advanced feature selection will be investigated as part of our future work.

In chapter 4, lane changing intention is predicted when the vehicles are guided by only a specific SAS speed (e.g., SAS speed = 80 km/h). In future work, it is worth investigating the prediction performance when the vehicles are driving at different SAS-recommended speeds. The lane changing intention is described as different static probabilities (e.g., probability 0.1, 0.5, 0.9). However, in the real world, it is more complicated to describe the intention, since the probability of lane changing can be varied depending on drivers' characteristics. In future work, we shall factor in such complexity in modeling drivers for a more accurate analysis for real world scenarios. Also, the real driver behaviours (e.g., the lane changing behaviours in real world) would be investigated, in order to figure out what level of detection accuracy would be enough for real-world applications. Finally, we wish to note that the SAS system will cover all parts of M50 highway network in the near future. In this context, different attributes of the road segments, e.g., length of lanes, number of lanes, are required to be redesigned for a better modelling of the graph, which forms another part of our future work.


\bibliographystyle{ieeetr}
\bibliography{references} 

\begin{thebibliography}{100}

\bibitem{DBLP:conf/ijcai/YuYZ18}
B.~Yu, H.~Yin, and Z.~Zhu, ``Spatio-temporal graph convolutional networks: {A}
  deep learning framework for traffic forecasting,'' in {\em Proceedings of the
  Twenty-Seventh International Joint Conference on Artificial Intelligence,
  {IJCAI} 2018, July 13-19, 2018, Stockholm, Sweden}, pp.~3634--3640,
  ijcai.org, 2018.

\bibitem{ye2020coupled}
J.~Ye, L.~Sun, B.~Du, Y.~Fu, and H.~Xiong, ``Coupled layer-wise graph
  convolution for transportation demand prediction,'' {\em arXiv preprint
  arXiv:2012.08080}, 2020.

\bibitem{giusto2010internet}
D.~Giusto, A.~Iera, G.~Morabito, and L.~Atzori, {\em The internet of things:
  20th Tyrrhenian workshop on digital communications}.
\newblock Springer Science \& Business Media, 2010.

\bibitem{madakam2015internet}
S.~Madakam, V.~Lake, V.~Lake, V.~Lake, {\em et~al.}, ``Internet of things
  {(IoT)}: A literature review,'' {\em Journal of Computer and Communications},
  vol.~3, no.~05, p.~164, 2015.

\bibitem{jia2012rfid}
X.~Jia, Q.~Feng, T.~Fan, and Q.~Lei, ``{RFID technology and its applications in
  Internet of Things (IoT)},'' in {\em 2012 2nd international conference on
  consumer electronics, communications and networks (CECNet)}, pp.~1282--1285,
  IEEE, 2012.

\bibitem{2011A}
H.~Chen, X.~Jia, and H.~Li, ``A brief introduction to {IoT} gateway,'' in {\em
  IET international conference on communication technology and application
  (ICCTA 2011)}, pp.~610--613, IET, 2011.

\bibitem{2018Brokering}
C.~S. Singh, E.~S. Pilli, R.~C. Joshi, S.~Girdhari, and M.~C. Govil,
  ``Brokering in interconnected cloud computing environments: A survey,'' {\em
  Journal of Parallel and Distributed Computing}, pp.~S0743731518305719--,
  2018.

\bibitem{2018Economic}
M.~Nir, A.~Matrawy, and M.~St-Hilaire, ``Economic and energy considerations for
  resource augmentation in mobile cloud computing,'' {\em IEEE Transactions on
  Cloud Computing}, pp.~1--1, 2018.

\bibitem{St2016Cloud}
M.~Aazam, E.-N. Huh, M.~St-Hilaire, C.-H. Lung, and I.~Lambadaris, ``Cloud
  customer's historical record based resource pricing,'' {\em IEEE Transactions
  on Parallel and Distributed Systems}, vol.~27, no.~7, pp.~1929--1940, 2015.

\bibitem{li2012compressed}
S.~Li, L.~Da~Xu, and X.~Wang, ``Compressed sensing signal and data acquisition
  in wireless sensor networks and internet of things,'' {\em IEEE Transactions
  on Industrial Informatics}, vol.~9, no.~4, pp.~2177--2186, 2012.

\bibitem{he2012integration}
W.~He and L.~Da~Xu, ``Integration of distributed enterprise applications: A
  survey,'' {\em IEEE Transactions on industrial informatics}, vol.~10, no.~1,
  pp.~35--42, 2012.

\bibitem{9301390}
J.~{Ni}, K.~{Zhang}, and A.~V. {Vasilakos}, ``Security and privacy for mobile
  edge caching: Challenges and solutions,'' {\em IEEE Wireless Communications},
  pp.~1--7, 2020.

\bibitem{9163078}
A.~Alwarafy, K.~Al-Thelaya, M.~Abdallah, J.~Schneider, and M.~Hamdi, ``A survey
  on security and privacy issues in edge-computing-assisted internet of
  things,'' {\em IEEE Internet of Things Journal}, vol.~8, no.~6,
  pp.~4004--4022, 2020.

\bibitem{10.1117/12.2571307}
M.-A. Sachian, G.~Suciu, F.~Osiac, R.~Roșcăneanu, and R.~Streche,
  ``{Cyber-physical healthcare security system based on a Raspberry Pi},'' in
  {\em Advanced Topics in Optoelectronics, Microelectronics and
  Nanotechnologies X} (M.~Vladescu, R.~D. Tamas, and I.~Cristea, eds.),
  vol.~11718, pp.~517 -- 525, International Society for Optics and Photonics,
  SPIE, 2020.

\bibitem{8230004}
A.~{Kaur} and A.~{Jasuja}, ``Health monitoring based on {IoT} using {Raspberry
  PI},'' in {\em 2017 International Conference on Computing, Communication and
  Automation (ICCCA)}, pp.~1335--1340, 2017.

\bibitem{9287960}
M.~{Verucchi}, L.~{Bartoli}, F.~{Bagni}, F.~{Gatti}, P.~{Burgio}, and
  M.~{Bertogna}, ``Real-time clustering and lidar-camera fusion on embedded
  platforms for self-driving cars,'' in {\em 2020 Fourth IEEE International
  Conference on Robotic Computing (IRC)}, pp.~398--405, 2020.

\bibitem{9289509}
Y.~{Yang}, J.~{Kim}, J.~{Lee}, S.~{Lee}, S.~{Hong}, and S.~H. {Shah Newaz},
  ``Adaptive queue management in embedded edge devices for object detection
  with low latency,'' in {\em 2020 International Conference on Information and
  Communication Technology Convergence (ICTC)}, pp.~1426--1428, 2020.

\bibitem{batty2012smart}
M.~Batty, K.~Axhausen, F.~Giannotti, A.~Pozdnoukhov, A.~Bazzani, M.~Wachowicz,
  G.~Ouzounis, and Y.~Portugali, ``Smart cities of the future,'' {\em The
  European Physical Journal Special Topics}, vol.~214, no.~1, pp.~481--518,
  2012.

\bibitem{albino2015smart}
V.~Albino, U.~Berardi, and R.~M. Dangelico, ``Smart cities: Definitions,
  dimensions, performance, and initiatives,'' {\em Journal of urban
  technology}, vol.~22, no.~1, pp.~3--21, 2015.

\bibitem{lin2017intelligent}
Y.~Lin, P.~Wang, and M.~Ma, ``Intelligent transportation system {(ITS)}:
  Concept, challenge and opportunity,'' in {\em 2017 IEEE 3rd international
  conference on big data security on cloud (bigdatasecurity), IEEE
  international conference on high performance and smart computing (HPSC), and
  IEEE international conference on intelligent data and security (IDS)},
  pp.~167--172, IEEE, 2017.

\bibitem{khanna2016iot}
A.~Khanna and R.~Anand, ``{IoT} based smart parking system,'' in {\em 2016
  International Conference on Internet of Things and Applications (IOTA)},
  pp.~266--270, IEEE, 2016.

\bibitem{agarana2017minimizing}
M.~Agarana, S.~Bishop, and O.~Agboola, ``Minimizing carbon emissions from
  transportation projects in sub-saharan africa cities using mathematical
  model: A focus on lagos, nigeria,'' {\em Procedia Manufacturing}, vol.~7,
  pp.~596--601, 2017.

\bibitem{rodrigue2020geography}
J.-P. Rodrigue, {\em The geography of transport systems}.
\newblock Routledge, 2020.

\bibitem{otero2018health}
I.~Otero, M.~Nieuwenhuijsen, and D.~Rojas-Rueda, ``Health impacts of bike
  sharing systems in europe,'' {\em Environment international}, vol.~115,
  pp.~387--394, 2018.

\bibitem{hulot2018towards}
P.~Hulot, D.~Aloise, and S.~D. Jena, ``Towards station-level demand prediction
  for effective rebalancing in bike-sharing systems,'' in {\em Proceedings of
  the 24th ACM SIGKDD International Conference on Knowledge Discovery \& Data
  Mining}, pp.~378--386, 2018.

\bibitem{raviv2013optimal}
T.~Raviv and O.~Kolka, ``Optimal inventory management of a bike-sharing
  station,'' {\em Iie Transactions}, vol.~45, no.~10, pp.~1077--1093, 2013.

\bibitem{7313194}
S.~Reiss and K.~Bogenberger, ``{GPS}-data analysis of {Munich's} free-floating
  bike sharing system and application of an operator-based relocation
  strategy,'' in {\em 2015 IEEE 18th International Conference on Intelligent
  Transportation Systems}, pp.~584--589, 2015.

\bibitem{raviv2013static}
T.~Raviv, M.~Tzur, and I.~A. Forma, ``Static repositioning in a bike-sharing
  system: models and solution approaches,'' {\em EURO Journal on Transportation
  and Logistics}, vol.~2, no.~3, pp.~187--229, 2013.

\bibitem{yao2018deep}
H.~Yao, F.~Wu, J.~Ke, X.~Tang, Y.~Jia, S.~Lu, P.~Gong, J.~Ye, and Z.~Li, ``Deep
  multi-view spatial-temporal network for taxi demand prediction,'' in {\em
  Proceedings of the AAAI Conference on Artificial Intelligence}, vol.~32,
  2018.

\bibitem{yao2019revisiting}
H.~Yao, X.~Tang, H.~Wei, G.~Zheng, and Z.~Li, ``Revisiting spatial-temporal
  similarity: A deep learning framework for traffic prediction,'' in {\em
  Proceedings of the AAAI conference on artificial intelligence}, vol.~33,
  pp.~5668--5675, 2019.

\bibitem{jeon2014effects}
M.~Jeon, B.~N. Walker, and J.-B. Yim, ``Effects of specific emotions on
  subjective judgment, driving performance, and perceived workload,'' {\em
  Transportation research part F: traffic psychology and behaviour}, vol.~24,
  pp.~197--209, 2014.

\bibitem{asakura2017incident}
Y.~Asakura, T.~Kusakabe, L.~X. Nguyen, and T.~Ushiki, ``Incident detection
  methods using probe vehicles with on-board {GPS} equipment,'' {\em
  Transportation research part C: emerging technologies}, vol.~81,
  pp.~330--341, 2017.

\bibitem{hawas2007fuzzy}
Y.~E. Hawas, ``A fuzzy-based system for incident detection in urban street
  networks,'' {\em Transportation Research Part C: Emerging Technologies},
  vol.~15, no.~2, pp.~69--95, 2007.

\bibitem{wang2018locality}
Q.~Wang, J.~Wan, and Y.~Yuan, ``Locality constraint distance metric learning
  for traffic congestion detection,'' {\em Pattern Recognition}, vol.~75,
  pp.~272--281, 2018.

\bibitem{8886013}
K.~Yardy, A.~Almehmadi, and K.~El-Khatib, ``Detecting malicious driving with
  machine learning,'' in {\em 2019 IEEE Wireless Communications and Networking
  Conference (WCNC)}, pp.~1--6, 2019.

\bibitem{li2016lane}
K.~Li, X.~Wang, Y.~Xu, and J.~Wang, ``Lane changing intention recognition based
  on speech recognition models,'' {\em Transportation Research Part C: Emerging
  Technologies}, vol.~69, 2016.

\bibitem{tang2020driver}
L.~Tang, H.~Wang, W.~Zhang, Z.~Mei, and L.~Li, ``Driver lane change intention
  recognition of intelligent vehicle based on long short-term memory network,''
  {\em IEEE Access}, vol.~8, pp.~136898--136905, 2020.

\bibitem{8813987}
T.~Han, J.~Jing, and m.~Özgüner, ``Driving intention recognition and lane
  change prediction on the highway,'' in {\em 2019 IEEE Intelligent Vehicles
  Symposium (IV)}, pp.~957--962, 2019.

\bibitem{wu2021admm}
H.~Wu, N.~E. O’Connor, J.~Bruton, and M.~Liu, ``An admm-based optimal
  transmission frequency management system for {IoT} edge intelligence,'' in
  {\em 2021 IEEE 7th World Forum on Internet of Things (WF-IoT)}, pp.~217--222,
  IEEE, 2021.

\bibitem{wu2022real}
H.~Wu, N.~E. O’Connor, J.~Bruton, A.~Hall, and M.~Liu, ``Real-time anomaly
  detection for an admm-based optimal transmission frequency management system
  for {IoT} devices,'' {\em Sensors}, vol.~22, no.~16, p.~5945, 2022.

\bibitem{bienko2015ibm}
C.~Bienko, M.~Greenstein, S.~Holt, and R.~Phillips, {\em IBM Cloudant: Database
  as a Service Advanced Topics}.
\newblock IBM Redbooks, 2015.

\bibitem{2020A}
A.~K. Tyagi, K.~Agarwal, D.~Goyal, and N.~Sreenath, ``A review on security and
  privacy issues in internet of things,'' in {\em Advances in Computing and
  Intelligent Systems}, pp.~489--502, Springer, 2020.

\bibitem{7106504}
R.~{Zhang}, K.~{Wu}, M.~{Li}, and J.~{Wang}, ``Online resource scheduling under
  concave pricing for cloud computing,'' {\em IEEE Transactions on Parallel and
  Distributed Systems}, vol.~27, no.~4, pp.~1131--1145, 2016.

\bibitem{2019Utility}
B.~Bian, X.~Chen, and Z.~Q. Xu, ``Utility maximization under trading
  constraints with discontinuous utility,'' {\em SIAM Journal on Financial
  Mathematics}, vol.~10, no.~1, pp.~243--260, 2019.

\bibitem{LV202190}
Z.~Lv, D.~Chen, R.~Lou, and Q.~Wang, ``Intelligent edge computing based on
  machine learning for smart city,'' {\em Future Generation Computer Systems},
  vol.~115, pp.~90 -- 99, 2021.

\bibitem{boyd2011distributed}
S.~Boyd, N.~Parikh, and E.~Chu, {\em Distributed optimization and statistical
  learning via the alternating direction method of multipliers}.
\newblock Now Publishers Inc, 2011.

\bibitem{8408731}
X.~Liu, Y.~Liu, A.~Liu, and L.~T. Yang, ``Defending on–off attacks using
  light probing messages in smart sensors for industrial communication
  systems,'' {\em IEEE Transactions on Industrial Informatics}, vol.~14, no.~9,
  pp.~3801--3811, 2018.

\bibitem{8379722}
E.~Anthi, L.~Williams, and P.~Burnap, ``Pulse: An adaptive intrusion detection
  for the internet of things,'' in {\em Living in the Internet of Things:
  Cybersecurity of the IoT - 2018}, pp.~1--4, 2018.

\bibitem{7474197}
A.~Ukil, S.~Bandyoapdhyay, C.~Puri, and A.~Pal, ``{IoT} healthcare analytics:
  The importance of anomaly detection,'' in {\em 2016 IEEE 30th International
  Conference on Advanced Information Networking and Applications (AINA)},
  pp.~994--997, 2016.

\bibitem{traffic}
Y.~Hu, A.~Qu, and D.~Work, ``Graph convolutional networks for traffic
  anomaly,'' {\em CoRR}, vol.~abs/2012.13637, 2020.

\bibitem{miller2016detection}
B.~N. Miller, {\em Detection of malicious content in JSON structured data using
  multiple concurrent anomaly detection methods}.
\newblock Eastern Michigan University, 2016.

\bibitem{srikant2004mathematics}
R.~Srikant, {\em The mathematics of Internet congestion control}.
\newblock Springer Science \& Business Media, 2004.

\bibitem{gordon2012karush}
G.~Gordon and R.~Tibshirani, ``{Karush-Kuhn-Tucker} conditions,'' {\em
  Optimization}, vol.~10, no.~725/36, p.~725, 2012.

\bibitem{schmidhuber1997long}
J.~Schmidhuber, S.~Hochreiter, {\em et~al.}, ``Long short-term memory,'' {\em
  Neural Comput}, vol.~9, no.~8, pp.~1735--1780, 1997.

\bibitem{ergen2019unsupervised}
T.~Ergen and S.~S. Kozat, ``Unsupervised anomaly detection with {LSTM} neural
  networks,'' {\em IEEE transactions on neural networks and learning systems},
  vol.~31, no.~8, pp.~3127--3141, 2019.

\bibitem{lindemann2021survey}
B.~Lindemann, B.~Maschler, N.~Sahlab, and M.~Weyrich, ``A survey on anomaly
  detection for technical systems using {LSTM} networks,'' {\em Computers in
  Industry}, vol.~131, p.~103498, 2021.

\bibitem{aljbali2020anomaly}
S.~Aljbali and K.~Roy, ``Anomaly detection using bidirectional {LSTM},'' in
  {\em Proceedings of SAI Intelligent Systems Conference}, pp.~612--619,
  Springer, 2020.

\bibitem{thill2019anomaly}
M.~Thill, S.~D{\"a}ubener, W.~Konen, T.~B{\"a}ck, P.~Barancikova, M.~Holena,
  T.~Horvat, M.~Pleva, and R.~Rosa, ``Anomaly detection in electrocardiogram
  readings with stacked {LSTM} networks,'' in {\em Proceedings of the 19th
  Conference Information Technologies-Applications and Theory (ITAT 2019)},
  pp.~17--25, CEUR-WS, 2019.

\bibitem{xia2021new}
L.~Xia and Z.~Li, ``A new method of abnormal behavior detection using {LSTM}
  network with temporal attention mechanism,'' {\em The Journal of
  Supercomputing}, vol.~77, no.~4, pp.~3223--3241, 2021.

\bibitem{nguyen2021forecasting}
H.~Nguyen, K.~P. Tran, S.~Thomassey, and M.~Hamad, ``Forecasting and anomaly
  detection approaches using {LSTM} and {LSTM} autoencoder techniques with the
  applications in supply chain management,'' {\em International Journal of
  Information Management}, vol.~57, p.~102282, 2021.

\bibitem{5990666}
Z.~Fan, ``Distributed demand response and user adaptation in smart grids,'' in
  {\em 12th IFIP/IEEE International Symposium on Integrated Network Management
  (IM 2011) and Workshops}, pp.~726--729, 2011.

\bibitem{chen2021comparative}
Z.~Chen, H.~Wu, N.~E. O'Connor, and M.~Liu, ``A comparative study of using
  spatial-temporal graph convolutional networks for predicting availability in
  bike sharing schemes,'' in {\em 2021 IEEE International Intelligent
  Transportation Systems Conference (ITSC)}, pp.~1299--1305, IEEE, 2021.

\bibitem{shi2015convolutional}
X.~Shi, Z.~Chen, H.~Wang, D.~Y. Yeung, W.~K. Wong, and W.~C. Woo,
  ``Convolutional lstm network: A machine learning approach for precipitation
  nowcasting,'' {\em Advances in neural information processing systems},
  vol.~2015, pp.~802--810, 2015.

\bibitem{wu2016short}
Y.~Wu and H.~Tan, ``Short-term traffic flow forecasting with spatial-temporal
  correlation in a hybrid deep learning framework,'' {\em arXiv preprint
  arXiv:1612.01022}, 2016.

\bibitem{DBLP:conf/nips/DefferrardBV16}
M.~Defferrard, X.~Bresson, and P.~Vandergheynst, ``Convolutional neural
  networks on graphs with fast localized spectral filtering,'' in {\em Advances
  in Neural Information Processing Systems 29: Annual Conference on Neural
  Information Processing Systems 2016, December 5-10, 2016, Barcelona, Spain},
  pp.~3837--3845, 2016.

\bibitem{bai2019stg2seq}
L.~Bai, L.~Yao, S.~Kanhere, X.~Wang, and Q.~Sheng, ``Stg2seq: Spatial-temporal
  graph to sequence model for multi-step passenger demand forecasting,'' in
  {\em Proceedings of the Twenty-Eighth International Joint Conference on
  Artificial Intelligence, {IJCAI-19}}, pp.~1981--1987, International Joint
  Conferences on Artificial Intelligence Organization, 2019.

\bibitem{zhang2017deep}
J.~Zhang, Y.~Zheng, and D.~Qi, ``Deep spatio-temporal residual networks for
  citywide crowd flows prediction,'' in {\em Proceedings of the AAAI Conference
  on Artificial Intelligence}, vol.~31, 2017.

\bibitem{chu2020passenger}
J.~Chu, X.~Wang, K.~Qian, L.~Yao, F.~Xiao, J.~Li, and Z.~Yang, ``Passenger
  demand prediction with cellular footprints,'' {\em IEEE Transactions on
  Mobile Computing}, 2020.

\bibitem{velivckovic2017graph}
P.~Veli{\v{c}}kovi{\'c}, G.~Cucurull, A.~Casanova, A.~Romero, P.~Lio, and
  Y.~Bengio, ``Graph attention networks,'' {\em arXiv preprint
  arXiv:1710.10903}, 2017.

\bibitem{chen2020multitask}
Z.~Chen, B.~Zhao, Y.~Wang, Z.~Duan, and X.~Zhao, ``Multitask learning and
  {GCN}-based taxi demand prediction for a traffic road network,'' {\em
  Sensors}, vol.~20, no.~13, p.~3776, 2020.

\bibitem{kim2019graph}
T.~S. Kim, W.~K. Lee, and S.~Y. Sohn, ``Graph convolutional network approach
  applied to predict hourly bike-sharing demands considering spatial, temporal,
  and global effects,'' {\em PloS one}, vol.~14, no.~9, p.~e0220782, 2019.

\bibitem{wu2019graph}
Z.~Wu, S.~Pan, G.~Long, J.~Jiang, and C.~Zhang, ``Graph wavenet for deep
  spatial-temporal graph modeling,'' in {\em International Joint Conference on
  Artificial Intelligence 2019}, pp.~1907--1913, Association for the
  Advancement of Artificial Intelligence (AAAI), 2019.

\bibitem{chiang2019cluster}
W.-L. Chiang, X.~Liu, S.~Si, Y.~Li, S.~Bengio, and C.-J. Hsieh,
  ``Cluster-{GCN}: An efficient algorithm for training deep and large graph
  convolutional networks,'' in {\em Proceedings of the 25th ACM SIGKDD
  International Conference on Knowledge Discovery \& Data Mining},
  pp.~257--266, 2019.

\bibitem{shiraki2020spatial}
K.~Shiraki, T.~Hirakawa, T.~Yamashita, and H.~Fujiyoshi, ``Spatial temporal
  attention graph convolutional networks with mechanics-stream for
  skeleton-based action recognition,'' in {\em Proceedings of the Asian
  Conference on Computer Vision}, 2020.

\bibitem{zhang2020sta}
W.~Zhang, Z.~Lin, J.~Cheng, C.~Ma, X.~Deng, and H.~Wang, ``{STA-GCN}:
  two-stream graph convolutional network with spatial--temporal attention for
  hand gesture recognition,'' {\em The Visual Computer}, vol.~36, no.~10,
  pp.~2433--2444, 2020.

\bibitem{2016DNN}
J.~Zhang, Y.~Zheng, D.~Qi, R.~Li, and X.~Yi, ``{DNN-based} prediction model for
  spatio-temporal data,'' in {\em the 24th ACM SIGSPATIAL International
  Conference}, 2016.

\bibitem{xgboost2016}
T.~Chen and C.~Guestrin, ``Xgboost: A scalable tree boosting system,'' in {\em
  Proceedings of the 22nd ACM SIGKDD International Conference on Knowledge
  Discovery and Data Mining}, KDD '16, (New York, NY, USA), p.~785–794,
  Association for Computing Machinery, 2016.

\bibitem{lstm1997}
S.~Hochreiter and J.~Schmidhuber, ``Long short-term memory,'' {\em Neural
  computation}, vol.~9, no.~8, pp.~1735--1780, 1997.

\bibitem{2017arXiv170701926L}
Y.~{Li}, R.~{Yu}, C.~{Shahabi}, and Y.~{Liu}, ``{Diffusion Convolutional
  Recurrent Neural Network: Data-Driven Traffic Forecasting},'' {\em arXiv
  e-prints}, p.~arXiv:1707.01926, July 2017.

\bibitem{kingma2017adam}
D.~P. Kingma and J.~Ba, ``Adam: A method for stochastic optimization,'' 2017.

\bibitem{wu2022lane}
H.~Wu and M.~Liu, ``{Lane-GNN}: Integrating {GNN} for predicting drivers lane
  change intention,'' {\em arXiv preprint arXiv:2207.00824}, 2022.

\bibitem{calibaba2017road}
J.~Calibaba, ``The road to vision zero: Zero traffic fatalities and serious
  injuries,'' 2017.

\bibitem{mejia2021vehicle}
H.~Mejia, E.~Palomo, E.~L{\'o}pez-Rubio, I.~Pineda, and R.~Fonseca, ``Vehicle
  speed estimation using computer vision and evolutionary camera calibration,''
  in {\em NeurIPS 2021 Workshop LatinX in AI}, 2021.

\bibitem{nam2020deep}
D.~Nam, R.~Lavanya, R.~Jayakrishnan, I.~Yang, and W.~H. Jeon, ``A deep learning
  approach for estimating traffic density using data obtained from connected
  and autonomous probes,'' {\em Sensors}, vol.~20, no.~17, p.~4824, 2020.

\bibitem{kuvsic2020extended}
K.~Ku{\v{s}}i{\'c}, I.~Dusparic, M.~Gu{\'e}riau, M.~Greguri{\'c}, and
  E.~Ivanjko, ``Extended variable speed limit control using multi-agent
  reinforcement learning,'' in {\em 2020 IEEE 23rd International Conference on
  Intelligent Transportation Systems (ITSC)}, pp.~1--8, IEEE, 2020.

\bibitem{7350149}
M.~Liu, R.~H. Ordóñez-Hurtado, F.~Wirth, Y.~Gu, E.~Crisostomi, and
  R.~Shorten, ``A distributed and privacy-aware speed advisory system for
  optimizing conventional and electric vehicle networks,'' {\em IEEE
  Transactions on Intelligent Transportation Systems}, vol.~17, no.~5,
  pp.~1308--1318, 2016.

\bibitem{liu2021mpc}
M.~Liu, L.~Cheng, Y.~Gu, Y.~Wang, Q.~Liu, and N.~E. O'Connor, ``{MPC-CSAS}:
  Multi-party computation for real-time privacy-preserving speed advisory
  systems,'' {\em IEEE Transactions on Intelligent Transportation Systems},
  2021.

\bibitem{li2013impacts}
H.~Li, D.~J. Graham, and A.~Majumdar, ``The impacts of speed cameras on road
  accidents: An application of propensity score matching methods,'' {\em
  Accident Analysis \& Prevention}, vol.~60, pp.~148--157, 2013.

\bibitem{mandalia2005using}
H.~M. Mandalia and M.~D.~D. Salvucci, ``Using support vector machines for
  lane-change detection,'' in {\em Proceedings of the human factors and
  ergonomics society annual meeting}, vol.~49, pp.~1965--1969, SAGE
  Publications Sage CA: Los Angeles, CA, 2005.

\bibitem{7835731}
H.~Woo, Y.~Ji, H.~Kono, Y.~Tamura, Y.~Kuroda, T.~Sugano, Y.~Yamamoto,
  A.~Yamashita, and H.~Asama, ``Lane-change detection based on
  vehicle-trajectory prediction,'' {\em IEEE Robotics and Automation Letters},
  vol.~2, no.~2, pp.~1109--1116, 2017.

\bibitem{hounsell2009review}
N.~Hounsell, B.~Shrestha, J.~Piao, and M.~McDonald, ``Review of urban traffic
  management and the impacts of new vehicle technologies,'' {\em IET
  intelligent transport systems}, vol.~3, no.~4, pp.~419--428, 2009.

\bibitem{tradisauskas2009map}
N.~Tradisauskas, J.~Juhl, H.~Lahrmann, and C.~S. Jensen, ``Map matching for
  intelligent speed adaptation,'' {\em IET Intelligent Transport Systems},
  vol.~3, no.~1, pp.~57--66, 2009.

\bibitem{gu2018design}
Y.~Gu, M.~Liu, M.~Souza, and R.~N. Shorten, ``On the design of an intelligent
  speed advisory system for cyclists,'' in {\em 2018 21st International
  Conference on Intelligent Transportation Systems (ITSC)}, pp.~3892--3897,
  IEEE, 2018.

\bibitem{chen2021intelligent}
B.~Chen, M.~Liu, Y.~Zhang, Z.~Chen, Y.~Gu, and N.~E. O’Connor, ``An
  intelligent multi-speed advisory system using improved whale optimisation
  algorithm,'' in {\em 2021 IEEE 93rd Vehicular Technology Conference
  (VTC2021-Spring)}, pp.~1--6, IEEE, 2021.

\bibitem{liu2015topics}
M.~Liu, {\em Topics in electromobility and related applications}.
\newblock PhD thesis, National University of Ireland Maynooth, 2015.

\bibitem{huang2014deep}
W.~Huang, G.~Song, H.~Hong, and K.~Xie, ``Deep architecture for traffic flow
  prediction: deep belief networks with multitask learning,'' {\em IEEE
  Transactions on Intelligent Transportation Systems}, vol.~15, no.~5,
  pp.~2191--2201, 2014.

\bibitem{lv2014traffic}
Y.~Lv, Y.~Duan, W.~Kang, Z.~Li, and F.-Y. Wang, ``Traffic flow prediction with
  big data: a deep learning approach,'' {\em IEEE Transactions on Intelligent
  Transportation Systems}, vol.~16, no.~2, pp.~865--873, 2014.

\bibitem{tian2018lstm}
Y.~Tian, K.~Zhang, J.~Li, X.~Lin, and B.~Yang, ``{LSTM-based} traffic flow
  prediction with missing data,'' {\em Neurocomputing}, vol.~318, pp.~297--305,
  2018.

\bibitem{8526506}
W.~Jiang and L.~Zhang, ``Geospatial data to images: A deep-learning framework
  for traffic forecasting,'' {\em Tsinghua Science and Technology}, vol.~24,
  no.~1, pp.~52--64, 2019.

\bibitem{ma2017learning}
X.~Ma, Z.~Dai, Z.~He, J.~Ma, Y.~Wang, and Y.~Wang, ``Learning traffic as
  images: a deep convolutional neural network for large-scale transportation
  network speed prediction,'' {\em Sensors}, vol.~17, no.~4, p.~818, 2017.

\bibitem{wu2020comprehensive}
Z.~Wu, S.~Pan, F.~Chen, G.~Long, C.~Zhang, and S.~Y. Philip, ``A comprehensive
  survey on graph neural networks,'' {\em IEEE transactions on neural networks
  and learning systems}, vol.~32, no.~1, pp.~4--24, 2020.

\bibitem{li2018dcrnn_traffic}
Y.~Li, R.~Yu, C.~Shahabi, and Y.~Liu, ``Diffusion convolutional recurrent
  neural network: Data-driven traffic forecasting,'' in {\em International
  Conference on Learning Representations (ICLR '18)}, 2018.

\bibitem{9413270}
T.~Mallick, P.~Balaprakash, E.~Rask, and J.~Macfarlane, ``Transfer learning
  with graph neural networks for short-term highway traffic forecasting,'' in
  {\em 2020 25th International Conference on Pattern Recognition (ICPR)},
  pp.~10367--10374, 2021.

\bibitem{behrisch2011sumo}
M.~Behrisch, L.~Bieker, J.~Erdmann, and D.~Krajzewicz, ``Sumo--simulation of
  urban mobility: an overview,'' in {\em Proceedings of SIMUL 2011, The Third
  International Conference on Advances in System Simulation}, ThinkMind, 2011.

\bibitem{773549}
C.-I. Chang, ``Spectral information divergence for hyperspectral image
  analysis,'' in {\em IEEE 1999 International Geoscience and Remote Sensing
  Symposium. IGARSS'99 (Cat. No.99CH36293)}, vol.~1, pp.~509--511 vol.1, 1999.

\bibitem{deng2021graph}
A.~Deng and B.~Hooi, ``Graph neural network-based anomaly detection in
  multivariate time series,'' in {\em Proceedings of the AAAI Conference on
  Artificial Intelligence}, vol.~35, pp.~4027--4035, 2021.

\bibitem{zhao2020multivariate}
H.~Zhao, Y.~Wang, J.~Duan, C.~Huang, D.~Cao, Y.~Tong, B.~Xu, J.~Bai, J.~Tong,
  and Q.~Zhang, ``Multivariate time-series anomaly detection via graph
  attention network,'' in {\em 2020 IEEE International Conference on Data
  Mining (ICDM)}, pp.~841--850, IEEE, 2020.

\bibitem{huang2020hitanomaly}
S.~Huang, Y.~Liu, C.~Fung, R.~He, Y.~Zhao, H.~Yang, and Z.~Luan, ``Hitanomaly:
  Hierarchical transformers for anomaly detection in system log,'' {\em IEEE
  Transactions on Network and Service Management}, vol.~17, no.~4,
  pp.~2064--2076, 2020.

\bibitem{jun2010partial}
M.~Jun, D.~Woo, and E.-Y. Chung, ``Partial connection-aware topology synthesis
  for on-chip cascaded crossbar network,'' {\em IEEE Transactions on
  Computers}, vol.~61, no.~1, pp.~73--86, 2010.

\end{thebibliography}

\end{document}